\documentclass[english,letterpaper, 10 pt, journal, twoside]{IEEEtran}

\IEEEoverridecommandlockouts

\usepackage{biblatex}
\usepackage{color}
\usepackage{babel}
\usepackage{verbatim}
\usepackage{float}
\usepackage{amsmath}
\usepackage{amssymb}
\usepackage{graphicx}
\usepackage{epstopdf}
\usepackage[left=48pt,right=48pt,top=57pt,bottom=43pt]{geometry}
\usepackage[caption=false]{subfig}
\usepackage{setspace,algorithm,algorithmicx,algpseudocode}
\usepackage{xpatch}
\makeatletter
\xpatchcmd{\algorithmic}{\ALG@tlm\z@}{\ALG@tlm\z@\leftmargin 0pt}{}{}
\renewcommand{\ALG@beginalgorithmic}{\small}

\def\bstctlcite#1{\@bsphack
\@for\@citeb:=#1\do{%
\edef\@citeb{\expandafter\@firstofone\@citeb}%
\if@filesw\immediate\write\@auxout{\string\citation{\@citeb}}\fi}%
\@esphack}
\makeatother

\usepackage{setspace,algorithm,algorithmicx,algpseudocode}

\usepackage{xcolor}

\newcommand{\Ag}{\mbox{$\alpha$}}

\newcommand{\Bg}{\mbox{$\beta$}}

\newcommand{\Gg}{\mbox{$\gamma$}}

\newcommand{\Kg}{\mbox{$\kappa$}}
\newcommand{\kp}{\mbox{$\kappa$}}

\newcommand{\Lg}{\mbox{$\lambda$}}
\newcommand{\Lgs}{\mbox{\small $\lambda$}}

\newcommand{\Sg}{\mbox{$\sigma$}}

\newcommand{\Bs}{\mbox{$\cal B$}}

\newcommand{\Cs}{\mbox{$\cal C$}}

\newcommand{\Es}{\mbox{$\cal E$}}

\newcommand{\Os}{\mbox{$\cal O$}}

\newcommand{\Ps}{\mbox{$\cal P$}}

\newcommand{\real}{\mbox{$I \!\! R$}}

\newcommand{\hlf}{ \mbox{$ \frac{1}{2} $} }

\newcommand{\nin}{ \mbox{$ \not{\!\in} $} }

\newcommand{\xb}{\mbox{$\bar{x}$}}

\newcommand{\absvl}[1]{ \left| #1 \right| }

\newcommand{\rtxt}[1]{ \hspace{1em}\mbox{#1} }

\newcommand{ \vctwo }[2] {
 \left( \begin{array}{c }
  #1        \\
  #2
\end{array} \right)  }

\newcommand{ \matwo }[4] {
 \left[ \begin{array}{c c }
   #1  &  #2     \\
   #3  &  #4
\end{array} \right]  }

\def \begpf{ \begin{description} \item[{\bf Proof:\hspace{1.5em} }]  }
\def \endpf{ \begin{flushright} $\Box $ \end{flushright}   \end{description}}

\def \epf{ \hfill $\square$ }

\def \eex{ \hfill $\circ$ }
\def \begpf{ \begin{description} \item[{\bf Proof:\hspace{0.75em} }]  }
\def \endpf{ \begin{flushright} $\Box $ \end{flushright}   \end{description}}
\def \begrm{ \begin{description} \item[{\bf Remark:\hspace{0.5em} }]  }

\def \endrm{  \end{description} }
\def \begex{ \begin{description} \item[{\bf Example:\hspace{0.5em} }]  }
\def \endex{  \end{description} }

\newcommand{\beq}[1]{ \begin{equation} \label{eq.#1} }
\newcommand{\eeq}{ \end{equation} }

\newcommand{\blemm}[1]{ \begin{lemm} \label{lemm.#1} }
\newcommand{\elemm}{ \end{lemm} }
\newcommand{\barr}{ \begin{array}{cl} }
\newcommand{\earr}{ \end{array} }

\color{black}

\begin{document}


\title{\LARGE\bf Steering Flexible Linear Objects in\\[2pt]
 Planar  Environments by Two Robot Hands\\[2pt]
 Using Euler's Elastica Solutions}

\author{A. Levin$^{1}$ and E. D. Rimon$^{1}$ and A. Shapiro$^{2}$
\thanks{This work receives funding from the European Union's Horizon Europe research
and innovation program under grant agreement no. 101070600, project SoftEnable. $^{1}\!\!$~Dept.~of~{\scriptsize ME,}~Technion,~Israel.  $^{2}\!\!$~Dept. of {\scriptsize ME,}~Ben-Gurion~University,~Israel.}
\vspace{-.3in}
}

\date{}


\maketitle

\thispagestyle{empty}

\pagestyle{empty}

\vspace{-.2in}

\begin{abstract}
The manipulation of flexible objects
such as cables, wires and fresh food items by robot hands
forms a special challenge in robot grasp mechanics.
This paper considers the steering of flexible linear objects in planar environments by two robot hands.
The flexible linear object, modeled as an elastic non-stretchable rod, is manipulated by varying the gripping endpoint positions
while keeping equal endpoint tangents. The flexible linear object shape has a~closed form solution in terms of the grasp endpoint positions and tangents, called {\em Euler's elastica.} This paper obtains the elastica solutions under the optimal control framework, then uses the elastica solutions to obtain closed form criteria for {\em non self-intersection,  stability  and  obstacle avoidance} of the flexible linear object. The new tools are incorporated into a~planning scheme for steering flexible linear objects in planar environments populated by sparsely spaced obstacles. The scheme  is fully implemented and demonstrated with detailed examples.
\end{abstract}

%
%
\vspace{-.15in}

\section{\bf Introduction}
\vspace{-.02in}

\noindent   Robotic manipulation of flexible linear objects such as cables, wires and fresh food items forms a~special
challenge in robot grasp mechanics. In these problems, one or two robot hands apply endpoint forces and torques that
together with external influences such as gravity and contacts with the environment affect the object shape during manipulation.
Robotic applications include cable routing and untangling~\cite{cable_routing11,bretl_multiple,cable_rss22}, surgical suturing~\cite{jackson&cavusogl,javdani}, knot tying~\cite{balkcom15,hirai_knot}, compliant mechanisms~\cite{till17}, fresh food handling~\cite{food_survey22},  architectural elements fabrication~\cite{brander} and  agricultural robotics~\cite{adaptive_LDO22}.

This paper focuses on robotic steering of flexible linear objects in planar environments.
The object,  modeled as a~non-stretchable elastic rod, is to be steered from start to target positions
by two robot hands while avoiding self-collision and contact with the environment,
 except  for endpoint contacts at the start or target positions (Fig.~1).
Examples of flexible linear objects that can be steered in two-dimensions
are strip like objects such as ribbon cables,  plastic ties and fresh food items, since
such objects preserve their strip flatness during manipulation by two robot hands.
This paper provides analysis of flexible linear objects  mechanics 
in two-dimensions, then a~scheme for steering such objects based on closed form solutions for their~shape~and stability that depend only on the gripping hands relative position during manipulation.

\begin{figure}
\centerline{\includegraphics[scale=0.25]{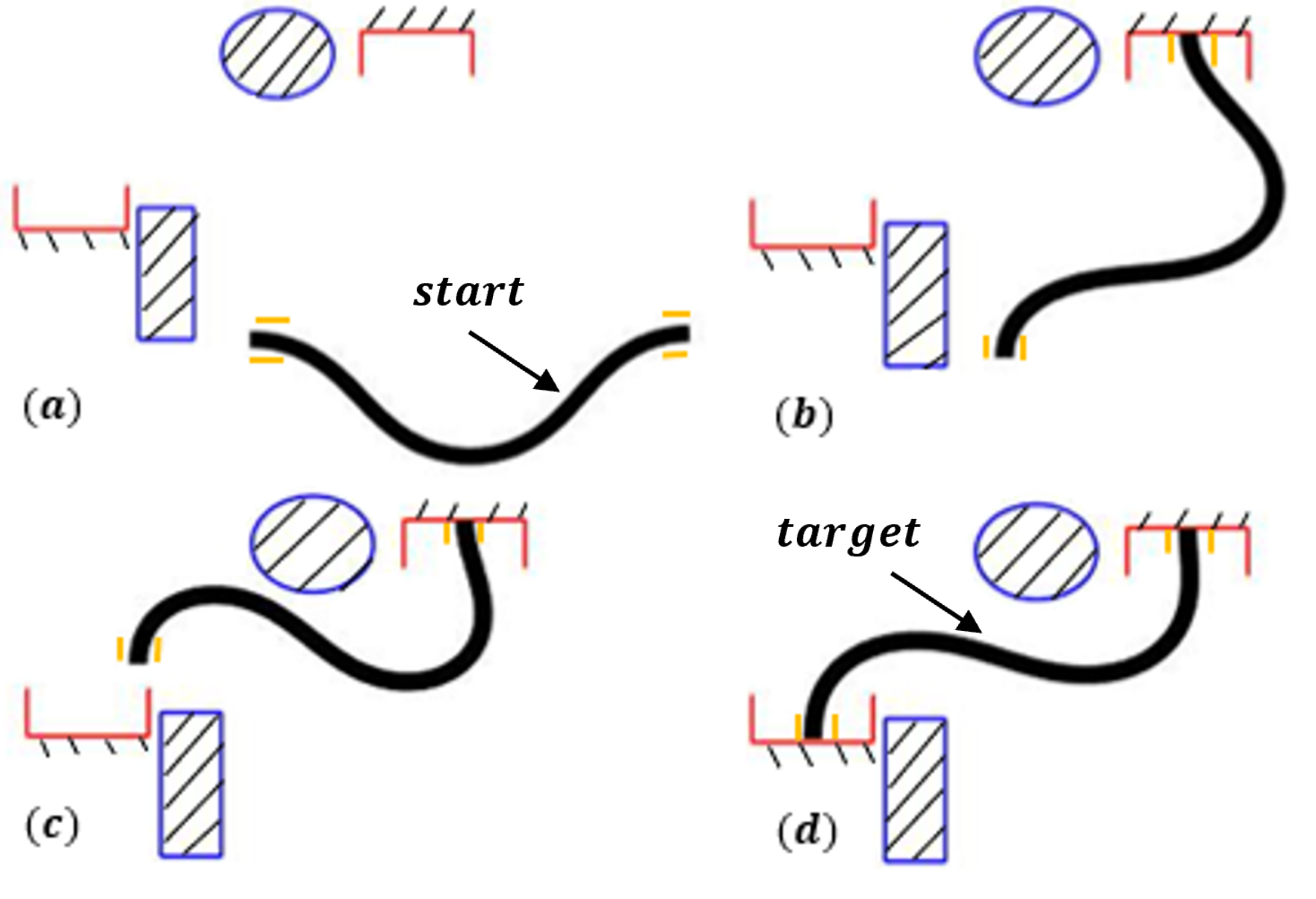}}
\vspace{-.2in}
\caption{Top view of a flexible cable steered by two robot grippers in a~planar environment populated by obstacles. The cable must reach target position 
while avoiding self-collision and contact with obstacles (except for endpoint contacts at the start or target).} 
\label{motiv.fig}
\vspace{-.24in}
\end{figure} 


{\bf Related work:}  We use the term {\em flexible cable} for flexible linear objects manipulated in two-dimensions.
The modeling of these objects  relies on {\em Euler-Bernoulli bending moment law}~\cite{goss_thesis}. When a~flexible cable is held  at an~equilibrium state by endpoint forces and moments in the plane, the cable curvature at every point is proportional  to the cable bending moment  at this point. The coefficient of proportionality is the {\em cable stiffness} at each point.

The mechanics literature has used the Euler-Bernoulli law to obtain solutions for flexible cable equilibrium shapes~in~two-dim\-ensions~\cite{love}. The solutions, called {\em Euler's elastica,} form a~one-parameter family of periodic shapes aligned along a~linear axis~\cite{levyakov10}. The axis is parallel to the opposing forces applied at the cable endpoints and passes through the zero curvature cable points. 
While flexible cable equilibrium shapes have been fully characterized in the mechanics literature,  verification of their stability typically relies on numerical techniques~\cite{batista15,levyakov09}. Sachkov~\cite{sachkov07,sachkov_conjugate,sachkov_vertices} describes analytic bounds~for~\mbox{flexible}~cable stability that are used in~this~paper's~\mbox{steering}~scheme. 

In the robotics~literature,~\mbox{sampling}~based~\mbox{approaches}~are~used to plan flexible cable steering paths.
Moll~\cite{kavraki06}~\mbox{samples}~\mbox{cable} endpoint positions, then computes their~\mbox{stable}~\mbox{shapes}~by~num\-erical optimization of the flexible cable total elastic energy.
Bretl~\cite{bretl_ijrr14} uses Sachkov's approach to describe the cable total elastic energy minimization as an optimal control problem. Bretl shows that the adjoint differential equation which describes the equilibrium shapes of a~flexible cable held by robot hands is fully determined by a~small number of {\em co-state variables}~\cite{bretl_ijrr14}. In {\small 2-D} environments these are endpoint forces and moments (three variables) while in {\small 3-D} environments these are endpoint forces and toques (six variables).

Using this insight, Bretl developed sampling based planners for steering flexible cables in {\small 2-D} and {\small 3-D} environments~\cite{bretl_ijrr14}. However, each sampled co-state requires solution of the adjoint differential equation to obtain the cable shape, then computation of the cable co-state to endpoint Jacobian
to verify that each sampled cable shape is stable.
Sintov~\cite{sintov20} extended this work into a~two stage approach. First a~roadmap of stable cable shapes is computed for sampled co-states.  Then a~cable steering path is computed in the physical environment using numerical self-intersection and obstacle~avoidance~tests.

While this paper focuses on Euler's elastica as a~means to model flexible object shapes,
robotic manipulation of heterogeneous flexible objects in agricultural robotics~\cite{zhu_survey22}
and fresh food handling~\cite{food_survey22} requires complementary approaches.  Papers
such as~\cite{adaptive_LDO22}, \cite{zhu_iros18} and \cite{lagneau20} describe
adaptive and learning based approaches to manipulation of such complex flexible objects. One hopes  that
this paper's modeling approach will become part of adaptive and learning based manipulation techniques
for complex flexible objects.

{\bf Paper contributions:}
This paper considers steering flexible cables in two-dimensions using two robot hands.
By maintaining equal endpoint tangents, 
the cable steering problem is reduced to {\em five
configuration variables:} the cable base-frame configuration and the cable endpoints relative position.
The paper starts with a~derivation of Euler's elastica solutions for flexible cable equilibrium shapes
under the optimal control approach.  The derivation complements  Bretle~\cite{bretl_ijrr14} that focused on the adjoint equation and the use of its co-states to steer flexible cables in {\small 3-D} environments. This paper focuses on flexible cable steering in two-dimensions in order to take advantage of the closed form elastica solutions.

When considering flexible cables in \mbox{\small 2-D} environments, Euler's elastica form periodic shapes aligned with a~linear  axis. This paper describes analytic equations that determine the cable shape in terms of the cable's relative endpoint positions
imposed by the robot hands.
The paper next describes the range of the elastica {\em modulus parameter} that
ensures non self-intersection, followed by 
a~simple geometric rule that ensures stability of the flexible cable during steering.

To ensure obstacle avoidance during
steering, the paper describes an approximation of
the flexible cable equilibrium shapes by 
quadratic arcs determined by control points located on the cable.
The piecewise quadratic approximation allows efficient collision checks
against obstacles in the environment. The new tools are incorporated into a~fully implemented planning scheme for steering flexible linear objects in planar environments populated by sparsely spaced obstacles.


The paper is structured as follows. Section~\ref{sec.elastica} derives the flexible cable equilibrium shapes 
and formulates the flexible cable steering problem. Section~\ref{sec.self_intersect} 
characterizes  the non self-intersecting flexible cable shapes. Section~\ref{sec.stability} describes a~geometric rule that ensures flexible cable stability during steering. Section~\ref{sec.bezier} describes an approximation of the flexible cable equilibrium shapes
by 
quadratic arcs.
Section~\ref{sec.scheme} incorporates these tools into  a~motion planning scheme that steers flexible cable among sparsely spaced obstacles.
Section~\ref{sec.examples} describes execution examples of the steering scheme. The conclusion discusses future research such as gravitational effects and interaction with the environment. An~appendix verifies formulas for flexible cable curvature and tangent used in this paper.
\vspace{-.02in}

\section{\bf Euler's Elastica as an Optimal Control Problem} \label{sec.elastica}
\vspace{-.02in}

\noindent  This section obtains the flexible cable equilibrium shapes as an~optimal control problem solution.
The solution parameters are then related
to the cable endpoint positions and the cable steering problem is formulated.

Consider a non-stretchable flexible cable of length $L$ in $\real^2$.
The cable is parametrized by $(x\mbox{\small $(s)$},y\mbox{\small $(s)$})$ for $s \!\in\! [0,L]$.
The cable state variables are its $(x\mbox{\small $(s)$},y\mbox{\small $(s)$})$ coordinates and
tangent direction $\phi(s)$, thus defining the {\em state vector} $S \!=\! (x,y,\phi)$.  Under arclength parametrization with unit norm tangent,
the cable curvature is given by $\kp(s) \!=\! \tfrac{d}{ds} \phi(s)$.
 The cable curvature forms a~continuous and piecewise smooth control input, $u(s) \!=\! \kp(s)$,
for the {\em cable system equations}~given~by
\vspace{-.08in}
\begin{equation} \label{eq.sys}
\barr
\dot{x}(s) &= \cos \phi(s)  \\
\dot{y}(s) &= \sin \phi(s) \\
\dot{\phi}(s) &= u(s)
\earr \rtxt{\hspace{1.5em} $s \in [0,L]$.}
\vspace{-.07in}
\end{equation}

\noindent Eq.~\eqref{eq.sys} states that the cable shape, parametrized by arclength, is determined by its
curvature as control input. When the flexible cable is modeled as an elastic rod and there are no external influences such as gravity and contacts with the environment, the cable's {\em total elastic energy} is given by
\vspace{-.07in}
\begin{equation} \label{eq.energy}
\Es =  \int_0^L \! \hlf \mbox{\small $ EI$} \!\cdot\! \kp^2(s) ds
\vspace{-.06in}
\end{equation}

\noindent where $E \!>\! 0 $ is the cable Young's modulus~of~\mbox{elasticity}~and $I \!>\! 0 $ is the cable cross-sectional 2'nd moment of inertia~\cite{timoshenko&goodier:1969}. The cable stiffness, {\small $EI$}, is assumed to be a~known parameter.\\
\indent Let two robot hands impose positions and tangents at the flexible cable endpoints,
\mbox{\small $S(0) \!=\! (x(0),y(0),\phi(0))$} and \mbox{\small $S(\mbox{\small $L$}) \!=\! (x(\mbox{\small $L$}),y(\mbox{\small $L$}),\phi(\mbox{\small $L$}))$}. The locally stable cable shapes are local minima of $\Es$ under the endpoint constraints and the cable fixed length constraint captured by Eq.~\eqref{eq.sys}.
Since $\Es$ represents {\em bending energy} that grows monotonically with increasing cable curvature
(up to plastic yield limit), stable cable shapes always exist when held with fixed endpoint positions and tangents.\\
\indent The {\em Hamiltonian}~\cite{ben-asher,opt_survey} of the cable system defined by  Eq.~\eqref{eq.sys} and the elastic energy~$\Es$
is given by
\vspace{-.05in}
\[
\barr
& H(\mbox{\small $S$}(s),\Lg(s),u(s))  \\
&= \Lg_x(s) \cos\phi(s) \!+\! \Lg_y(s) \sin\phi(s) \!+\! \Lg_{\phi}(s) u(s) \!+\!  \hlf \mbox{$\scriptsize EI$} \!\cdot\!  u^2(s)
\earr
\vspace{-.05in}
\]
\noindent where $\Lg(s) \!=\! (\Lg_x(s),\Lg_y(s),\Lg_{\phi}(s))$ are the {\em co-state variables.} The co-states
$(\Lg_x(s),\Lg_y(s))$ and $\Lg_{\phi}(s)$ represent internal force and bending moment that develop at the cable points.
The energy extremal cable shapes represent equilibrium states of the cable held by the robot hands.
Along an~extremal cable shape, the vector $\Lg(s)$ is determined by  the {\em adjoint equation}~\cite{pontryagin}
\vspace{-.07in}
\[
	\dot{\lambda}(s) = - \frac{\partial}{\partial S} H \big( \mbox{\small $S$}(s), \Lg(s), u(s)  \big) \rtxt{\hspace{1em}\mbox{$S = ( x,y, \phi)$}}
\vspace{-.04in}
\]
\noindent thus leading to the system of adjoint differntial equations
\vspace{-.04in}
\begin{align}\label{eq:AdjointSystem}
\begin{split}
\dot{\Lg}_x(s) &=0\\
\dot{\Lg}_y(s) &=0\\
\dot{\Lg}_{\phi}(s) &=    \Lg _x(s) \sin\phi(s) \!-\! \Lg_y(s) \cos\phi(s)
\end{split}
\end{align}
\vspace{-.14in}

\noindent while the control $u(s) \!=\! \kp(s)$ satisfies the additional condition
\vspace{-.11in}
\begin{equation} \label{eq.u_cond}
\frac{\partial}{\partial u} H\big( \mbox{\small $S$}(s), \Lg(s),  u(s)  \big)  = 0,
\vspace{-.06in}
\end{equation}

\noindent thus leading to the algebraic equation
\vspace{-.1in}
\begin{equation} \label{eq.partial_u}
\Lg_{\phi}(s)  + \mbox{\small $EI$} \!\cdot\! u(s) = 0
\vspace{-.09in}
\end{equation}
\noindent which is the Euler-Bernoulli bending moment law.

\begin{figure}
\centering \includegraphics[width=.45\textwidth]{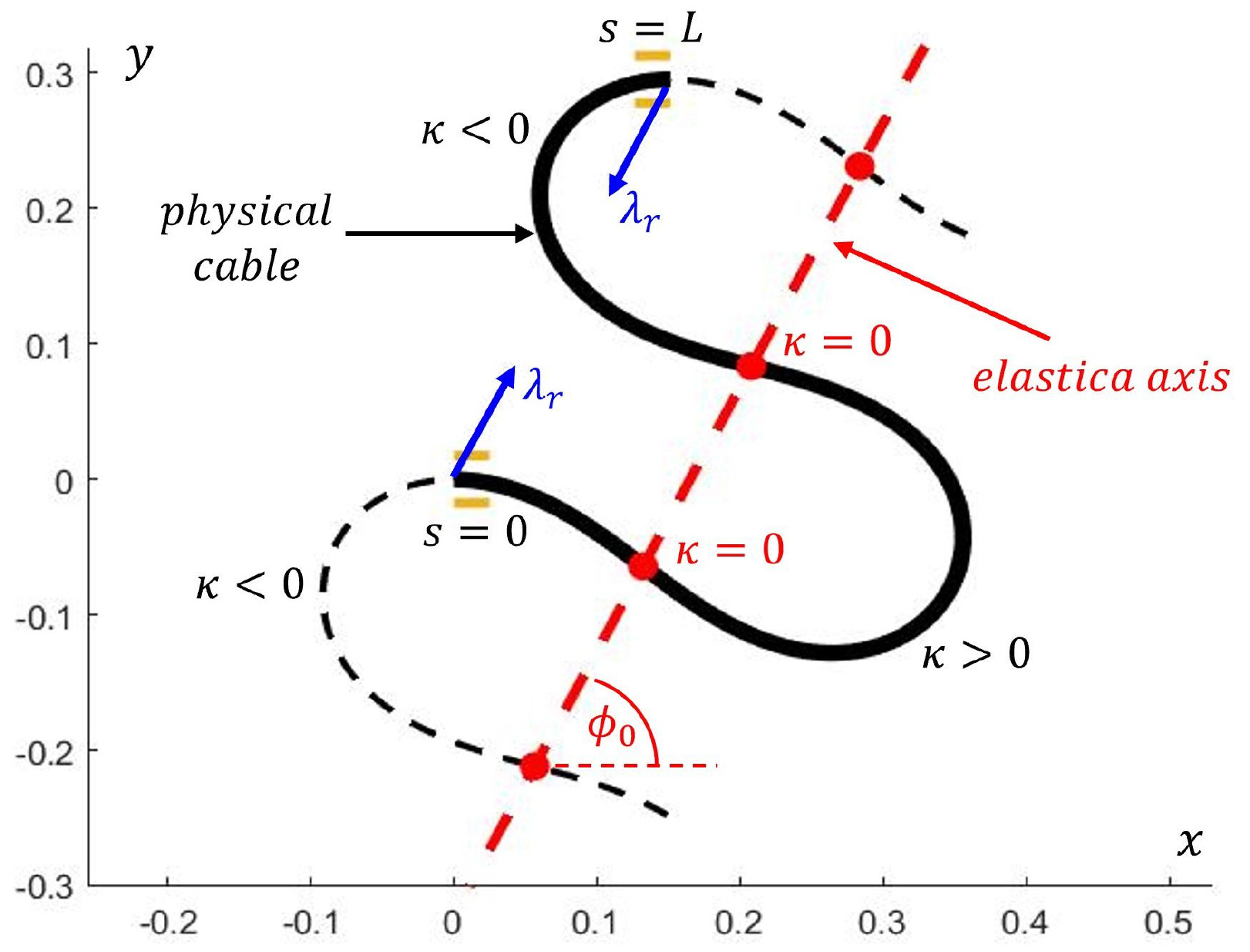}
\vspace{-.15in}
	\caption{Top view of a~flexible cable of length $L$ embedded in its periodic elastica solution. The elastica axis with angle $\phi_0$
passes through the elastica  zero curvature points and is parallel to the opposing forces of magnitude $\Lg_r$ applied at the cable endpoints.}
\label{el_axis.fig}
\vspace{-.25in}
	\end{figure}

From Eq.~\eqref{eq:AdjointSystem} one obtains that $\Lg_x(s)$ and $\Lg_y(s)$ are constants along energy extremal cable shapes. These constants
define the co-state parameters $\Lgs_r$ and $\phi_0$
\vspace{-.08in}
\[
\vctwo{\Lg_x}{\Lg_y} = \Lg_r \!\cdot\! \vctwo{\cos\phi_0}{\sin \phi_0} \rtxt{ $\Lg_r \!=\! \mbox{\small $\sqrt{ \Lg^2_x + \Lg^2_y}$ }$ }
\vspace{-.08in}
\]
\noindent which represent the magnitude  and direction of the opposing forces applied at the cable endpoints (blue arrows in Fig.~\ref{el_axis.fig}).
The system described by Eq.~\eqref{eq.sys} is autonomous (no terms explicitly depend on $s$).
Hence the Hamiltonian is {\em constant} along energy extremal cable shapes, $H(s) \!=\! H^*$ for $s \!\in\! [0,L]$.
Substituting $\Lgs_{\phi}(s) \!=\! - \mbox{\small $EI$} \!\cdot \!u(s)$  according to Eq.~\eqref{eq.partial_u} in 
$H(s)$ and then substituting $u(s)  \!=\! \kp(s)$  gives
\vspace{-.08in}
\begin{equation} \label{eq.Hstar}
\Lgs_r \cdot \big( \cos\phi\mbox{\small $(s)$}\cos\phi_0 + \sin\phi\mbox{\small $(s)$}\sin\phi_0 \big)  -  \hlf \mbox{\small $EI$} \cdot \kp^2(s) = H^* .
\vspace{-.07in}
\end{equation}

\noindent Taking the derivative of both sides with respect to $s$ gives
\vspace{-.07in}
\begin{equation} \label{eq.H_deriv}
\Lg_r \!\cdot\! \big( -\! \sin\phi\mbox{\small $(s)$}\cos\phi_0  + \cos\phi\mbox{\small $(s)$}\sin\phi_0 \big)  -  \mbox{\small $EI$} \cdot \frac{d}{ds} \kp(s) \!=\! 0
\vspace{-.05in}
\end{equation}

\noindent where we canceled the common factor $\kp(s) \!=\! \tfrac{d}{ds} \phi(s)$. Substituting the system equations $\dot{x}(s) \!=\! \cos \phi(s)$
and $\dot{y}(s) \!=\! \sin \phi(s)$ into Eq.~\eqref{eq.Hstar}--\eqref{eq.H_deriv} gives
\vspace{-.05in}
\[
\Lg_r \!\cdot\!  \matwo{\cos\phi_0}{\sin\phi_0 }{\sin\phi_0}{-\cos\phi_0 }
\vctwo{\!\!\dot{x}\mbox{\small $(s)$}\!\!}{\!\!\dot{y}\mbox{\small $(s)$}\!\!} =  \vctwo{\!\!\hlf \mbox{\small $EI$}\!\cdot\! \kp^2(s) + H^* \!\!}
{ \!\!\mbox{\small $EI$} \!\cdot\! \frac{d}{ds} \kp(s) \!\!}.
\vspace{-.04in}
\]

\begin{figure}
\vspace{-.05in}
\centering \includegraphics[width=.435\textwidth]{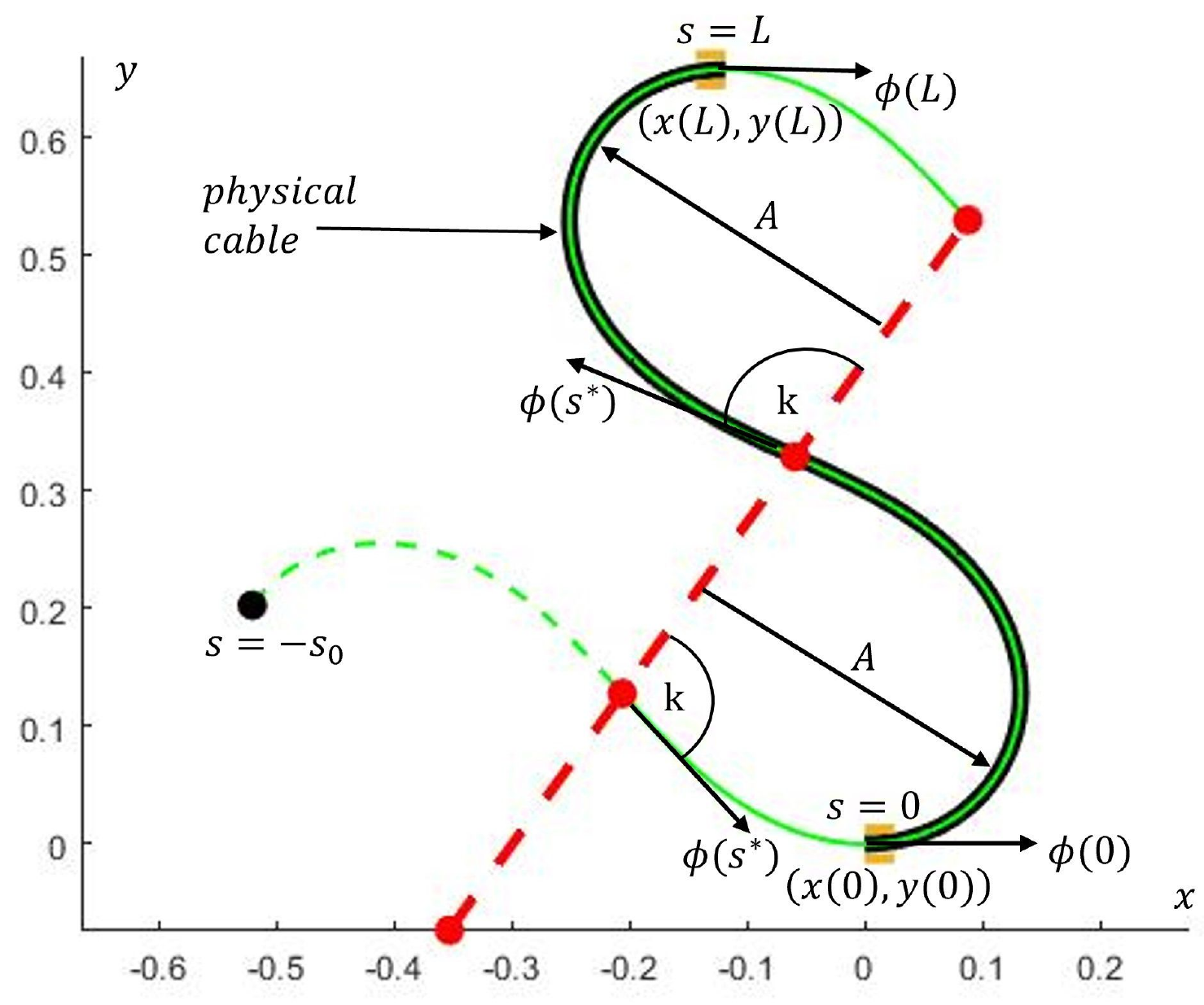}
\vspace{-.18in}
	\caption{The elastica solutions are characterized by the phase~parameter, $s_0$, the amplitude parameter, {\footnotesize $A$}, and the modulus
parameter, $0 \!<\! \mathrm{k}\!<\! 1$, determined by $\phi(s^*)$. Note that $\mbox{\footnotesize $A$} \!=\! 2\mathrm{k} / \mbox{\scriptsize $\sqrt{\Lg}$}$, where $\Lgs$ represents the  magnitude of the force applied at the cable endpoints.}
\label{el_extra.fig}
\vspace{-.2in}
\end{figure}

\noindent Integrating both sides,  $x(s) \!=\! \int_0^s \dot{x}(t) dt$ and $y(s) \!=\! \int_0^s \dot{y}(t) dt$,
gives the cable $(x,y)$ coordinates in terms of its curvature 
\vspace{-.06in}
\begin{equation} \label{eq.xy}
\barr
  &  \vctwo{\!\! x(s) \!\!}{\!\! y(s) \!\!}    \!=\!    \vctwo{\!\! x(0) \!\!}{\!\! y(0) \!\!} \\[6pt]
   &   +   \frac{1}{\Lgs_r} \mbox{\small  $ \matwo{\cos\phi_0}{\sin\phi_0 }{\sin\phi_0}{-\cos\phi_0 }$}
  \vctwo{\!\! \int_0^s \hlf \mbox{\small $EI$}\cdot \kp^2(t) dt + H^* \!\!\cdot\! s \!\!}{\!\! \mbox{\small $EI$} \!\cdot\! ( \kp(s) \!-\!  \kp(\mbox{\small $0$}) ) \!\!}
  \earr
\vspace{-.06in}
\end{equation}

\noindent where $s \!\in\! [0,L]$. When a flexible cable is steered with equal endpoint tangents, its equilibrium shapes possess {\em inflection points,} zero curvature points
at which 
the cable curvature switches sign (Fig.~\ref{el_axis.fig}).
Such cable shapes are called {\em inflectional elastica.}\footnote{Non-inflectional elastica form spiral-like shapes when held with equal endpoint tangents. Such shapes  
may be useful in robotic applications, but the current paper focuses on the more common inflectional elastica.}
The curvature of the inflectional elastica is given by an~elliptic cosine function
of the cable path length parameter~\cite{love}[p. 402-404]
\vspace{-.06in}
\begin{equation} \label{eq.kp}
\kp(s) = - \Lgs  \mbox{\small $A$} \!\cdot\! \mbox{\sf $cn$}\big( \mbox{\small $\sqrt{\Lgs}$} \cdot (s \!+\! s_0),\mathrm{k}\big)   \rtxt{\hspace{1em} $s \in [0,L]$}
\vspace{-.07in}
\end{equation}

\noindent where $\mbox{\sf $cn$}(\cdot)$ has an~{\em ellipse modulus} $0 \!<\! \mathrm{k} \!<\! 1$ discussed below,
$\Lg \!=\! \Lg_r / \mbox{\small $EI$}$, $A$ is an~{\em amplitude parameter} discussed below, and $s_0$ is  a~{\em phase  parameter} that is used to locate the cable start point  (Fig.~\ref{el_extra.fig}).
Verification that $\kp(s)$ satisfies the energy extremal conditions of Eqs.~\eqref{eq:AdjointSystem}-\eqref{eq.u_cond}
appears in the appendix.

The elliptic cosine function 
is a~periodic cosine-like function having two zeroes per period~\cite{ellip_book}. The {\em elastica axis} shown in Fig.~\ref{el_axis.fig} is defined as the line that passes through the elastica inflection points. Some intuition on this axis can be gained by the following observations.
In Eq.~\eqref{eq.xy}, the term $\mbox{\small $EI$} \!\cdot\! ( \kp\mbox{\small $(s)$} \!-\!  \kp(\mbox{\small $0$}) )$ multiplies the column vector
$(\sin \phi_0, - \! \cos \phi_0)$. Since $\kp(s)$ is a~periodic cosine-like function, $\phi_0$ is the elastica axis angle.
It also follows from Eq.~\eqref{eq.xy} that the elastica {\em height} above its axis 
is $h(s) \!=\! \mbox{\small $A$} \!\cdot\!  \mbox{\sf $cn$}( \mbox{\small $\sqrt{\Lgs}$} \!\cdot\! (s \!+\! s_0),\mathrm{k})$ (Fig.~\ref{el_extra.fig}),
which explains the amplitude parameter~$A$.

The geometric meaning of the modulus parameter, $\mathrm{k}$, is based on
the cable tangent, $\phi(s)$, which is
given in terms of an~elliptic sine function, $\mbox{\sf $sn$}(\cdot)$ (see appendix for verification)
\vspace{-.06in}
\begin{equation} \label{eq.phi}
\sin\big( \tfrac{1}{2}(\phi(s) - \phi_0) \big) \!=\! -\mathrm{k}\cdot \mbox{\sf $sn$}\big( \mbox{\small $\sqrt{\Lg}$} \cdot (s +  s_0),\mathrm{k}\big)  \rtxt{$ s \in [0,L]$.}
\vspace{-.07in}
\end{equation}

\noindent The elliptic sine function  
is a~periodic sine-like function satisfying  $\mbox{\sf $cn$}^2(u) \!+\! \mbox{\sf $sn$}^2(u) \!=\! 1$.
It follows from~\mbox{Eqs.~\eqref{eq.kp}-\eqref{eq.phi}} that $\mbox{\sf $sn$}( \mbox{\scriptsize $\sqrt{\Lg}$} \!\cdot\! \mbox{\small $(s \!+\! s_0)$},\mathrm{k}) \!=\! \pm 1$ at the zero curvature points. Substituting this value in Eq.~\eqref{eq.phi} gives
\vspace{-.08in}
\begin{equation} \label{eq.k}
k = \absvl{ \sin\big( \tfrac{1}{2}( \phi(s^*) - \phi_0 ) \big) } \rtxt{\hspace{1em} $\kp(s^*)\!=\!0$.}
\vspace{-.08in}
\end{equation}

\noindent The modulus parameter, $0 \!<\!\mathrm{k}\!<\! 1$,  thus represents the flexible cable {\em incidence angle} 
with its elastica axis (Fig.~\ref{el_extra.fig}). This paper will also use an equivalent parameter that
captures the {\em sign} of the incidence angle, $-1 \!<\! \Sg \!<\! 1$, defined as
\vspace{-.09in}
\[
\Sg =  \cos\big( \phi(s^*) - \phi_0 \big)  \rtxt{\hspace{1em} $\kp(s^*)\!=\!0$,}
\vspace{-.09in}
\]
\noindent where $\mathrm{k}^2 \!=\! (1 \!-\! \Sg)/2$.

{\bf Determination of~flexible~cable~shape~from~endpoint constraints:} Let us formulate three equations that capture
the flexible cable equilibrium shapes in terms of the endpoint constraints,
\mbox{\small $S(0) \!=\! (x(0),y(0),\phi(0))$} and \mbox{\small $S(\mbox{\small $L$}) \!=\! (x(\mbox{\small $L$}),y(\mbox{\small $L$}),\phi(\mbox{\small $L$}))$}. The flexible cable equilibrium shapes will be characterized by three parameters:
the co-state $\Lg_r$ or equivalently $\Lg \!=\! \Lg_r / \mbox{\small $EI$}$, the elastica modulus parameter~$\mathrm{k}$ and the phase parameter~$s_0$.

First consider the parameter $H^*$  in Eq.~\eqref{eq.xy}.~Since $H(s) \!=\! H^*$ for $s \!\in\! [0,\mbox{\small $L$}]$,
the value of  $H^*$  can be determined at any point along  the cable.  At the zero curvature points, $\kp(s^*) \!=\! 0$, Eq.~\eqref{eq.Hstar} gives
 \vspace{-.12in}
 \[
 H^*  \!=\! \Lg_r \cdot ( \cos\phi\mbox{\small $(s^*)$}\cos\phi_0 \!+\! \sin\phi\mbox{\small $(s^*)$}\sin\phi_0 )
 \!=\! \Lg_r  \Sg .
 \vspace{-.08in}
 \]
\noindent Next consider the amplitude~$A$ in Eq.~\eqref{eq.kp}.
 The elastica highest point with respect to its axis occurs at $s \!=\! -s_0$, where
$s \!+\! s_0 \!=\! 0$ and $\mbox{\sf $cn$}( 0,\mathrm{k}) \!=\! 1$ (Fig.~\ref{el_extra.fig}).
Substituting $H^* \!=\! \Lg_r \Sg$ in Eq.~\eqref{eq.Hstar},  then evaluating $H(s)$ at 
$s \!=\! -s_0$ using the relations $\Lg \!=\! \Lg_r / \mbox{\small $EI$}$ and $\mathrm{k}^2 \!=\! (1 \!-\! \Sg)/2$ gives the amplitude
\vspace{-.13in}
\begin{equation} \label{eq.ms_0}
\Lg_r  -  \hlf \mbox{\small $EI$} \cdot (\Lg  \mbox{\small $A$})^2
\!=\! \Lg_r  \Sg  \,\, \Rightarrow \,\,
\mbox{\small $A$} \!=\! \frac{2\mathrm{k}}{\mbox{\small $\sqrt{\Lgs}$}} \, .
\vspace{-.08in}
\end{equation}

\noindent We can now formulate three equations for $\Lg$, $\mathrm{k}$ and $s_0$ in terms of the cable endpoint constraints.
 The cable $(x,y)$ coordinates, Eq.~\eqref{eq.xy}, evaluated at $s \!=\! \mbox{\small $L$}$  give two equations
\vspace{-.05in}
\begin{equation} \label{eq.xy_L}
\barr
  &  \vctwo{\!\! x(\mbox{\small $L$}) \!\!}{\!\! y(\mbox{\small $L$}) \!\!}    \!-\!    \vctwo{\!\! x\mbox{\small $(0)$} \!\!}{\!\! y\mbox{\small $(0)$} \!\!}  \\[6pt]
   & =  \frac{1}{\Lgs_r} \mbox{\small  $ \matwo{\cos\phi_0}{\sin\phi_0 }{\sin\phi_0}{-\cos\phi_0 }$}
  \vctwo{\int_0^L \hlf \mbox{\small $EI$}\cdot \kp^2(t) dt + (\Lgs_r \Sg ) \cdot L }{\mbox{\small $EI$} \!\cdot\! ( \kp(\mbox{\small $L$})
   \!-\!  \kp(\mbox{\small $0$}) )} \\[6pt]
&  = \frac{1}{\Lgs} \mbox{\small  $ \matwo{\cos\phi_0}{\sin\phi_0 }{\sin\phi_0}{-\cos\phi_0 }$}
  \vctwo{\int_0^L \hlf \kp^2(t) dt + (\Lgs \Sg ) \cdot L }{ \kp(\mbox{\small $L$})
   \!-\!  \kp(\mbox{\small $0$}) } \earr
\vspace{-.02in}
\end{equation}
\noindent where we substituted $H^*   \!=\! \Lg_r  \Sg$ and $\Lg \!=\! \Lg_r / \mbox{\small $EI$}$.
The endpoint curvatures in Eq.~\eqref{eq.xy_L} are
$\kp(0) \!=\! -  2\mathrm{k}\mbox{\small $\sqrt{\Lg}$}   \cdot  \mbox{\sf $cn$}( \mbox{\small $\sqrt{\Lg}$} \cdot s_0,\mathrm{k})$
and $\kp(\mbox{\small $L$}) \!=\! -  2\mathrm{k}\mbox{\small $\sqrt{\Lg}$}    \! \cdot\!  \mbox{\sf $cn$}( \mbox{\small $\sqrt{\Lg}$} \!\cdot \! (s_0 \!+\! \mbox{\small $L$}),\mathrm{k})$, based on Eq.~\eqref{eq.kp}. The elastica axis direction, $\phi_0$,
is obtained by evaluating Eq.~\eqref{eq.phi} for $\phi(s)$ at $s \!=\! 0$
\vspace{-.12in}
\begin{equation} \label{eq.phi0}
\phi_0 = \phi(\mbox{\small $0$}) + 2\sin^{-1}\big(\mathrm{k}\cdot \mbox{\sf $sn$}( \mbox{\small $\sqrt{\Lg}$} \!\cdot\! s_0, \mathrm{k}) \big) .
\vspace{-.08in}
\end{equation}
\noindent The third equation is obtained by evaluating Eq.~\eqref{eq.phi} 
at the cable endpoints
$s \!=\! 0$ and $s \!=\! \mbox{\small $L$}$
\vspace{-.06in}
\begin{equation} \label{eq.phi_0L}
\barr
& \tfrac{1}{2} \big( \phi(\mbox{\small $L$}) - \phi(0) \big) = \\[6pt]
&  \sin^{-1}\big(\mathrm{k}\!\cdot\! \mbox{\sf $sn$}( \mbox{\small $\sqrt{\Lgs}$} \!\cdot\! s_0,\mathrm{k})\big)
- \sin^{-1}\big(\mathrm{k}\!\cdot\! \mbox{\sf $sn$}( \mbox{\small $\sqrt{\Lgs}$} \!\cdot\! (s_0 \!+\! L), \mathrm{k}) \big).
\earr
\vspace{-.02in}
\end{equation}

\noindent Equations \eqref{eq.xy_L}-\eqref{eq.phi_0L} provide three equations~in~$\Lg$,~$\mathrm{k}$~and~$s_0$ whose solution
by numerical means gives~the~\mbox{flexible}~\mbox{cable} shape as a~function of its relative endpoint states, \mbox{\small $S(L) \!-\!  S(0)$}.
The cable  {\em base frame configuration} is defined as $(x(0),y(0),\phi(0))$, and together with \mbox{\small $S(L) \!-\! S(0)$} gives a~total of six configuration variables. This paper describes a~steering scheme that maintains equal endpoint tangets, thus giving a~total of five configuration variables.

{\bf The elastica full period length:}
The elliptic cosine and sine functions period is $4 K(\mathrm{k})$, where $K(\mathrm{k})$ is the complete elliptic integral of
the first kind.\footnote{It is defined as $K(\mathrm{k}) \!=\! \int_0^{\pi/2} \!\! \tfrac{d \theta}{\sqrt{1 - \mathrm{k}^2 \sin^2 \theta}}$,
a~standard  analytic function of the modulus parameter $\mathrm{k}$~\cite{ellip_book}.}
Denote by $\tilde{L}$ the {\em full period length} of the elastica
that contains the physical cable.  The argument $\mbox{\small $\sqrt{\Lg}$} \cdot (s \!+\!  s_0)$ in Eqs.~\eqref{eq.kp}-\eqref{eq.phi} satisfies the full-period relation $\mbox{\small $\sqrt{\Lg}$} \cdot \tilde{L}  \!=\! 4 K(\mathrm{k})$, which gives  $\tilde{L}  \!=\! 4 K(\mathrm{k}) / \mbox{\small $\sqrt{\Lg}$}$.

{\bf The {\small 2-D} cable steering problem:} A flexible cable
modeled by Euler's elastica
is held by two robot grippers in a~planar environment populated by sparsely spaced
obstacles. Find a~path for the robot grippers that steers the flexible cable from start to target endpoint states using equal endpoint tangents, such that the cable maintains stable equilibrium shapes while  avoiding self-intersection and contact with the environment, except  for endpoint contacts at the start or target positions.\\
\indent The sparse obstacles assumption highlights~a~caveat~in~the cable steering problem.
Two robot grippers are able to control the flexible cable shape only within the elastica family of solutions. Obstacle avoidance is therefore feasible
only among sparsely spaced obstacles
(see Section~\ref{sec.examples}).

\vspace{-.04in}

\section{\bf Non Self-Intersecting Flexible Cable Shapes} \label{sec.self_intersect}
\vspace{-.02in}

\noindent This section uses the elastica solutions to identify the non self-intersecting flexible cable shapes, which
remarkably
involve only the modulus parameter~$\mathrm{k}$.

A flexible cable  can possibly self-intersect when its tangent angle with respect to its elastica axis, $\phi(s) \!-\! \phi_0$, exceeds $\pm 90^\circ$  for some $s \!\in \! [0,L]$ (Fig.~\ref{el_extra.fig}). Using Eq.~\eqref{eq.phi} for $\phi(s)$,
the flexible cable can self-intersect only when $\mbox{\small $\absvl{\sin(\tfrac{1}{2}(\phi\mbox{\small $(s)$} \!-\! \phi_0))}$} \!\geq\! \tfrac{1}{\sqrt{2}}$ for some $s \!\in \! [0,L]$.  When this happens,~the~cable~crosses~its~elas\-tica axis at an~angle $\mbox{\small $\absvl{\sin(\tfrac{1}{2}(\phi\mbox{\small $(s^*)$} - \phi_0))}$} \!\geq\! \tfrac{1}{\sqrt{2}}$ where~\mbox{$\kp(s^*) \!=\! 0$.}
Hence, according to Eq.~\eqref{eq.k} for the modulus parameter, $0 \!<\!\mathrm{k}\!<\! 1$,
self-intersection can possibly occur when
\vspace{-.07in}
\[
\mathrm{k} = \absvl{ \sin\big( \tfrac{1}{2}( \phi(s^*) - \phi_0 ) \big) } \geq \mbox{\small $\frac{1}{\sqrt{2}}$} \rtxt{\hspace{1em} $\kp(s^*)\!=\!0$.}
\vspace{-.05in}
\]

\noindent Next consider the projection of the flexible cable  $(x(s),y(s))$ coordinates on its elastica axis, denoted $\xb(s)$. 
A full period of the elastica that contains the physical cable starts at $s \!+\! s_0 \!=\! 0$ and ends at
$s \!+\! s_0 \!=\! \tilde{L}$, where $\tilde{L}$ is the full-period~length~of~the elastica that contains the physical cable (Fig.~\ref{el_axis.fig}).
Define the equivalent path length parameter $\bar{s} \!=\! s \!+\! s_0$.
According to Eq.~\eqref{eq.xy}
\vspace{-.12in}
\begin{equation} \label{eq.xb}
\xb(\bar{s}) =  \frac{\mbox{\small $1$}}{\Lgs_r} \!\cdot \! \int_0^{\bar{s}} \!\! \hlf \mbox{\small $EI$}\cdot \kp^2(t) dt + \frac{\mbox{\small $1$}}{\Lgs_r} H^* \!\cdot\! \bar{s}  \rtxt{\hspace{.5em} $\bar{s} \in [0, \tilde{L}]$.}
\vspace{-.07in}
\end{equation}

\begin{figure}
\centering \includegraphics[width=.45\textwidth]{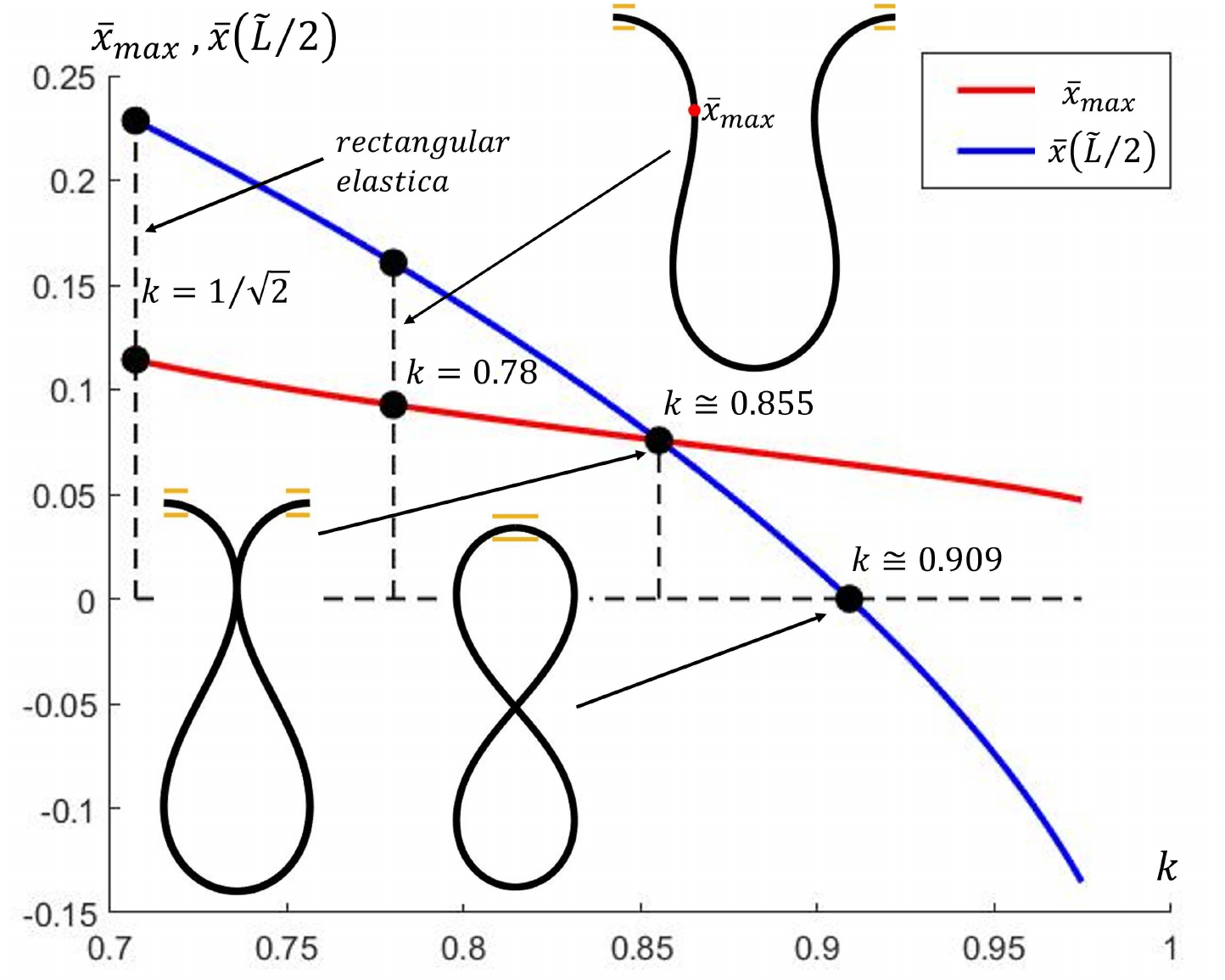}
\vspace{-.2in}
	\caption{Numerical solution of the non self-intersection condition, $\xb_{max} \! \leq \! \xb( \mbox{\scriptsize $\tilde{L}/2$})$
where  $\xb_{max} \!=\! \xb(\bar{s}_{max})$, gives the
modulus parameter $\mathrm{k}_{max} \!=\! 0.855$. The range $0 \!<\!\mathrm{k}\!<\! \mathrm{k}_{max}$ guarantees that the flexible cable is {\em not} self-intersecting.}
\label{intersect.fig}
\vspace{-.2in}
	\end{figure}

\noindent When $\mathrm{k} \!\geq\! \mbox{\small $\tfrac{1}{\sqrt{2}}$}$,~each~full~\mbox{period}~of~the~\mbox{elastica}~\mbox{contains}~two 
{\em fold} {\em  points} located at $\bar{s}_{max}$ and $\tilde{L} \!-\! \bar{s}_{max}$
at which $\mbox{\small $\absvl{\phi(\bar{s}) \!-\! \phi_0}$} \!=\! 90^\circ \!$ (Fig.~\ref{intersect.fig}).
Evaluating this condition
using~Eq.~\eqref{eq.phi} for $\phi(s)$ gives 
\vspace{-.14in}
\[
\mathrm{k}\cdot \mbox{\sf $sn$}\big( \mbox{\small $\sqrt{\Lg}$} \cdot \bar{s}_{max},\mathrm{k}\big) \!=\! \mbox{\small $\frac{1}{\sqrt{2}}$}
  \,\Rightarrow\,
  \bar{s}_{max}  \!=\!  \mbox{\small $\frac{1}{\mbox{\scriptsize $\sqrt{\Lg}$} }$} \mbox{\sf $sn$}^{-1}\big( \tfrac{1}{\sqrt{2} \cdot \mathrm{k}},\mathrm{k}\big).
\vspace{-.08in}
\]
where $\mbox{\sf $sn$}^{-1}(\cdot)$ is the inverse elliptic sine function.
As long as  $\xb(\bar{s}_{max} ) \leq \xb( \tilde{L}/2)$, the fold points
do not touch each other (Fig.~\ref{intersect.fig}). Hence, using Eq.~\eqref{eq.xb} for $\xb(\bar{s})$, the flexible cable is {\em not} self-intersecting when
\vspace{-.1in}
\begin{equation} \label{eq.non_int}
  \int_0^{\bar{s}_{max}}\!\! \hlf \mbox{\small $EI$}\cdot \kp^2(t) dt +  \mbox{\small $H^*$} \!\cdot\! \bar{s}_{max}
 \leq
\int_0^{\tilde{L}/2} \! \hlf \mbox{\small $EI$}\cdot \kp^2(t) dt + \mbox{\small $H^*$} \!\cdot\! \mbox{\small $\frac{\tilde{L}}{2}$}
\vspace{-.02in}
\end{equation}

\noindent where we canceled the common factor $1 / \Lgs_r$.
Substituting $\mbox{\small $H$}^* \!=\! \Lg_r \!\cdot\! (1 \!-\! 2 \mathrm{k}^2) $ and $\kp(t) \!=\! - 2 \mbox{\small $\sqrt{\Lg}$}\mathrm{k}\!\cdot \! \mbox{\sf $cn$}( \mbox{\small $\sqrt{\Lg}$} t,\mathrm{k})$ in Eq.~\eqref{eq.non_int} gives the non self-intersection condition
\vspace{-.03in}
\[
\barr
  \int_0^{\bar{s}_{max}} \mbox{\small $2$} \mathrm{k}^2 \hspace{.08em}  \mbox{\sf $cn$}^2( \mbox{\small $\sqrt{\Lg}$} t,\mathrm{k}) dt +  (1 \!-\! 2 \mathrm{k}^2)  \!\cdot\! \bar{s}_{max} & \\[4pt]
 \leq
\int_0^{\tilde{L}/2} \mbox{\small $2$}   \mathrm{k}^2\hspace{.08em} \mbox{\sf $cn$}^2( \mbox{\small $\sqrt{\Lg}$} t,\mathrm{k}) dt +  (1 \!-\! 2 \mathrm{k}^2) \!\cdot\! \frac{\tilde{L}}{2}
\earr
\vspace{-.05in}
  \]

\noindent where we canceled the common factor $\Lgs_r \!=\! \mbox{\small $EI$} \!\cdot\! \Lgs$.
A~\mbox{change}~of integration variable, $u \!=\! \mbox{\small $\sqrt{\Lgs}$} \cdot t $ with $du \!=\! \mbox{\small $\sqrt{\Lgs}$} \cdot dt$,
gives an equivalent non self-intersection condition that~\mbox{depends}~\mbox{only}~on~$\mathrm{k}$
\vspace{-.08in}
\begin{equation} \label{eq.self}
\barr
 \int_0^{u_{max}(\mathrm{k})}  \mbox{\small $2$} \mathrm{k}^2 \hspace{.08em} \mbox{\sf $cn$}^2( u,\mathrm{k}) du +   (1 \!-\! 2 \mathrm{k}^2)  \!\cdot\! u_{max}(\mathrm{k}) & \\[4pt]
 \leq
\int_0^{2K(\mathrm{k})}  \mbox{\small $2$} \mathrm{k}^2 \hspace{.08em} \mbox{\sf $cn$}^2( u,\mathrm{k}) du  +  (1 \!-\! 2 \mathrm{k}^2) \!\cdot\!  2K(\mathrm{k})
\earr
\vspace{-.08in}
\end{equation}

\noindent with integration limits $u_{max}(\mathrm{k}) \!=\! \mbox{\small $\sqrt{\Lgs}$} \!\cdot\!   \bar{s}_{max}\!=\!  \mbox{\sf $sn$}^{-1}\big( \tfrac{1}{\sqrt{2} \cdot \mathrm{k}},\mathrm{k} \big)$ and $u \!=\! \mbox{\small $\sqrt{\Lgs}$} \cdot \tfrac{1}{2} \tilde{L} \!=\! 2K(\mathrm{k})$ since \mbox{\small $\tilde{L} \!=\! 4 K(\mathrm{k}) / \mbox{\small $\sqrt{\Lg}$}$}.~The~two~sides of Eq.~\eqref{eq.self} depend only on
$\mathrm{k}$ and intersect at $\mathrm{k}_{max} \!=\! 0.855$ (Fig. \ref{intersect.fig}). The range $0 \!<\!\mathrm{k}\!<\! \mathrm{k}_{max}$
guarantees that the flexible cable is  {\em not} self-intersecting, irrespective
of the other elastica parameters.

{\bf Remark:} The range $0 \!<\! \mathrm{k} \!<\! 0.855$ provides
a~conservative rule for non self-intersection.
Elastica solutions with modulus parameter in the range $0.855 \! \leq \! \mathrm{k} \!<\! 1$ are self-intersecting,
but the physical cable may occupy a~short segment of such elastica that is not self-intersecting.~\eex
\vspace{-.04in}

\section{\bf Flexible Cable Stability } \label{sec.stability}
\vspace{-.04in}

\noindent This section uses the elastica solutions to obtain a simple geometric characterization of the stable cable shapes to be used during steering.
Recall that the elastica inflection points are zero curvature points (Fig.~\ref{el_axis.fig}).
The following theorem adapted from Sachkov~\cite{sachkov07}
characterizes the stable cable shapes in terms of the number of inflection points.

\begin{figure}
\centering \includegraphics[width=.5\textwidth]{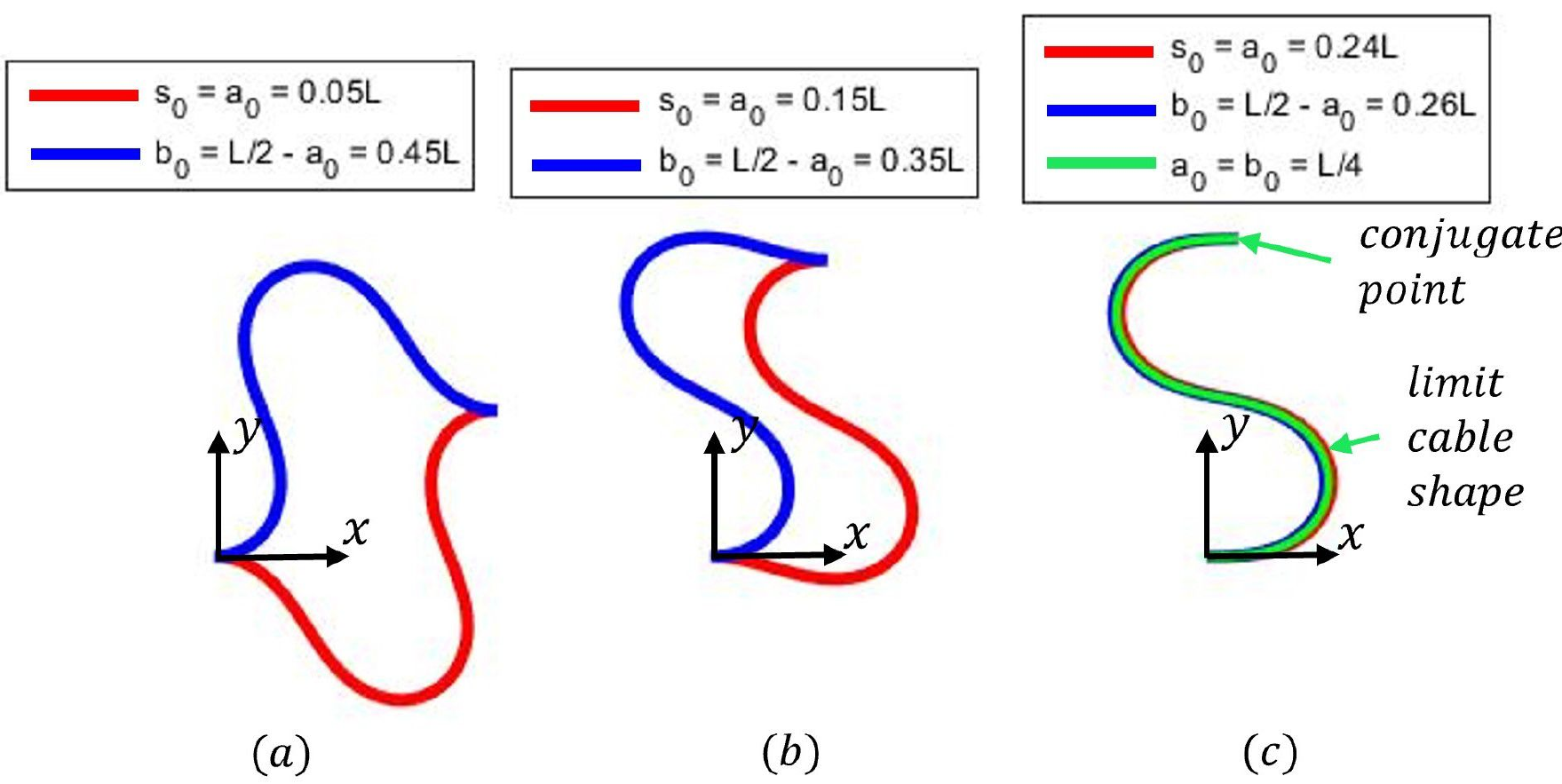}
\vspace{-.28in}
	\caption{(a)-(b) Two families of energy extremal cable shapes, $(x_1(s),y_1(s))$~deter\-mined by $s_0 \!=\! a_0$ and
$(x_2(s),y_2(s))$ determined by $s_0 \!=\! b_0$. All cable shapes have same length of $\mbox{\scriptsize $L$} \!=\! 1$. (c)~When the two families of extremal cable shapes approach the limit $a_0 \!=\! b_0 \!=\! \tfrac{\tilde{L}}{4}$, their common endpoint reaches
a~conjugate point along the limit cable shape.}
\label{conj_cable.fig}
\vspace{-.15in}
\end{figure}

\indent {\bf Theorem 1 [Stability]:}
{\em Let a flexible cable be held with fixed endpoint positions and tangents
at an~equilibrium state. Cable shapes that contain zero, one or two inflection points are {\bf possibly stable} while cable shapes
that contain three or more inflection points are {\bf definitely unstable.}
}

{\bf Proof sketch:} Stable cable shapes form local minima
of their total elastic energy, $\Es \!=\!  \int_0^L \! \hlf \mbox{\small $ EI$} \!\cdot\! \kp^2(s) ds$.
First consider cables whose length is much shorter than the full period length of the elastica containing the physical cable, $L \!\ll\! \tilde{L}$.
Equilibrium shapes of these cables 
form local minima of $\Es$,
since $\Es$  possesses a~positive definite second variation at the equilibrium state.
As $L$ increases relative to $\tilde{L}$, the second variation of $\Es$ eventually ceases to be positive definite at a~{\em  conjugate point}~\cite{gelfand}[Theorem 26.3].

In order to describe the notion of a~conjugate point, consider a~flexible cable held at a~fixed initial point and fixed tangent at this point.
The cable equilibrium shape is fully determined as the solution of the cable system equations (Eq.~\eqref{eq.sys})  and the cable adjoint  equations (Eq.~\eqref{eq:AdjointSystem}) for each initial value of the co-states $(\Lg_x,\Lg_y)$ and $\Lg_{\phi}(\mbox{\small $0$})$.
In Section~\ref{sec.elastica}, the co-states were replaced by the co-state
parameter $\Lgs$, the elastica modulus~$\mathrm{k}$ and the phase parameter~$s_0$.

One technique to identify a~conjugate point (which marks loss of stability) is described in ~\cite{gelfand}[Definition 27.4]. Consider two families of energy extremal cable shapes sharing a~fixed initial point and fixed tangent at this point, parameterized by $\Lgs$, $\mathrm{k}$ and $s_0$. If the two distinct families merge into a~single extremal cable shape at the limit $(\Lgs,\mathrm{k},s_0) \rightarrow (\Lgs^*,\mathrm{k}^*,s^*_0)$ while intersecting at their respective endpoints,
the {\em endpoint} of the limit cable shape associated with $(\Lgs^*,\mathrm{k}^*,s^*_0)$ forms a~{\em conjugate point} along the limit cable shape. 

Using this insight, consider two families of energy extremal cable shapes having the same  modulus parameter, $\mathrm{k} \!=\! \mathrm{k}^*$, the same co-state parameter, $\Lgs \!=\! \Lgs^*$, and variable phase parameter~$s_0$.
From Section~\ref{sec.elastica}, the full period  length of the elastica that contains the physical cable is given by  $\tilde{L}  \!=\! 4 K(\mathrm{k}) / \mbox{\small $\sqrt{\Lg}$}$. Hence all cable shapes of both families have full period  length
of $\tilde{L}  \!=\! 4 K(\mathrm{k}^*) / \mbox{\small $\sqrt{\Lg^*}$}$.

Consider the family of energy extremal cable shapes given by Eq.~\eqref{eq.xy}, $(x_1 \mbox{\small $(s)$},y_1 \mbox{\small $(s)$})$, whose phase parameter is $s_0 \!=\! a_0$
\vspace{-.05in}
\[
\mbox{\small $  \vctwo{\!\! x_1(s) \!\!}{\!\! y_1(s) \!\!}    \!=\!    \vctwo{\!\! x(0) \!\!}{\!\! y(0) \!\!}
     \!+\!   \frac{1}{\Lgs^*} \mbox{\small  $ \matwo{\cos\phi_0}{\sin\phi_0 }{\sin\phi_0}{-\cos\phi_0 }$} \!
  \vctwo{\!\! \int_0^s \hlf  \kp^2(t) dt +  \Sg \!\cdot\! s \!\!}{\!\! \kp_1(s) \!-\!  \kp_1(\mbox{\small $0$})  \!\!}
  $}
\]
\vspace{-.15in}

\noindent where $s \!\in\! [0,\mbox{\small $\tilde{L}$}]$, $\Sg  \!=\! 1 \!-\! 2(\mathrm{k}^*)^2$
and $\kp_1(s) = - \Lgs^*  \mbox{\small $A$} \!\cdot\! \mbox{\sf $cn$}( \mbox{\small $\sqrt{\Lgs^*}$} \cdot (s \!+\! a_0), \mathrm{k}^* )$ (using Eq.~\eqref{eq.kp}), where $\mbox{\small $A$} \!=\! 2\mathrm{k}^* / \mbox{\small $\sqrt{\Lgs^*}$}$.

Next consider an~alternative family of energy extremal cable shapes, $(x_2 \mbox{\small $(s)$},y_2 \mbox{\small $(s)$})$, whose  phase parameter $s_0 \!=\! b_0$ is set symmetrically with respect to $a_0$ about the inflection points:
if $a_0$ lies in the interval $[0,\tfrac{\tilde{L}}{2}]$ then $b_0 \!=\! \tfrac{\tilde{L}}{2} \!-\! a_0$; if $a_0$ lies in the complimentary interval
$(\tfrac{\tilde{L}}{2}, \tilde{L}]$ then $b_0 \!=\!  \tfrac{3\tilde{L}}{2} \!-\! a_0$. Thus
\vspace{-.06in}
\[
\mbox{\small $  \vctwo{\!\! x_2(s) \!\!}{\!\! y_2(s) \!\!}    \!=\!    \vctwo{\!\! x(0) \!\!}{\!\! y(0) \!\!}
     \!+\!   \frac{1}{\Lgs^*} \mbox{\small  $ \matwo{\cos\phi_0}{\sin\phi_0 }{\sin\phi_0}{-\cos\phi_0 }$} \!
  \vctwo{\!\! \int_0^s \hlf  \kp^2(t) dt +  \Sg \!\cdot\! s \!\!}{\!\!  \kp_2(s) \!-\!  \kp_2(\mbox{\small $0$})  \!\!}
  $}
\vspace{-.09in}
\]

\noindent where $s \!\in\! [0,\mbox{\small $\tilde{L}$}]$, $\Sg$ is the same as above and
$\kp_2(s) = - \Lgs^*  \mbox{\small $A$} \!\cdot\! \mbox{\sf $cn$}( \mbox{\small $\sqrt{\Lgs^*}$} \cdot (s \!+\! b_0), \mathrm{k}^* )$ with
$A$ the same as above.

The two families of cable shapes start at $(x\mbox{\small $(0)$},y\mbox{\small $(0)$})$. One can also verify that  the two families start along the same tangent direction, $\phi_1(0) \!=\! \phi_2(0)$, for each matched choice of $a_0$ and $b_0$ (Fig.~\ref{conj_cable.fig}(a)-(b)). One can also verify that the two families of
cable shapes
intersect at their respective endpoints, $(x_1(\mbox{\small $\tilde{L}$}),y_1(\mbox{\small $\tilde{L}$})) \!=\! (x_2(\mbox{\small $\tilde{L}$}),y_2(\mbox{\small $\tilde{L}$}))$, for each matched choice of $a_0$ and $b_0$ (Fig.~\ref{conj_cable.fig}(a)-(b)).

The key step of the proof is to study the limit of the two families 
when $a_0 \!=\! b_0$.  When $a_0 \!\in\! [0,\tfrac{\tilde{L}}{2}]$, this limit occurs when $a_0 \!\rightarrow\! \tfrac{\tilde{L}}{4}$ and then $b_0 \!=\!  \tfrac{\tilde{L}}{2} \!-\! a_0 \!\rightarrow\! \tfrac{\tilde{L}}{4}$ (the elastica 1'st inflection point). When $a_0 \!\in\! (\tfrac{\tilde{L}}{2}, \tilde{L}]$, the limit occurs when $a_0 \!\rightarrow\! \tfrac{3\tilde{L}}{4}$ and then $b_0 \!=\!  \tfrac{3\tilde{L}}{2} \!-\! a_0 \!\rightarrow\! \tfrac{3\tilde{L}}{4}$ (the elastica 2'nd inflection point). In either case,
the limit cable shape starts at an~inflection point, passes through a~midpoint inflection point and
ends after a full period 
at a~third inflection point (Fig.~\ref{conj_cable.fig}(c)).  Since the two families of extremal cable shapes intersect at their respective endpoints, the endpoint of the limit cable shape at the third inflection point forms a~{\em conjugate point} along~the~limit~cable~shape.~\epf


Theorem~1 is next applied
to flexible cables held with equal  endpoint tangents. In this case  the stable cable shapes consist of two types
($L$ is the physical cable length, $\tilde{L}$  the full period length of the elastica that contains the physical cable).

\indent {\bf Corollary 4,1:}
{\em Under the conditions of Theorem~1, when a~flexible cable is held with equal endpoint tangents, the {\em stable cable shapes} either
contain {\bf a single inflection point} at the cable midpoint and then $L \!<\! \tilde{L}$, or  contain {\bf two
interior inflection points} and then $L \!=\! \tilde{L}$.
}

\begin{figure}
\centering \includegraphics[width=.465\textwidth]{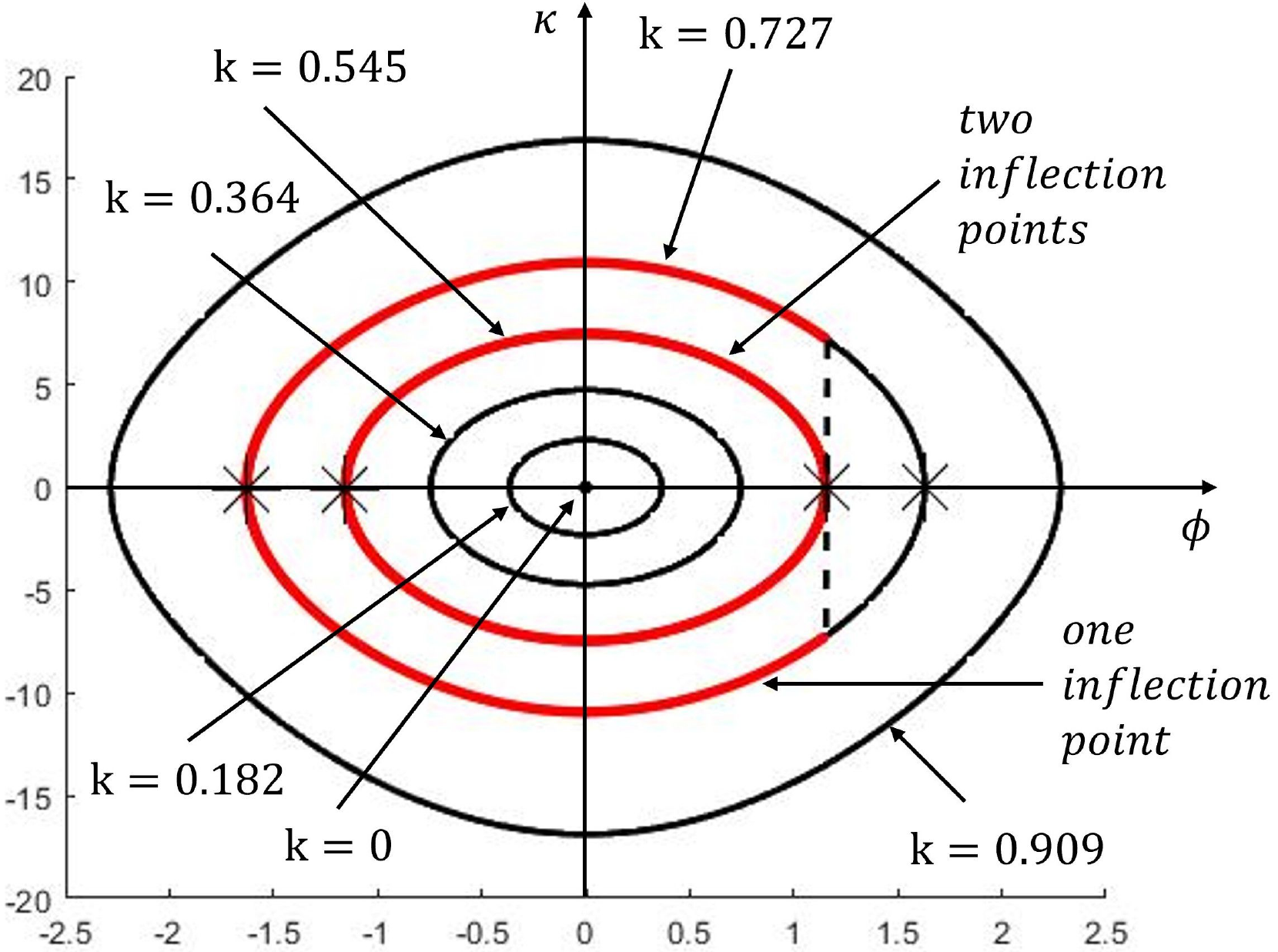}
\vspace{-.15in}
\caption{The $(\phi(s),\kp(s))$ contours of \mbox{\scriptsize $H(s) \!=\! H^*$} form convex loops that describe full periods of the elastica. When a~flexible cable is held with equal endpoint tangents, the cable occupies a~segment with vertically aligned endpoints on a particular loop, with one or two inflection points.}
\label{ovals.fig}
\vspace{-.15in}
\end{figure}

{\bf Proof:} The possibly stable cable shapes contain zero, one or two inflection points according to Theorem~1. From Section~\ref{sec.elastica},
the Hamiltonian is constant along energy extremal cable shapes,~\mbox{\small $H(s) \!=\! H^*$}~for~\mbox{$s \!\in\! [0,L]$.}~\mbox{According}~to~Eq.~\eqref{eq.Hstar}~for~$H(s)$
\vspace{-.12in}
\begin{equation} \label{eq.ovals}
\Lg_r \cdot \cos(\phi\mbox{\small $(s)$} \!-\! \phi_0 )   \!-\!  \hlf \mbox{\small $EI$} \cdot \kp^2(s) \!=\! \Lg_r  \cdot (1 \!-\! 2\mathrm{k}^2 )
\rtxt{$\!\! s \in [0,L]$}
\vspace{-.07in}
\end{equation}

\noindent where we substituted $H^*  \!=\!  \Lg_r  \!\cdot\! (1 \!-\! 2\mathrm{k}^2 )$.  The~\mbox{flexible}~\mbox{cable}~tan\-gent and curvature, $(\phi\mbox{\small $(s)$},\kp\mbox{\small $(s)$})$, are variables~while~$\Lgs_r$,~\mbox{\small $EI$}~and $\mathrm{k}$ are fixed parameters. 
The $(\phi\mbox{\small $(s)$},\kp\mbox{\small $(s)$})$ contours described~by Eq.~\eqref{eq.ovals} form convex loops centered at the origin,~shown~in Fig.~\ref{ovals.fig} for different values of the modulus parameter $\mathrm{k}$ (the other parameters do not affect these loops).
Each loop represents full period of the elastica, $s \in [0,\tilde{L}]$.
Since each loop crosses the $\phi$-axis exactly at two points~at~which~$\kp(s) \!=\! 0$,~each full period of the elastica contains exactly
two~\mbox{inflection}~points.


When a~flexible cable is held with equal endpoint tangents, its endpoints lie on the {\em same vertical line} in the $(\phi(s),\kp(s))$ plane
(Fig.~\ref{ovals.fig}). The physical cable therefore occupies one of two segments on a~particular loop, such that the segment is bounded by vertically aligned endpoints. Any such segment
crosses the $\phi$-axis at least once, hence the physical cable must have at least one inflection point. When a~segment crosses the $\phi$-axis at a single point (one inflection point), the segment occupies less than a~full loop and then $L \!<\! \tilde{L}$ (Fig.~\ref{ovals.fig}). When a~segment crosses the $\phi$-axis at two points (two inflection points), its vertically aligned endpoints merge and the segment forms a~full loop with $L \!=\! \tilde{L}$ (Fig.~\ref{ovals.fig}).~\epf

\begin{figure}
\centering \includegraphics[width=.465\textwidth]{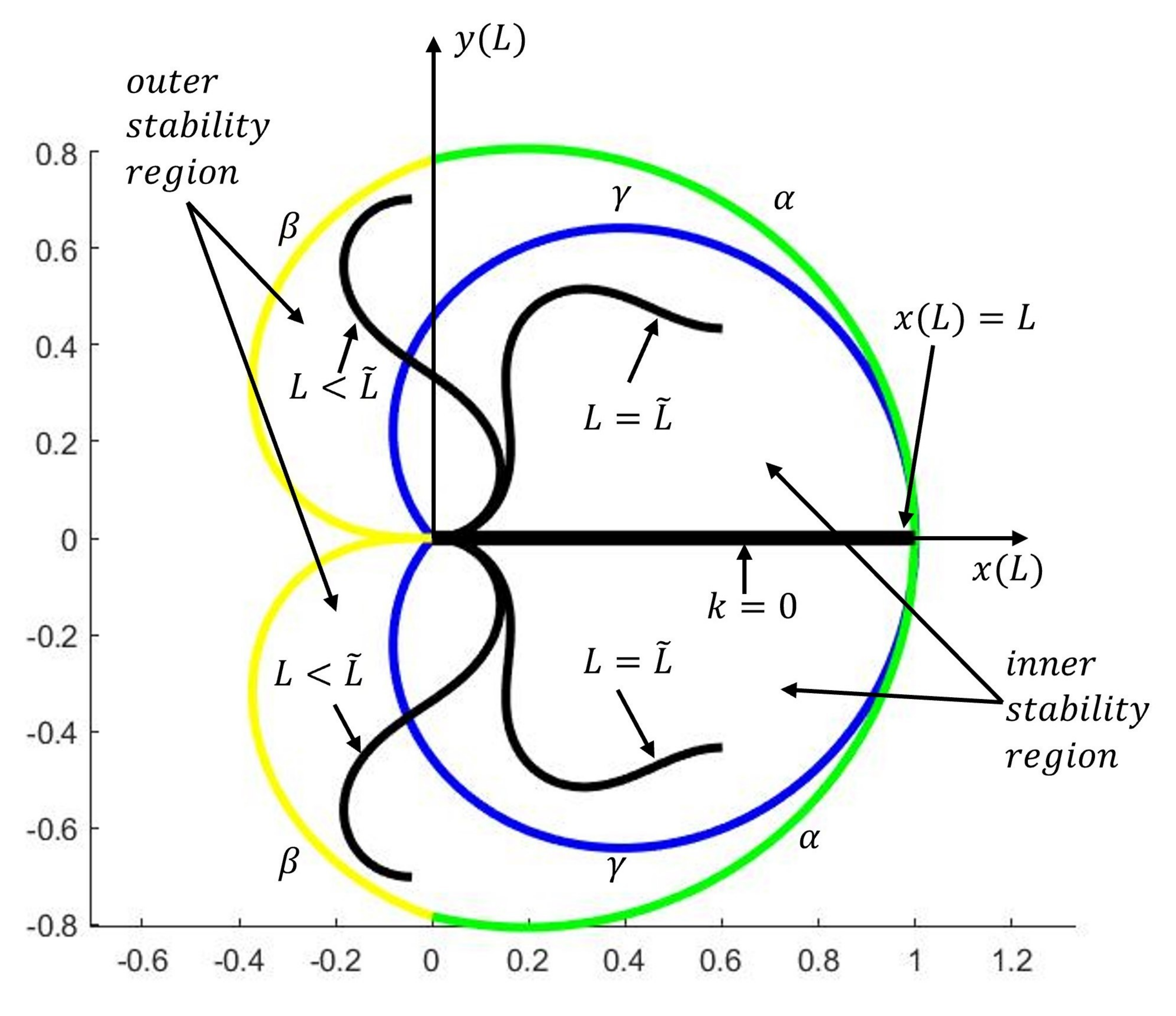}
\vspace{-.2in}
\caption{The cable start point is fixed at the origin with tangent fixed along the $x$-axis. The cable distal endpoint varies while its
tangent is fixed along the $x$-axis.
The outer stability region represents endpoint positions where the stable cable shapes have a~midpoint inflection point and  $L \!<\! \tilde{L}$. The inner stability region represents endpoint positions where the stable cable shapes have two
interior inflection points and $L \!=\! \tilde{L}$.}
\label{endpoint_regions.fig}
\vspace{-.2in}
\end{figure}

{\bf Determination~of~stable~cable~shapes from endpoint positions:} The two types of stable cable shapes
partition the cable relative endpoint positions as follows.
Let the cable start point be fixed at the origin with the cable tangent fixed along the $x$-axis (Fig.~\ref{endpoint_regions.fig}). The cable {\em relative endpoint space} is defined as the position of its distal endpoint, $(x(\mbox{\small $L$}),y(\mbox{\small $L$}))$, with the cable endpoint tangent, $\phi(\mbox{\small $L$})$,  fixed along the $x$-axis
(Fig.~\ref{endpoint_regions.fig}).

From Corollary~1, the length of any stable cable shape is either $L \!<\! \tilde{L}$ (one inflection point) or $L \!=\! \tilde{L}$ (two inflection points).
As $L$ becomes shorter than $\tilde{L}$, the physical cable occupies smaller portion of the full elastica period.
If the modulus parameter is held fixed when  $L$ becomes shorter than~$\tilde{L}$, the cable
flattens while its distal endpoint moves away from the origin. 
To allow the cable endpoint reach points located on 
the vertical $y$-axis,  a~maximal flattening parameter
limits the cable flattening to the range  $\rho \!\cdot\! \tilde{L} \!\leq \!  L \!\leq\! \tilde{L}$, where
$0 \!<\! \rho \!<\! 1$. For instance, $\rho \!=\! 1/2$ is used in Fig.~\ref{endpoint_regions.fig}.

We also need to limit the modulus parameter $\mathrm{k}$ as follows. At $\mathrm{k} \!=\! 0$ the cable
forms a~horizontal straight line with \mbox{\small $(x(\mbox{\small $L$}),y(\mbox{\small $L$})) \!=\! (0,\mbox{\small $L$})$} (Fig.~\ref{endpoint_regions.fig}). As
$\mathrm{k}$ increases, the flexible cable 
folds inward towards the origin. Eventually $\mathrm{k}$ reaches a~critical value 
at which the cable distal endpoint reaches the origin.
When $\mbox{\small $L$} \!=\!  \mbox{\small $\tilde{L}$}$, this event occurs when the cable forms a~figure eight.
The critical value, $\mathrm{k}_c$, is determined by the condition $\xb(\mbox{\small $\tilde{L}$} ) \!=\! \xb( \mbox{\small $0$} )$,
where  $\xb$ 
is the cable coordinate along its elastica axis. Using Eq.~\eqref{eq.xb} for $\xb$,  it can be verified
that $\mathrm{k}_c \!=\! 0.909$. In the following characterization of the
stability regions the parameter $\mathrm{k}$ will vary in the interval $[0,\mathrm{k}_c]$.


Consider the stability regions depicted in relative endpoint space of Fig.~\ref{endpoint_regions.fig}. The outer boundary consists
of two curves,  $\Ag$ and $\Bg$.
The curve $\Ag$
traces the endpoint of a~maximally flattened cable satisfying $L \!=\! \rho \cdot\! \tilde{L}$. It is specified by the elastica solution $\Ag(\mathrm{k}) \!=\! (x(\mbox{\small $L$}),y(\mbox{\small $L$}))$  of Eq.~\eqref{eq.xy}, using $\mathrm{k} \!\in\! [0,\mathrm{k}_c]$ as the curve parameter.
This curve consists of an upper piece determined by 
$s_0 \!=\! \tfrac{\tilde{L}}{4} \!+\! \tfrac{\tilde{L}-L}{2}$ (cable starts at a~distance $\tfrac{\tilde{L}-L}{2}$ from the elastica first inflection point at~$\tfrac{\tilde{L}}{4}$), and
a~lower piece determined by the phase parameter by $s_0 \!=\! \tfrac{3 \tilde{L}}{4} \!+\! \tfrac{\tilde{L}-L}{2}$ (cable starts at a~distance $\tfrac{\tilde{L}-L}{2}$ from the elastica second inflection point at~$\tfrac{3\tilde{L}}{4}$).

The curve $\Bg$ in Fig.~\ref{endpoint_regions.fig} traces the cable endpoint when it folds inward towards the origin, obtained by
decreasing the elastica full period length from  $\mbox{\small $\tilde{L}$} \!=\! \mbox{\small $L$} / \rho$ down to $\mbox{\small $\tilde{L}$} \!=\! \mbox{\small $L$}$
while keeping the modulus parameter fixed at~$\mathrm{k}_c$.  This curve  is specified by the elastica solution $\Bg(\mbox{\scriptsize $\tilde{L}$})  \!=\! (x(\mbox{\scriptsize $\tilde{L}$}),y(\mbox{\scriptsize $\tilde{L}$}))$  of Eq.~\eqref{eq.xy}, using $\mbox{\small $\tilde{L}$} \!\in\! [\mbox{\small $L$}, \mbox{\small $L$} / \rho]$ as the curve parameter. The upper and lower pieces of~$\Bg$ are determined by the phase parameters $s_0 \!=\! \tfrac{\tilde{L}}{4} \!+\! \tfrac{\tilde{L}-L}{2}$ and $s_0 \!=\! \tfrac{3 \tilde{L}}{4} \!+\! \tfrac{\tilde{L}-L}{2}$.

The region bounded by $\Ag$ and $\Bg$ contains an~inner loop marked as $\Gg$ in Fig.~\ref{endpoint_regions.fig}. The inner loop traces the endpoint of a~flexible cable of length $\mbox{\small $L$} \!=\!  \mbox{\small $\tilde{L}$}$,
having two inflection points at the cable endpoints in addition to a~midpoint inflection point.
This curve is specified by the elastica solution $\Gg(\mathrm{k}) \!=\! (x(\mbox{\small $L$}),y(\mbox{\small $L$}))$  of Eq.~\eqref{eq.xy}, using
$\mathrm{k} \!\in\! [0,\mathrm{k}_c]$ as the curve parameter. The upper piece of $\Gg$ is determined by the phase parameter
$s_0 \!=\! \tfrac{\tilde{L}}{4}$  (cable starts at the elastica first inflection point and ends after full period length). The lower piece of $\Gg$ is determined by $s_0 \!=\! \tfrac{3\tilde{L}}{4}$ (cable starts at the elastica second inflection point and ends after full period length).

The relative position of the cable endpoint indicates {\em two types} of stable cable shapes when held with equal endpoint tangents.  
All cable endpoints in the {\em outer region} between $\Gg$ and the outer loop formed by $\Ag$ and $\Bg$ are associated with stable cable shapes of length $L \!<\! \tilde{L}$.
The stable cable shapes for endpoints in this region contain a~single inflection point at their midpoint (Fig.~\ref{endpoint_regions.fig}). All cable endpoints in the {\em inner region} bounded by $\Gg$
are associated with stable cable shapes of length $L \!=\! \tilde{L}$.
The stable cable shapes for endpoints in this region contain two interior inflection points (Fig.~\ref{endpoint_regions.fig}). 
\section{\bf Flexible Cable Approximation by Quadratic Arcs} \label{sec.bezier}
\vspace{-.02in}

\noindent This section describes an~approximation of flexible cable stable equilibrium shapes
by convex quadratic arcs to be used for efficient collision checking
during flexible cable steering.

\begin{figure}
\centering \includegraphics[width=.485\textwidth]{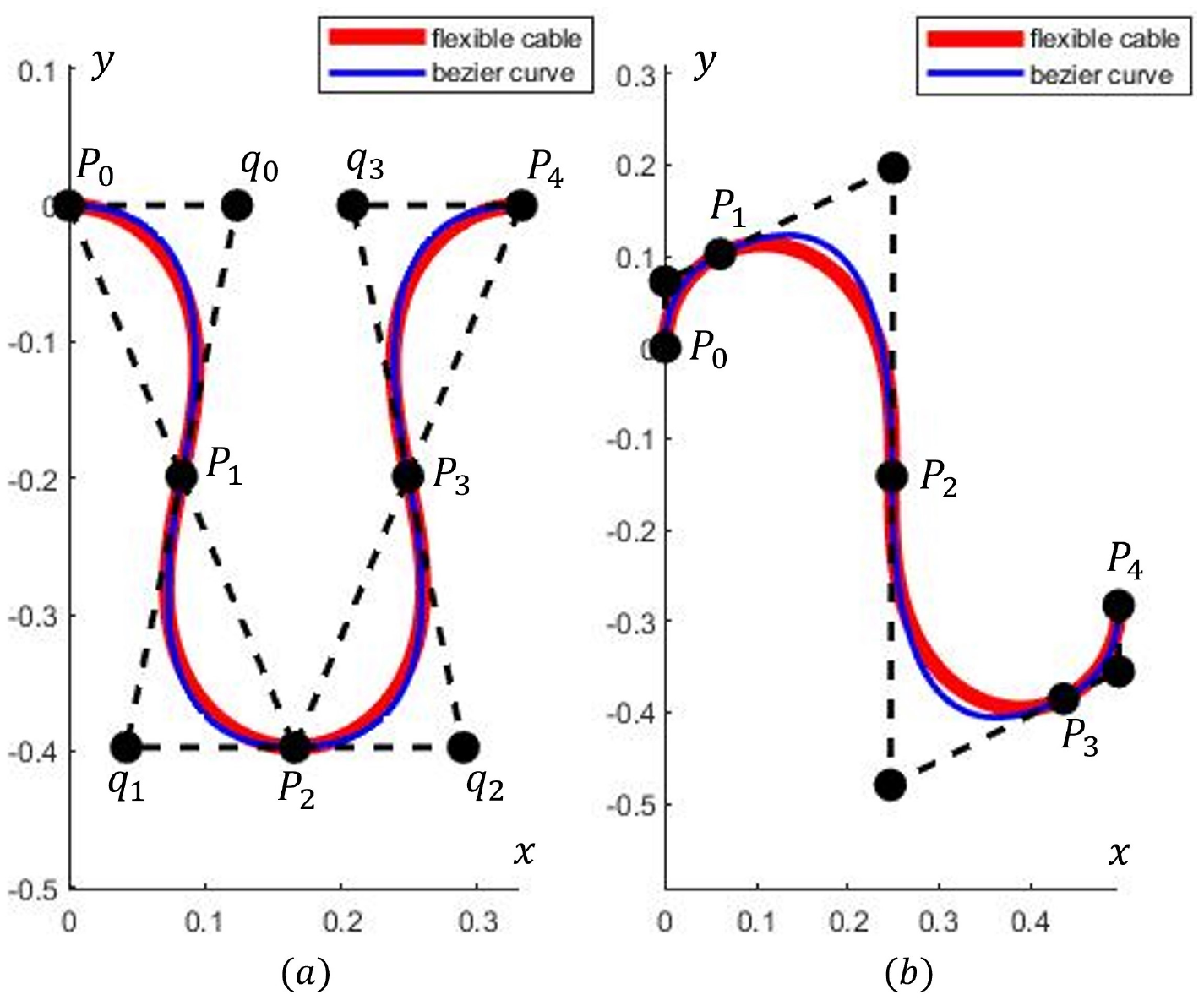}
\vspace{-.4in}
\caption{(a) Approximation of a~flexible cable shape (thin curve) having two interior inflection points $p_1$ and $p_3$ and a~maximum curvature point $p_2$ by four convex quadratic arcs. (b) Approximation of a~flexible cable shape (thin curve) having a~midpoint inflection point $p_2$
and two extreme curvature points $p_1$ and $p_3$ by four convex quadratic arcs, two of which form shorter arcs.}
\label{bezier.fig}
\vspace{-.2in}
\end{figure}

{\bf Control points selection:}
Each full period of the elastica
consists of four convex arcs having equal length.
To describe these arcs, consider the $(\phi\mbox{\small $(s)$},\kp\mbox{\small $(s)$})$ contours
depicted in Fig.~\ref{ovals.fig}. The contours form concentric convex loops that describe full periods of the elastica, $s \!\in\! [0,\mbox{\small $\tilde{L}$}]$.
Each loop contains four special points:  the upper point $(0,\kp_{max})$ of maximum curvature where $\phi(0) \!=\! 0$, the rightmost point $(\phi(\tfrac{\tilde{L}}{4}),0)$ which is the first inflection point at which $\kp \!=\! 0$, the bottom point
$(0,-\kp_{max})$ of minimum curvature where  $\phi(\tfrac{\tilde{L}}{2}) \!=\! 0$, and the leftmost point $(\phi(\tfrac{3\tilde{L}}{4}),0)$ which is the second inflection point at which $\kp \!=\! 0$.

When a flexible cable forms a~stable 
shape with equal endpoint tangents, it occupies up to full period length of the elastica (Corollary~4.1).
The cable starts according to its phase parameter, $s_0 \!\in\! [0,\mbox{\small $\tilde{L}$}]$,
and ends at $s_0 \!+\! \mbox{\small $L$}$ such that  $\mbox{\small $L$} \! \leq\! \mbox{\small $\tilde{L}$}$.
The flexible cable {\em control points} are
its two endpoints, its extreme curvature points when such points exist, and the cable inflection points (one or two such points exist in a~full period of the elastica).
The flexible cable can thus have up to six control points that divide the cable into up to five~convex~arcs~(Fig.~\ref{bezier.fig}).


{\bf B\'ezier polynomial construction:}  Let the flexible cable control points be indexed according to their appearance on the cable, $p_0,\ldots,p_n$ such that $p_0$ and $p_n$ are the cable endpoints and $n \!\leq\! 5$ (Fig.~\ref{bezier.fig}). Every pair of control points, $p_i$ and $p_{i+1}$, bounds a convex arc  approximated by a~convex quadratic arc as follows. The control points $p_i$ and $p_{i+1}$ are first augmented with an~intermediate control point, $q_i$, located at the intersection point of the cable's tangent lines at $p_i$ and $p_{i+1}$ (Fig.~\ref{bezier.fig}). The position of $p_i$ and $p_{i+1}$ and the tangents at these points are obtained from the elastica solution of Eq.~\eqref{eq.xy}. 
This data feeds a~simple formula for computing $q_i$ which is omitted. Using B\'ezier's technique~\cite{bezier}, each triplet $p_i$, $q_i$  and $p_{i+1}$ defines the quadratic polynomial
\vspace{-.08in}
\begin{equation} \label{eq.quadratic}
P_i(t) = (1 \!-\! t)^2 \cdot \vec{p}_i + 2(1 \!-\! t) t \cdot \vec{q}_i + t^2 \cdot \vec{p}_{i+1} \rtxt{ $t \in [0,1]$}
\vspace{-.1in}\end{equation}

\noindent where $i \!=\! 0 \ldots n \!-\! 1$.
Some intuition for $P_i(t)$ can be obtained by considering its midpoint  position 
\vspace{-.1in}
\[
P_i(\mbox{\small $\tfrac{1}{2}$}) \!=\!  \tfrac{1}{4} \vec{p}_i +  \tfrac{1}{2} \vec{q}_i + \tfrac{1}{4} \vec{p}_{i+1}.
\vspace{-.08in}
\]

\noindent The midpoint coefficients form barycentric coordinates that place the midpoint closer to the intermediate control point $q_i$ within the triangle formed by $p_i$, $q_i$ and $p_{i+1}$. Additional intuition can be gained from the quadratic polynomial tangent
\vspace{-.15in}
\[
 P'_i(t) = 2 \big( (t \!-\! 1) \cdot \vec{p}_i + (1 \!-\! 2t) \cdot \vec{q}_i + t \cdot \vec{p}_{i+1} \big) \rtxt{ $t \in [0,1]$}.
\vspace{-.08in}
\]

\noindent The endpoint tangents \mbox{\small $P'_i(0) \!=\! 2 ( \vec{q}_i \!-\! \vec{p}_i)$} and
\mbox{\small $P'_i(1) \!=\! 2 ( \vec{p}_{i+1} \!-\! \vec{q}_i )$} are aligned
with two edges of the triangle formed by  $p_i$, $q_i$ and $p_{i+1}$. In particular, these tangents are collinear with
 the elastica tangents at $p_i$ and $p_{i+1}$ for $i \!=\! 0 \ldots  n \!-\! 1$. The B\'ezier polynomials construction is
 summarized as Algorithms~$1$ and~$2$.

{\fontsize{10}{10}\selectfont
\begin{algorithm}{}
\caption{\bf Control Points Selection}
\begin{spacing}{0.95}
\begin{algorithmic}[0]
\renewcommand{\thealgorithm}{}

\Statex {{\bf Input:} cable start state $(x(0),y(0),\phi(0))$, cable length \mbox{\small $L$}, cable elastica  parameters
 $\Lgs$, $\mathrm{k}$ and $s_0$ with $\mbox{\small $\tilde{L}$}  \!=\! 4 K(\mathrm{k}) / \mbox{\scriptsize $\sqrt{\Lg}$}$.}
 \Statex {{\bf Data structures:} control points set $\Ps$, number of control points $2n \!+\! 1$, next control point position $\mbox{\small $s^*$}\!$.}
\Statex {{\bf Initialize:} $p_0 \!=\! (x(0),y(0))$, $\phi_0 \!=\! \phi(0)$, $\Ps \!=\! \{ p_0 \}$, $i \!=\! 0$ , $\mbox{\small $s^*$} \!=\! 0$.}
\Statex {\bf Set \boldmath{\small $s^*$}:} \vspace{-.05in}
\begin{enumerate}{}
\item If $s_0 \leq \tfrac{\mbox{\small $\tilde{L}$}}{4}$ and $s_0 \!+\! \mbox{\small $L$} \geq \tfrac{\mbox{\small $\tilde{L}$}}{4}$ set $\mbox{\small $s^*$} = \tfrac{\mbox{\small $\tilde{L}$}}{4}$.
\item If $s_0 \leq \tfrac{\mbox{\small $\tilde{L}$}}{2}$ and $s_0 \!+\! \mbox{\small $L$} \geq \tfrac{\mbox{\small $\tilde{L}$}}{2}$ set $\mbox{\small $s^*$} = \tfrac{\mbox{\small $\tilde{L}$}}{2}$.
\item If $s_0 \leq \tfrac{3\mbox{\small $\tilde{L}$}}{2}$ and $s_0 \!+\! \mbox{\small $L$} \geq \tfrac{3\mbox{\small $\tilde{L}$}}{2}$ set $\mbox{\small $s^*$} = \tfrac{3\mbox{\small $\tilde{L}$}}{2}$.
\item  If $s_0 \leq \mbox{\small $\tilde{L}$}$ and $s_0 \!+\! \mbox{\small $L$} \geq \mbox{\small $\tilde{L}$}$ set $\mbox{\small $s^*$} = \mbox{\small $\tilde{L}$}$.
\end{enumerate}

\While {$s_0 \leq \mbox{\small $s^*$} \leq  s_0 + \mbox{\small $L$}$}
\State {$p_{i+1} = ( x(\mbox{\small $s^*$} \!-\! s_0),  y(\mbox{\small $s^*$} \!-\! s_0) )$ using elastica solution of Eq.~\eqref{eq.xy}.}
\State {$\phi_{i+1} = \phi((\mbox{\small $s^*$} \!-\! s_0)$ using elastica tangent formula of Eq.~\eqref{eq.phi}.}
\State {Compute control point $q_i$ using  $(p_i,\phi_i)$ and $(p_{i+1},\phi_{i+1})$.}
\State {$\Ps \leftarrow \Ps \cup \{ q_i, p_{i+1} \}$.}
\State {$\mbox{\small $s^*$} = \mbox{\small $s^*$} + \mbox{\small $\tilde{L}$}/4$.}
\State {$i = i+1$.}
\EndWhile
\State { $n = i$, $p_n = (x(\mbox{\small $L$}),y(\mbox{\small $L$}))$, $\phi_n \!=\! \phi(0)$.}
\State {\hspace{.09em} Compute control point $q_{n-1}$ using  $(p_{n-1},\phi_{n-1})$ and $(p_n,\phi_n)$.}
\State {\hspace{.09em} $\Ps \leftarrow \Ps \cup \{  q_{n-1}, p_n \}$.}
\Statex {{\bf Output:} control points $\Ps = \{ p_0,q_0,p_1,q_1, \ldots, p_n\}$.}
\end{algorithmic}
\end{spacing}
\end{algorithm}

{\fontsize{10}{10}\selectfont
\begin{algorithm}{}
\caption{\bf B\'ezier Polynomials Construction}
\begin{spacing}{0.95}
\begin{algorithmic}[0]
\renewcommand{\thealgorithm}{}

\Statex {Use Algorithm~$1$ to compute control points $\Ps = \{ p_0,q_0,p_1, \ldots, p_n\}$, $n \leq 5$.}

\For {$i = 0$ to $n-1$} \\
Compute quadratic polynomial $P_i(t)$ according to Eq.~\eqref{eq.quadratic}.
\EndFor

\Statex {{\bf Output:}~\mbox{flexible}~\mbox{cable}~\mbox{approximation}~by~\mbox{$P_0(t),\ldots, P_{n-1}(t)$.}}
\end{algorithmic}
\end{spacing}
\end{algorithm}

\vspace{.02in}
{\bf Example:} Fig.~\ref{bezier.fig}(a) shows 
an~equilibrium shape that contains two interior inflection points $p_1$ and $p_3$
and a~maximum curvature point~$p_2$.
The 
elastica parameters 
are $\mbox{\small $L$} \!=\! \mbox{\small $\tilde{L}$}$, $s_0 \!=\! 0$ and 
$\mathrm{k} \!=\! 0.7746$.
The control points divide the cable into four convex arcs of length $\tilde{L}/4$.  Fig.~\ref{bezier.fig}(b) shows a~different stable equilibrium  shape that contains a~midpoint inflection point $p_2$ and two extreme curvature points $p_1$ and $p_3$.
The
elastica parameters in Fig.~\ref{bezier.fig}(b) are $\mbox{\small $L$} \!=\! \tfrac{2}{3} \mbox{\small $\tilde{L}$}$, $s_0 \!=\! 0.916 \mbox{\small $\tilde{L}$}$ and 
$\mathrm{k} \!=\! 0.8515$. The control points divide the cable into four convex arcs with shorter first and fourth arcs
(since $\mbox{\small $L$} \!=\! \tfrac{2}{3} \mbox{\small $\tilde{L}$}$).  In both examples, the four  quadratic polynomials
pass through the control points $p_0,\ldots,p_4$ with  tangents collinear with the cable tangents at the control points.
A measure of the  approximation quality is the {\em excess length} of the piecewise quadratic approximation relative to the cable length.
It is $1.6$\% longer in Fig.~\ref{bezier.fig}(a) and $4.2$\% longer in Fig.~\ref{bezier.fig}(b).~\eex
\vspace{-.04in}


\section{\bf Flexible Cable Steering Scheme} \label{sec.scheme}
\vspace{-.01in}

\noindent This section describes the flexible cable steering scheme
which relies on the tools developed in previous sections. The scheme consists of three stages.
In the first stage the flexible cable relative endpoint positions are computed as a~function of three elastica parameters.
The second stage uses the relative endpoint positions and their elastica parameters to initialize a~{\small 5-D} grid of the cable configuration space.
The third stage performs motion planning in the flexible cable {\small 5-D} configuration space.

{\bf Computation~of~cable~\mbox{relative}~\mbox{endpoint}~\mbox{positions:}} The flexible cable
relative endpoint positions, denoted $(\mbox{\small $X_L$},\mbox{\small $Y_L$}) \!=\! (x(\mbox{\small $L$}) \!-\! x(\mbox{\small $0$}),y(\mbox{\small $L$}) \!-\! y(\mbox{\small $0$}))$,  are computed using three elastica parameters:
the modulus parameter, $\mathrm{k}$, the phase parameter,~$s_0$,
and the full period length of the elastica, $\mbox{\small $\tilde{L}$} \!=\! 4 K(\mathrm{k}) / \mbox{\small $\sqrt{\Lg}$}$.  
Here $\mbox{\small $\tilde{L}$}$ replaces the co-state parameter $\mbox{\small $\Lgs$}$, since the stable cable shapes described in Section~\ref{sec.stability} are associated with cable lengths $\mbox{\small $L$} \!<\! \mbox{\small $\tilde{L}$}$ and $\mbox{\small $L$} \!=\! \mbox{\small $\tilde{L}$}$.

The three elastica parameters vary in a~manner that ensures stable non self-intersecting cable shapes.
The modulus parameter varies in the interval $0 \!\leq\! \mathrm{k} \!\leq\! \mathrm{k}_{max}$,
where $\mathrm{k} \!=\! 0$ is a~straight-line shape while $\mathrm{k} \!\leq\! \mathrm{k}_{max}$ ensures that the cable is not self-intersecting (Section~\ref{sec.self_intersect}). The phase parameter $s_0$ varies in the interval $[\mbox{\small $L$}/4, 3\mbox{\small $L$}/4 ]$. This interval ensures full coverage of the endpoints $(\mbox{\small $X_L$},\mbox{\small $Y_L$})$ associated with
the stable cable shapes.
The parameter \mbox{\small $\tilde{L}$} varies in the interval $[\mbox{\small $L$}, \mbox{\small $L$}/\rho]$, where $0 \!<\! \rho \!<\! 1$ is a~maximal flattening parameter (Section~\ref{sec.stability}).

\begin{figure}
\centering \includegraphics[width=.5\textwidth]{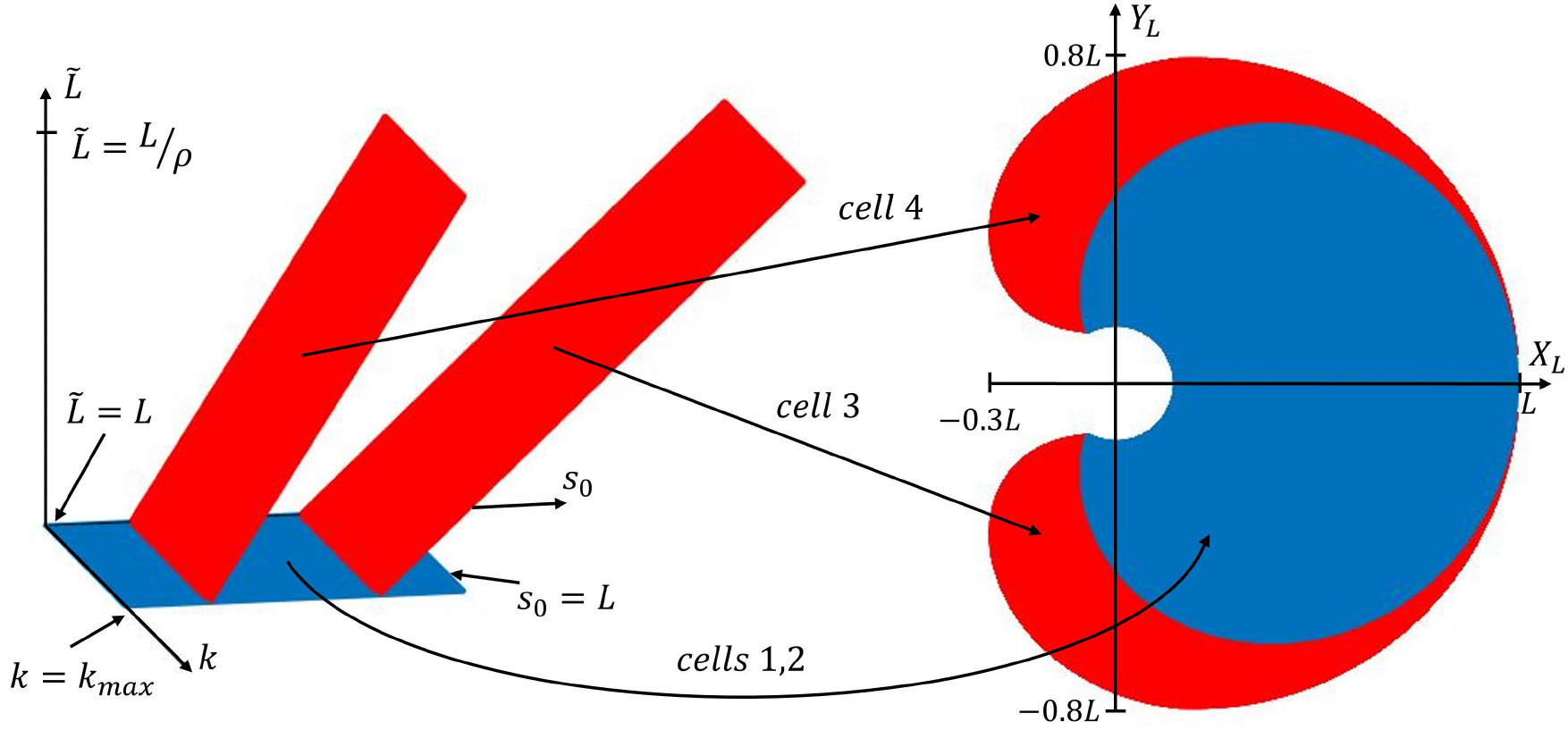}
\vspace{-.28in}
\caption{The relative endpoint positions \mbox{\scriptsize $(X_L,Y_L)$} are computed by Algorithm~3 over four cells
of the elastica parameters $(\mathrm{k},s_0,\mbox{\scriptsize $\tilde{L}$})$. Cells 1 and~2 parametrize stable non self-intersecting cable shapes satisfying $L \!=\! \tilde{L}$. Cells  3 and~4 parametrize stable non self-intersecting cable shapes satisfying $L \!<\! \tilde{L}$.}
\label{cells.fig}
\vspace{-.1in}
\end{figure}

The computation of $(\mbox{\small $X_L$},\mbox{\small $Y_L$})$ is summarized as Algorithm~3. The algorithm uses $n_{\mathrm{k}}$, $n_{s_0}$ and $n_{\mbox{\tiny $\tilde{L}$}}$ discrete values of the elastica parameters
to compute $(\mbox{\small $X_L$},\mbox{\small $Y_L$})$ over {\em four planar cells} embedded in the space $(\mathrm{k},s_0,\mbox{\small $\tilde{L}$})$. Two cells correspond to stable cable shapes satisfying $\mbox{\small $L$} \!=\! \mbox{\small $\tilde{L}$}$ (horizontal cells 1 and~2 in Fig.~\ref{cells.fig}(a)), the other two cells correspond to stable cable shapes satisfying $\mbox{\small $L$} \!<\! \mbox{\small $\tilde{L}$}$ (diagonal cells 3 and~4 in Fig.~\ref{cells.fig}(a)).  For each value of $(\mathrm{k},s_0,\mbox{\small $\tilde{L}$})$, the cable relative endpoint position is computed according to Eq.~\eqref{eq.xy}
\vspace{-.08in}
\begin{equation} \label{eq.XY}
 \vctwo{\!\! X_L(\mathrm{k},s_0,\mbox{\scriptsize $\tilde{L}$}) \!\!}{\!\! Y_L(\mathrm{k},s_0,\mbox{\scriptsize $\tilde{L}$}) \!\!}    \!=\!
   \frac{\mbox{\small $1$}}{\Lgs} \mbox{\small  $ \matwo{\cos\phi_0}{\sin\phi_0 }{\sin\phi_0}{-\cos\phi_0 }$}
  \vctwo{\!\! \int_0^L \hlf \kp^2(s) ds + \Lg \Sg \!\!\cdot\! \mbox{\small $L$} \!\!}{\!\!  \kp(\mbox{\small $L$}) \!-\!  \kp(\mbox{\small $0$})  \!\!}
\vspace{-.02in}
\end{equation}

\noindent where $\phi_0$ is specified in Eq.~\eqref{eq.phi0}, $\Sg  \!=\! 1 \!-\! 2\mathrm{k}^2$, $\Lgs \!=\! \Lgs_r / \mbox{\scriptsize $EI$}$, and the cable curvatures
$\kp(0)$ and $\kp(\mbox{\small $L$})$ are specified in Eq.~\eqref{eq.kp}.

{\fontsize{10}{10}\selectfont
\begin{algorithm}{}
\caption{\bf Relative Endpoint Positions}
\begin{spacing}{0.95}
\begin{algorithmic}[0]
\renewcommand{\thealgorithm}{}

\Statex {{\bf Input:}  cable length \mbox{\small $L$};~elastica~parameters~$(\mathrm{k},s_0,\mbox{\scriptsize $\tilde{L}$})$
discretized into $(n_{\mathrm{k}},n_{s_0},n_{\mbox{\tiny $\tilde{L}$}})$ values in $[0,\!\mathrm{k}_{max}]$, $[\tfrac{L}{4},\!\tfrac{3L}{4}]$ and $[L,\!\tfrac{L}{\rho}]$.}
 \Statex {{\bf Data structures:} $EndPoint$  array of $(X_L,Y_L)$, each storing elastica parameter triplets  $(\mathrm{k},s_{0,1},\mbox{\scriptsize $\tilde{L}$})$ and $(\mathrm{k},s_{0,2},\mbox{\scriptsize $\tilde{L}$})$.}
 \Statex {{\bf Initialize:} $EndPoint \gets [0]$.}
 \Statex {\bf /*full period cable shapes (cells 1,2 in Fig.~\ref{cells.fig})*/}
 \State{\bf Set $\tilde{L}=L$}
\For{$i=1$ to $n_{\mathrm{k}}$}
   \For{$j=1$ to $n_{s_0}$}
       \State{Compute $X_L\big(\mathrm{k}(i),s_0(j),\mbox{\scriptsize  $\tilde{L}$}\big) ,Y_L\big(\mathrm{k}(i),s_0(j),\mbox{\scriptsize  $\tilde{L}$}\big)$}
       \State{$s_{0,1} = s_0(j)$}
       \State{{\bf if} $0 \leq s_{0,1} \!\leq\! \tfrac{L}{2}$ set $s_{0,2} \!=\! \tfrac{L}{2} \!-\! s_{0,1}$ {\bf else} $s_{0,2} \!=\! \tfrac{3L}{2} \!-\! s_{0,1}$}
       \State{$EndPoint[X_L,Y_L] \gets \big( k(i),s_{0,1},\mbox{\scriptsize  $\tilde{L}$}\big), \big( k(i),s_{0,2},\mbox{\scriptsize  $\tilde{L}$}\big)$}
    \EndFor
\EndFor
\Statex {\bf /*less than full period cable shapes (cells 3,4 in Fig.~\ref{cells.fig})*/}
\For{$i=1$ to $n_{\mathrm{k}}$}
   \For{$j=1$ to $n_{\mbox{\scriptsize $\tilde{L}$}}$}
   \vspace{.01in}
   \State{\bf Set $s_{0,1} \!=\! \tfrac{3\tilde{L}(j)-2 L}{4}$, $s_{0,2} \!=\! \tfrac{5\tilde{L}(j)-2 L}{4}$}
   \State{Compute $X_L\big(\mathrm{k}(i),s_{0,1},\mbox{\scriptsize $\tilde{L}$}(j)\big),
             Y_L\big(\mathrm{k}(i),s_{0,1},\mbox{\scriptsize $\tilde{L}$}(j)\big)$}
   \State{$EndPoint[X_L,Y_L] \gets \big( k(i),s_{0,1},\mbox{\scriptsize $\tilde{L}$}(j) \big)$}
   \State{Compute $X_L\big(\mathrm{k}(i),s_{0,2},\mbox{\scriptsize $\tilde{L}$}(j)\big) ,Y_L\big(\mathrm{k}(i),s_{0,2},\mbox{\scriptsize $\tilde{L}$}(j)\big)$}
   \State{$EndPoint[X_L,Y_L] \gets \big( k(i),s_{0,2},\mbox{\scriptsize $\tilde{L}$}(j) \big)$}
 \EndFor
\EndFor
\Statex {{\bf Output:} $EndPoint$  array of $(X_L,Y_L)$.}

\end{algorithmic}
\end{spacing}
\end{algorithm}

\begin{figure}
\centering \includegraphics[width=.5\textwidth]{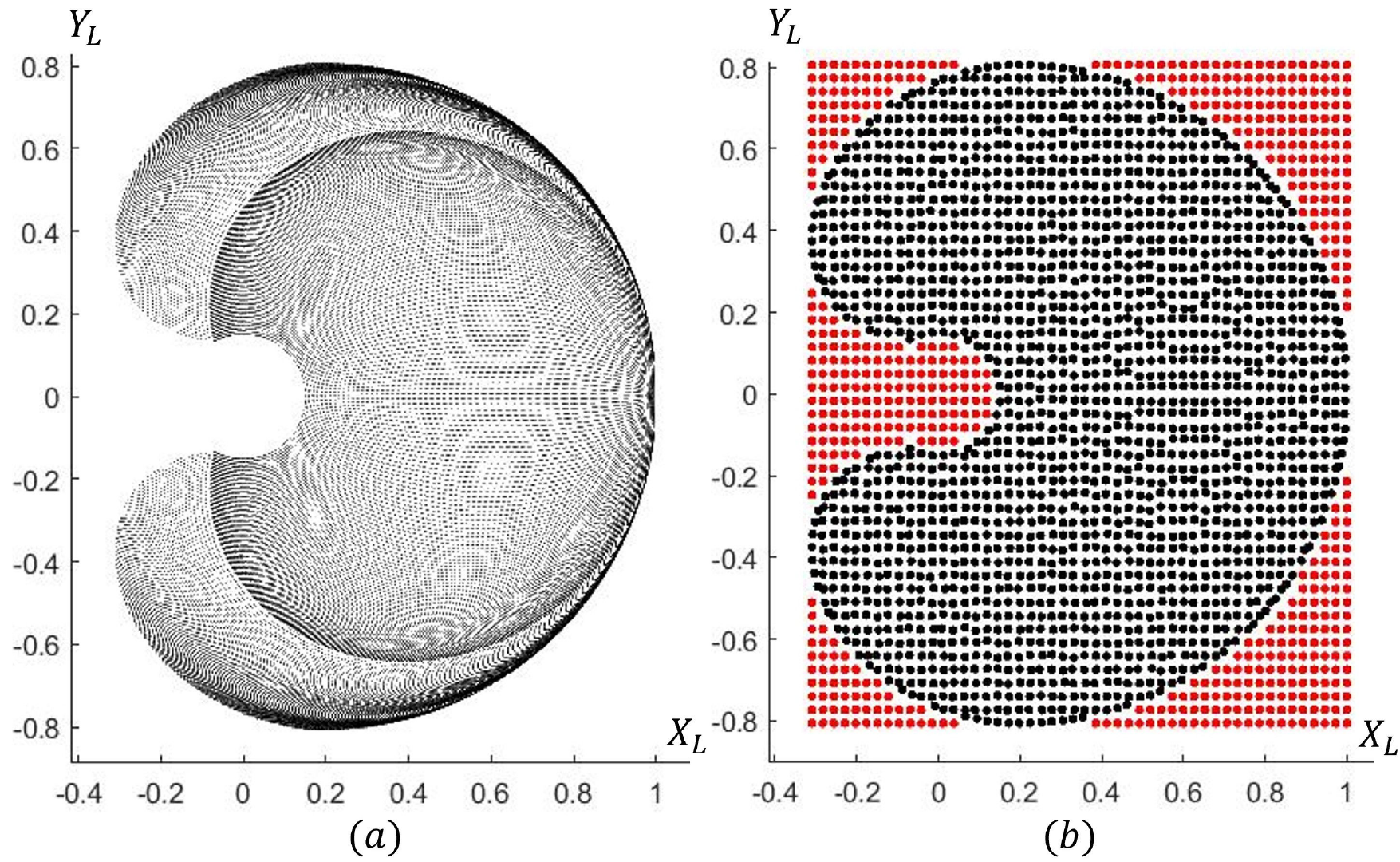}
\vspace{-.32in}
\caption{(a) The relative endpoint positions \mbox{\scriptsize $(X_L,Y_L)$} computed by Algorithm~3
over the elastica parameter cells shown in Fig.~\ref{cells.fig}.
(b) The relative endpoint positions
lumped into an $\mbox{\scriptsize $N$} \!\times \mbox{\scriptsize $N$}$
array whose inner black cells represent stable non self-intersecting cable shapes. These cells store
the elastica parameters that determine the corresponding cable shape.}
\label{2d_grid.fig}
\vspace{-.1in}
\end{figure}

\vspace{.02in}
{\bf Example:}  The  relative endpoint positions, $(\mbox{\small $X_L$},\mbox{\small $Y_L$})$,
computed by Algorithm~3 using Eq.~\eqref{eq.XY} are shown in Fig.~\ref{2d_grid.fig}(a).
These endpoints were computed over the cells shown in Fig.~\ref{cells.fig} using elastica parameters $(\mathrm{k},s_0,\mbox{\scriptsize $\tilde{L}$})$ discretized into $n_{\mathrm{k}} \!=\! 160$, $n_{s_0} \!=\! 200$ and $n_{\mbox{\scriptsize $\tilde{L}$}} \!=\! 100$ values, thus giving
$n_{\mathrm{k}} n_{s_0} \!+\! 2 n_{\mathrm{k}} n_{\mbox{\tiny $\tilde{L}$}} \!=\! \mbox{\small $64,000$}$  relative endpoint positions.
These endpoint positions are used
to initialize the flexible cable {\small 5-D} configuration space grid as next described.~\eex
\vspace{.02in}

{\bf Initialization~of~cable~configuration~space~grid:} The flexible cable {\small 5-D} configuration space grid, $CSpace$, consists of  $(x(0),y(0),\phi(0))$ and $(\mbox{\small $X_L$},\mbox{\small $Y_L$})$. The cable base frame position, $(x\mbox{\small $(0)$},y\mbox{\small $(0)$})$, varies in a~rectangular environment
discretized into $n \!\times\! n$ cells while the cable base frame orientation, $\phi(\mbox{\small $0$})$, 
is discretized into $m$ cells. The $CSpace$ grid
is initialized in two steps. In the first step the relative endpoint positions stored in $EndPoint$ are lumped into an~$\mbox{\small $N$} \!\times \! \mbox{\small $N$}$ array, $EndPointGrid$, shown as a~$50 \!\times \! 50$ grid in Fig.~\ref{2d_grid.fig}(b).
The inner black cells
represent stable non self-intersecting cable shapes. Each of these cells
stores two elastica triplets that represent two cable shapes in cells $1$ and $2$ of Fig~\ref{cells.fig}(a),  and a~single cable shape in cells $3$ and $4$  of Fig~\ref{cells.fig}(a).  For each inner black cell 
of $EndPointGrid$,  all cells of $CSpace$ having this $(\mbox{\small $X_L$},\mbox{\small $Y_L$})$ coordinate are marked as {\em feasible} in $CSpace$.
These cells correspond to stable and non self-intersecting cable shapes and their elastica
parameters are kept in the underlying $EndPointGrid$.
All cells of $CSpace$
that project to the outer red cells in Fig.~\ref{2d_grid.fig}(b) are marked as {\em infeasible.} 


{\fontsize{10}{10}\selectfont
\begin{algorithm}{}
\caption{\bf Flexible Cable Steering Scheme}
\begin{spacing}{0.95}
\begin{algorithmic}[0]
\renewcommand{\thealgorithm}{}

\Statex {{\bf Input:}  cable length \mbox{\small $L$}; cable $CSpace$ grid; start $S$ and target $T$ specified
as feasible cells of $CSpace$. Polygonal obstacles $\Bs_1,\ldots,\Bs_k$.}
\Statex {{\bf Data structures:} Open list $\Os$, closed list $\Cs$.}
\Statex {{\bf Initialize:} $\Os = \{ S \}$, $\Cs = \emptyset$.}
\Statex {\bf /*Search for shortest \boldmath{$CSpace$} path from start to target*/}
\While {$\Os \neq \emptyset$}
    \State{Select $z^* \!\!\in\! \Os$ according to $A^*$ search criterion.}  
    \State{{\bf if} $z^* = T$ {\bf then}  move to output stage.}
    \State{Move $z^*$ from $\Os$ to $\Cs$.}
    \For{each feasible neighbor $z$ of $z^*$ in $CSpace$ s.t. \hspace*{.2in}$z \nin \Cs$}
    \State{Extract cable base frame $(x(0),y(0),\phi(0))$  of $z$.}
    \State{Extract the cable elastica triplets 
    $(\mathrm{k}, s_{0,1}, \tilde{L})$ and \hspace*{.4in}$(\mathrm{k}, s_{0,2}, \tilde{L})$ from $z$.}
    \State{{\bf if} $L = \tilde{L}$ {\bf then}  set $s_0 = s_{0,1}$ {\small /*$s_{0,1} = s_{0,2}$ in this case*/}}
    \State{{\bf else} set $s_0 \!=\! s_{0,1}$ when $\kappa(0)>0$ and}
    \State{\hspace*{.22in} set $s_0 = s_{0,2}$ when $\kappa(0)<0$.}
    \State{$P(t) \gets \mathrm{BezierPolynomial}(\mathrm{k}, s_0, \tilde{L})$}
    \State{$Q(t) \gets \mathrm{Position}\big(P(t), x(0), y(0), \phi(0)\big)$}
    \State{{\bf if} $Collision(Q(t), \mathcal{B}_1,\ldots,\mathcal{B}_k) = \mathrm{FALSE}$ \\
    \hspace*{.4in}{\bf then} add $z$ to $\mathcal{O}$}
    \EndFor
\EndWhile
\Statex {{\bf Output:} $CSpace$ path from $S$ to $T$, when such a path exists.}
\end{algorithmic}
\end{spacing}
\end{algorithm}

{\bf Motion planning in {\small 5-D} configuration space:} The flexible cable steering scheme
is summarized as Algorithm~4. The algorithm is given  feasible start and target cells in $CSpace$ 
and a~description of polygonal obstacles in the environment.
The main loop  of Algorithm~4 selects the current best node in the open list, $z^* \!\in\! \Os$, as detailed below.
For each {\em feasible} neighbor $z \!\in\! CSpace$ of $z^* \!$,
the elastica parameters of~$z$ are extracted and fed into Algorithm~3 that computes the B\'ezier polynomial $P(t)$  for the cable shape.
The cable base frame position 
extracted from~$z$ is used to place
the B\'ezier polynomial in the physical environment as $Q(t)$, then  collision of $Q(t)$
with  the polygonal obstacles is efficiently checked by $Collision\mbox{\small $(Q(t), \Bs_1,\ldots,\Bs_k)$}$}.
All collision free neighboring  cells of $z^*$ are added  to the open list, $\Os$, and the main loop of Algorithm~4 resumes.
Eventually the target becomes the best node in~$\Os$ or the open list becomes empty.
In the first case  the path stored in the closed list, $\Cs$, provides the shortest $CSpace$ path from $S$ to $T$. In the latter case a~feasible $CSpace$
path from $S$ to $T$ does~not~exist. 
\vspace{-.06in}

\section{\bf Representative Examples} \label{sec.examples}
\vspace{-.01in}

\noindent This section describes execution examples of the steering scheme
described in Section~\ref{sec.scheme}. The scheme
was implemented on {\small MATLAB} using elliptic functions library~\cite{batista_matlab}.\footnote{All simulations run {\scriptsize MATLAB-22B} on {\scriptsize DELL} OptiPlex 7080, with Intel I7-10700 {\scriptsize CPU} running at 2.90 GHz and 16 {\scriptsize GB} main memory.}
In all examples the flexible cable
maximal flattening parameter is set to $\rho \!=\! 0.5$, so that  \mbox{\small $\tilde{L}$}
is at most twice the physical cable length.


\begin{figure}
\centering \includegraphics[width=.33\textwidth]{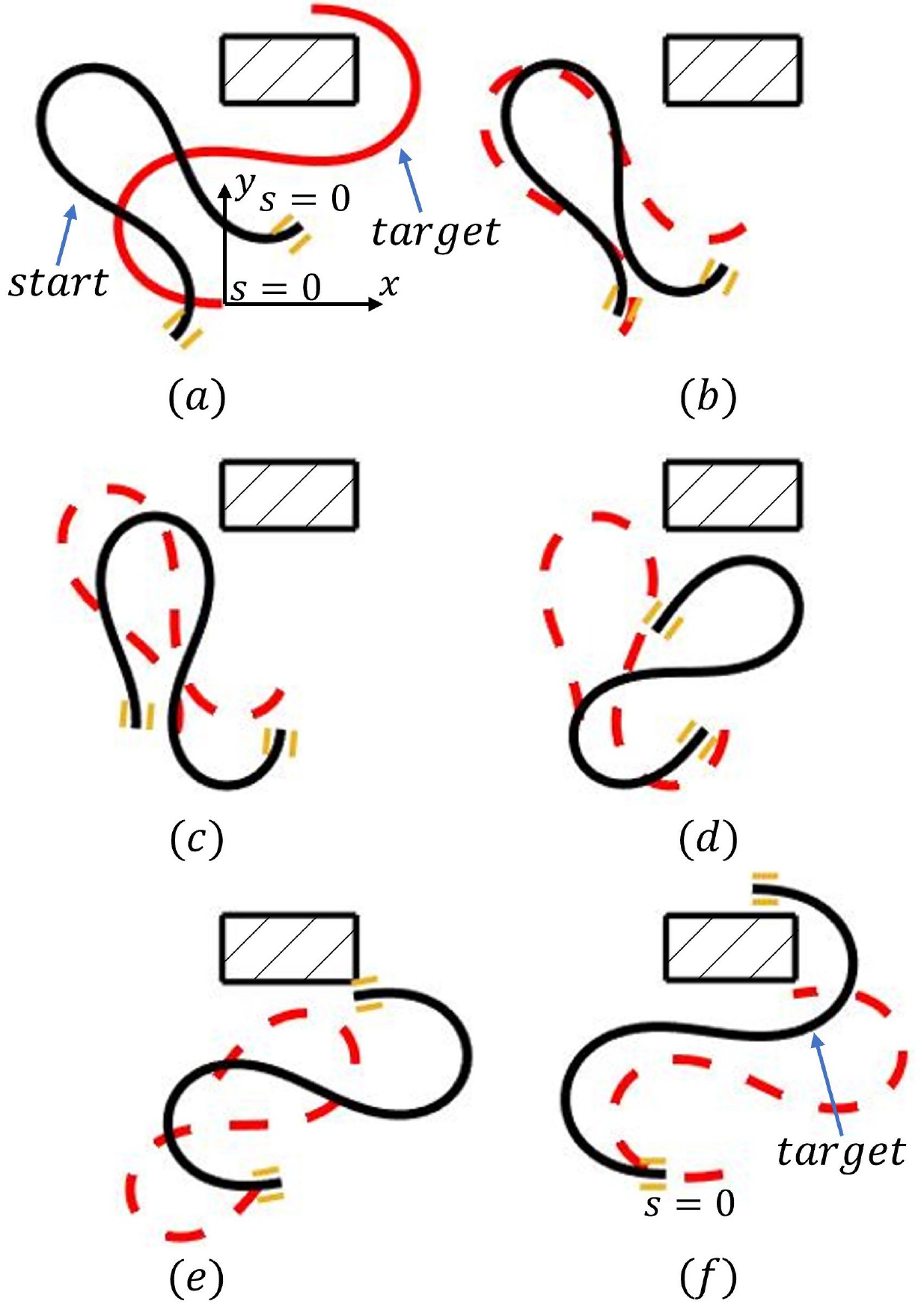}
\vspace{-.15in}
\caption{Flexible cable steering around a single obstacle. (a) Flexible cable position and shape at the start and target. (b)-(c)~Cable folds inward until reaching self-intersection limit while rotating away from the obstacle. (d)-(e)~Cable moves to the right along obstacle bottom edge while rotating into horizontal position. (f)~Cable stretches outward while circumnavigating the obstacle until its distal endpoint reaches the endpoint target position.}
\label{sim1.fig}
\vspace{-.2in}
\end{figure}

{\bf Example 1---single obstacle:} The example shown in Fig.~\ref{sim1.fig} steers a~unit length flexible cable
in the presence~of~a~single obstacle. The cable starts at
the base frame position $(x\mbox{\small $(0)$},y\mbox{\small $(0)$},\phi(0)) \!=\! (0.12, 0.12, -135^{\circ})$~with~\mbox{elastica}~paramet\-ers
$(\mathrm{k},s_0,\mbox{\small $\tilde{L}$}) \!=\! (0.671,0,\mbox{\small $L$})$. The 
start~shape~forms~full period of the elastica 
with co-axial endpoint tangents  (Fig.~\ref{sim1.fig}(a)).
The cable target base frame position is $(x\mbox{\small $(0)$},y\mbox{\small $(0)$},\phi(0)) \!=\! (0,0,180^{\circ})$ with elastica parameters $(\mathrm{k},s_0,\mbox{\small $\tilde{L}$}) \!=\! (0.707,0.9 \mbox{\small $L$}, 1.12 \mbox{\small $L$})$. The 
target
shape forms $1/1.12$ of the elastica full period with
equal endpoint tangents (Fig.~\ref{sim1.fig}(a)).

Algorithm~4 computed the steering path shown in Fig.~\ref{sim1.fig} in 
$138$ seconds (see video clip). The cable first folds inward until reaching self-intersection limit while rotating 
away from the obstacle (Fig.~\ref{sim1.fig}(b)-(c)). The cable now slides with small clearance along the obstacle bottom edge while rotating 
into horizontal position (Fig.~\ref{sim1.fig}(d)). When the cable distal endpoint reaches the obstacle bottom right corner (Fig.~\ref{sim1.fig}(e)), the cable stretches outward by increasing \mbox{\small $\tilde{L}$} and decreasing $\mathrm{k}$ until it reaches the endpoint target position (Fig.~\ref{sim1.fig}(f)). Note that equal endpoint tangents are maintained during cable steering~\eex

\begin{figure}
\centering \includegraphics[width=.5\textwidth]{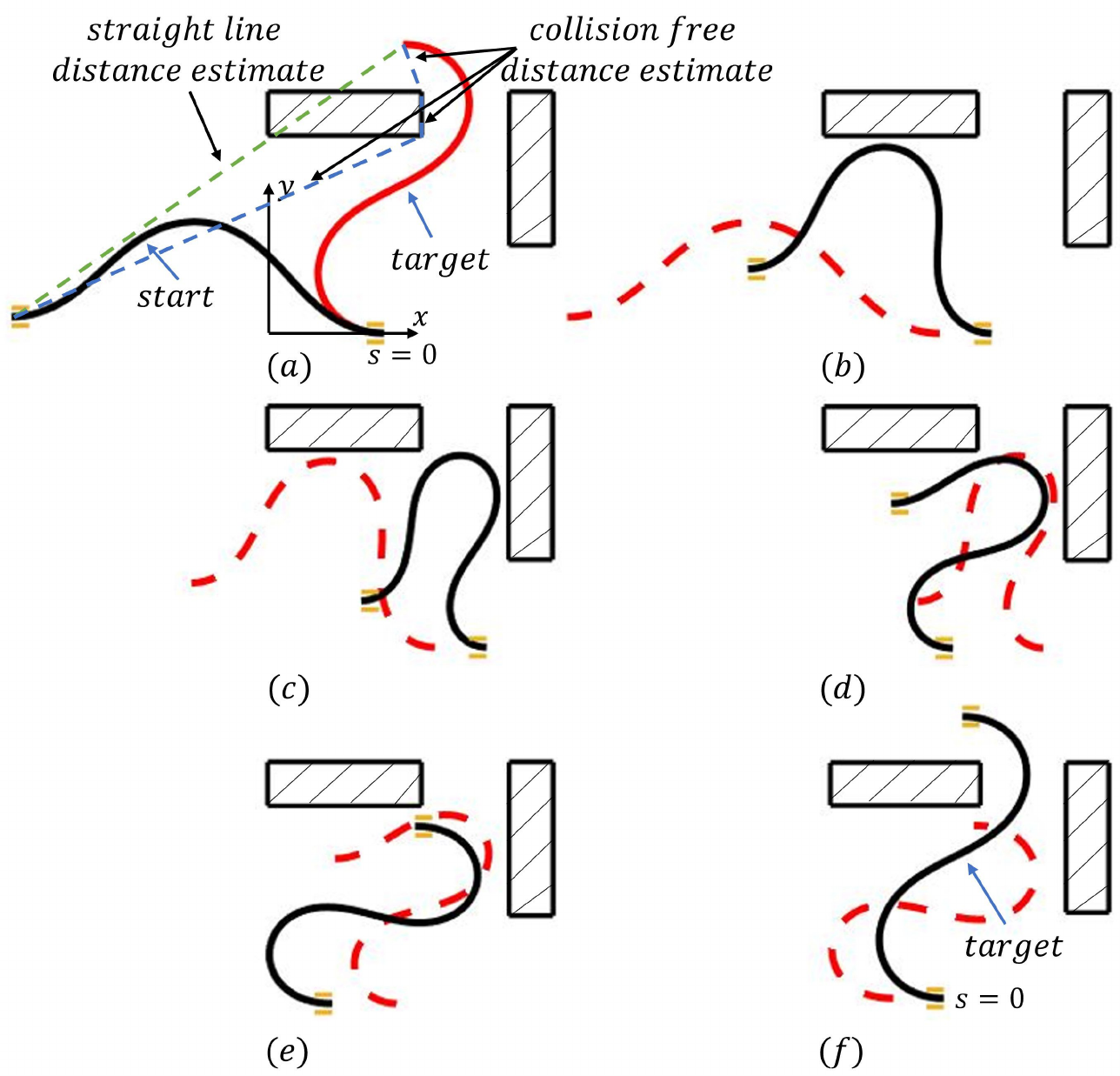}
\vspace{-.33in}
\caption{Flexible cable steering through an~opening between two obstacles. (a)~Start and target
have the same base frame position (see text for distance to target estimates). (b)~Cable folds inward while
its mid-section slides under left-obstacle bottom edge. (c)~Cable reaches the right obstacle. (d)-(e)~Cable
moves away from both obstacles while stretching.  (f)~Cable moves between obstacles
by further stretching until it reaches the distal endpoint position.}
\label{sim2.fig}
\vspace{-.2in}
\end{figure}

{\bf Example 2---two obstacles:} The example shown in Fig.~\ref{sim2.fig} steers a~unit length flexible cable between two obstacles.
The cable start and target share the same base frame position at  $(x\mbox{\small $(0)$},y\mbox{\small $(0)$},\phi(0)) \!=\! (0.26, 0, 180^{\circ})$. The cable elastica parameters for the start shape are $(\mathrm{k},s_0,\mbox{\small $\tilde{L}$}) \!=\! (0.5,0,\mbox{\small $L$})$. The 
start shape forms
full period of the elastica with
coaxial endpoint tangents (Fig.~\ref{sim2.fig}(a)). The cable elastica parameters for the target shape are
$(\mathrm{k},s_0,\mbox{\small $\tilde{L}$}) \!=\! (0.707,1.15 \mbox{\small $L$}, 1.32 \mbox{\small $L$})$.
The 
target shape forms $1/1.32$ of the elastica full period with
equal endpoint tangents (Fig.~\ref{sim2.fig}(a)).

Algorithm~4 computed the steering path shown in Fig.~\ref{sim2.fig} in 
$40$ seconds (see video clip). The cable first folds inward while
its mid-section slides with small clearance along the left-obstacle bottom edge  (Fig.~\ref{sim2.fig}(b)).  When the cable mid-section reaches the right obstacle (Fig.~\ref{sim2.fig}(c)), it moves away from both obstacles while increasing \mbox{\small $\tilde{L}$} in order to stretch and reduce height with respect to its elastica axis (Fig.~\ref{sim2.fig}(d)-(e)). The cable distal endpoint now moves between the two obstacles
by further increasing \mbox{\small $\tilde{L}$} while decreasing $\mathrm{k}$ in order to bring the cable endpoint to its target position
(Fig.~\ref{sim2.fig}(f)).~\eex
\vspace{.01in}

{\bf Practical \boldmath{$A^*$} path planning:} Algorithm~4 searches the feasible cells of $CSpace$ using the open list nodes, $z \!\in\! \Os$,  sorted by {\em trip length estimate} from {\small $S$} to {\small $T$} through~$z$.  Trip length is measured as the sum $l(\mbox{\small $S$},z) \!+\! l(z,\mbox{\small $T$})$, where $l(\mbox{\small $S$},z)$ measures the c-space path length traveled from $S$ to $z$ while   $l(z,\mbox{\small $T$})$  estimates the c-space path length yet to be travelled from $z$ to {\small $T$}. The simplest choice for $l(z,\mbox{\small $T$})$ would be Euclidean distance to the target computed in $CSpace$
\vspace{-.06in}
\[
\barr
l(z,\mbox{\small $T$}) \! \!  &  \!=\! \mbox{\small $\left( (x_z(0)\!-\!x_T(0))^2 \!\!+\!  (x_z(0)\!-\!x_T(0))^2 \!\!+\! a \!\cdot\! (\phi_z(0)\!-\!\phi_T(0))^2 \right. $}\\
& \mbox{\small $\left. + (\mbox{\small $X_{L,z}$} \!-\! \mbox{\small $X_{L,T}$})^2
 \!+\! (\mbox{\small $Y_{L,z}$} \!-\! \mbox{\small $Y_{L,T}$})^2
\right)^{1/2} $}
\earr
\vspace{-.07in}
\]

\noindent where \mbox{\small $(x_z(0),y_z(0),\phi_z(0),\mbox{\small $X_{L,z}$},\mbox{\small $Y_{L,z}$})$} are the c-space coordinates of $z$,
\mbox{\small $(x_T(0),y_T(0),\phi_T(0),\mbox{\small $X_{L,T}$},\mbox{\small $Y_{L,T}$})$} are the c-space coordinates of {\small $T$} and $a \!>\! 0$ is
a~scaling constant.  Using the Euclidean distance estimate,
Algorithm~4 took 
$168$ seconds to compute  a~steering path for Example~2. When the start and target were switched in Example~2,
Algorithm~4 took significantly longer time of  
$816$ seconds  to compute a~steering path.
However, one expects {\em shorter} runtime when the
cable starts within the opening between the two obstacles.

A practical speedup that also gives shorter computation time when the start and target are switched
works when the cable's start-and-target base frames are located close to each other.
In these situations one can compute the {\em shortest  collision free path} for the cable distal endpoint from start to target
(Fig.~\ref{sim2.fig}(a)). The shortest  collision free path is piecewise linear with vertices located at convex obstacle vertices. Each vertex of the  shortest  collision free path
defines intermediate target position for the cable distal endpoint. The cable is now steered
using the Euclidean distance estimate for each intermediate target.
In Example~2, three intermediate endpoint targets allowed  Algorithm~4 to compute a~steering path from start to target in 
$40$ seconds and in shorter time of 
$37$ seconds for the switched start and target in this example.

\begin{figure}
\centering \includegraphics[width=.4\textwidth]{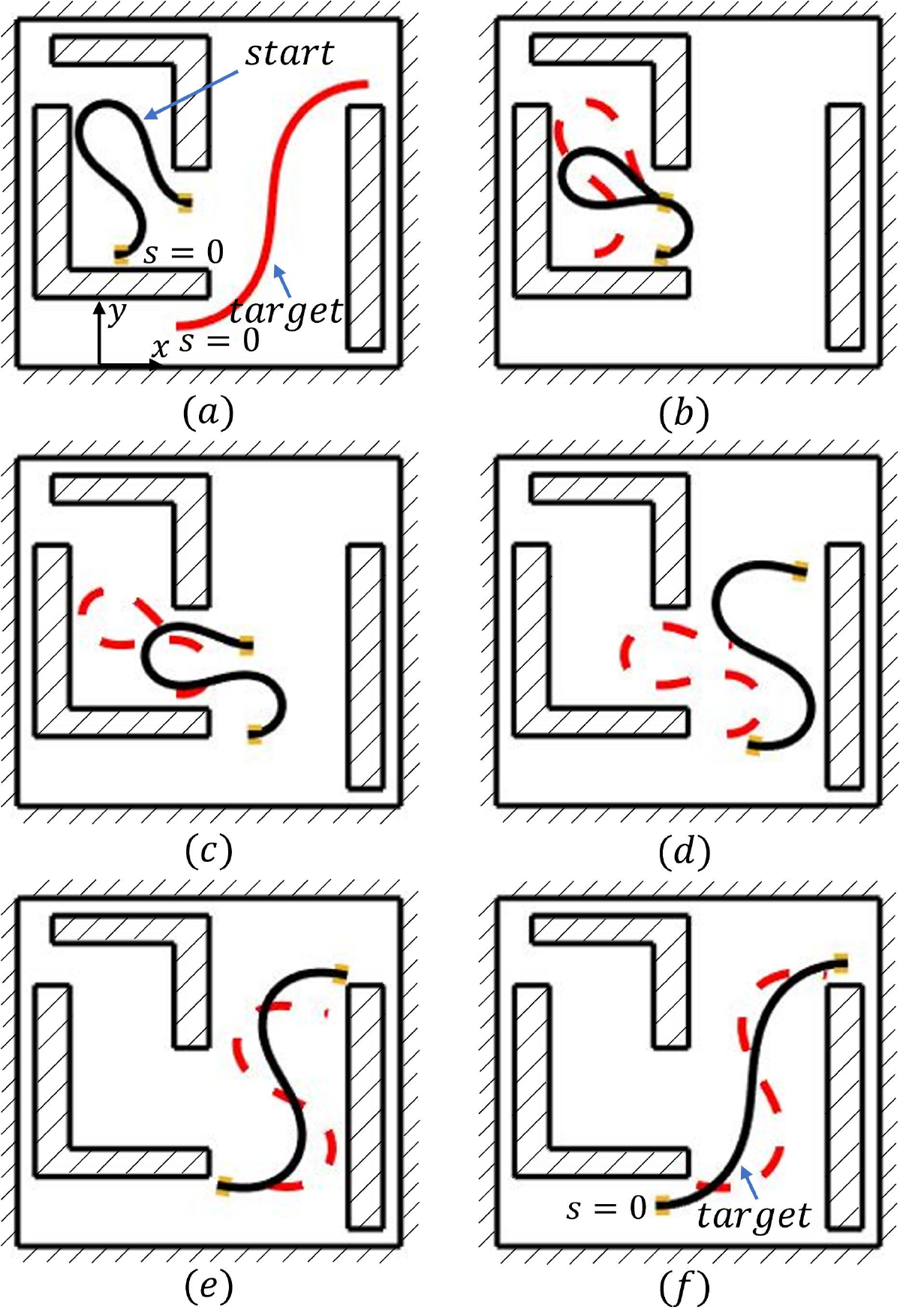}
\vspace{-.18in}
\caption{Flexible cable steering in a~confined environment containing three obstacles.
(a)~Flexible cable position and shape at the start and target.  (b)~Cable folds inward until reaching self-intersection limit while rotating in-place.
(c)~Cable slides in maximally compressed shape through narrow
opening between left obstacles. (d)~Cable rotates into vertical position.
(e)-(f)~Cable stretches outward until both endpoints reach their target positions.}
\label{sim3.fig}
\vspace{-.2in}
\end{figure}

{\bf Example 3---multiple obstacles:} The example shown in Fig.~\ref{sim3.fig} steers a~unit length flexible cable in a~confined environment having
unit $x$ and $y$ dimensions and
containing three obstacles. The cable starts at
the base frame position $(x\mbox{\small $(0)$},y\mbox{\small $(0)$},\phi(0)) \!=\! (0.03, 0.33,0)$ with elastica parameters
$(\mathrm{k},s_0,\mbox{\small $\tilde{L}$}) \!=\! (0.675,0.45 \mbox{\small $L$},\mbox{\small $L$})$. The 
start~shape~forms~full period of the elastica 
with equal endpoint tangents  (Fig.~\ref{sim3.fig}(a)). The cable target is located at the base frame position $(x\mbox{\small $(0)$},y\mbox{\small $(0)$},\phi(0)) \!=\! (0.21,0.12,0)$ with elastica parameters $(\mathrm{k},s_0,\mbox{\small $\tilde{L}$}) \!=\! (0.55,0.5 \mbox{\small $L$}, 1.37 \mbox{\small $L$})$. The 
target
shape forms $1/1.37$ of the elastica full period with
equal endpoint tangents (Fig.~\ref{sim3.fig}(a)).

Algorithm~4 computed the steering path shown in Fig.~\ref{sim3.fig} in 
$180$ seconds (see video clip). The cable first folds inward until reaching self-intersection limit while rotating in-place into horizontal position (Fig.~\ref{sim3.fig}(b)). The cable next slides in maximally compressed shape through the narrow opening between the left obstacles (Fig.~\ref{sim3.fig}(c)).  When the cable reaches the open area beyond the narrow opening it rotates into vertical position (Fig.~\ref{sim3.fig}(d)). The cable now stretches outward at both endpoints by increasing \mbox{\small $\tilde{L}$} and decreasing $\mathrm{k}$ (Fig.~\ref{sim3.fig}(e)), until both endpoints reach their target positions (Fig.~\ref{sim3.fig}(f)).~\eex

\vspace{.02in}
The environment depicted in Fig.~\ref{sim3.fig} highlights the steering scheme limitation to
sparsely spaced obstacles. Consider the confined environment of Fig.~\ref{sim4.fig} which has unit $x$ and $y$ dimensions. In Fig.~\ref{sim4.fig}(a), the gap
$d_1 \!=\! 0.29$ allows cable steering through the opening (Example~3). In Fig.~\ref{sim4.fig}(b), the gap
$d_2 \!=\! 0.23$ barely allows the cable to move through the opening in a~self-folded shape, roughly confined to a~rectangle
of dimensions  $0.15L \!\times\! 0.4L$ (Fig.~\ref{intersect.fig}). When the gap reduces to $d_3 \!=\! 0.15$, Algorithm~4 reports that a~feasible steering path no longer exists between start and target.
\vspace{-.04in}


\begin{figure}
\centering \includegraphics[width=.502\textwidth]{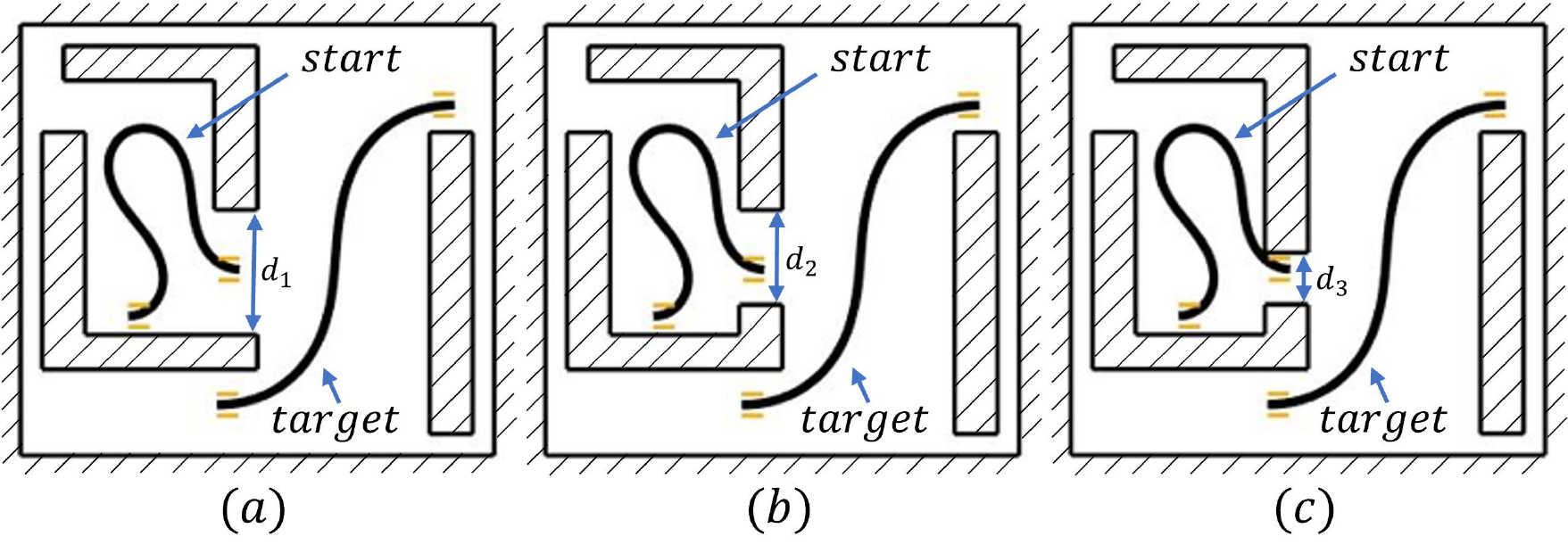}
\vspace{-.34in}
\caption{Flexible cable steering through an opening between two obstacles.
(a)~The gap $d_1 \!=\! 0.29$ allows cable steering through the opening. (b)~The gap $d_2 \!=\! 0.23$ barely allows the cable to move through the opening in a~self-folded shape. (c)~The gap $d_3 \!=\! 0.15$ disconnects the start and target when the flexible cable is not allowed to
deform against obstacles.}
\label{sim4.fig}
\vspace{-.2in}
\end{figure}

\section{\bf Conclusion}
\vspace{-.02in}

\noindent The paper considered flexible cable steering by two robot hands
that impose endpoint positions and tangents in two-dimensions.
The robot hands steer the flexible cable while maintaining equal endpoint tangents, thus reducing the planning problem into
five  configuration variables.
The paper described the Euler's elastica solutions for the flexible cable equilibrium shapes.
The elastica solutions were then used to develop  several tools that were incorporated into a~flexible cable steering scheme among sparsely spaced obstacles.

The paper first developed analytic equations that determine the cable elastica parameters in terms of the endpoint positions and tangents imposed by the robot hands. The paper next described  the range of the elastica modulus parameter that
ensures non self-intersection. Using the notion of conjugate point, the paper established a~criterion for the flexible cable
stability:
all cable shapes that form less
than full period of the elastica (midpoint inflection point) or full period of the elastica (two internal inflection points) are automatically stable. Last, the paper described an approximation of the
flexible cable
equilibrium shapes by convex quadratic arcs that allow efficient collision check against obstacles during steering.

The new tools were incorporated into a~steering scheme for flexible cables
in two-dimensions by two robot hands.
The steering scheme has been implemented and
execution examples demonstrated 
the flexible cable motion between start and target while
avoiding self-intersection and collision with sparsely placed obstacles in the environment.


Future research will extend this paper in two important ways. The first extension
will consider flexible cable steering by two robot hands under gravity in two-dimensions. The flexible cable
equilibrium shapes now  minimize the cable elastic and gravitational energies
\vspace{-.08in}
\[
\Es =  \int_0^L \! \left( \hlf \mbox{\small $ EI$} \!\cdot\! \kp^2(s) + \rho g \!\cdot\! y(s) \right) ds
\vspace{-.08in}
\]

\noindent where $\kp(s)$ is the cable curvature, $y(s)$ is the cable vertical coordinate and \mbox{\small $L$} is
the cable length. The cable stiffness, {\small $EI$}, and the cable mass density, $\rho$, are~known~\mbox{parameters}~such that $m \!=\! \rho \mbox{\small $L$}$ is the cable mass.
When $\rho$~is ~small~\mbox{relative}~to~{\small $EI$}, the elastica solutions characterize the cable equilibrium shapes. When
$\rho$ is large relative to {\small $EI$}, the cable behaves like a~chain that attains {\em catenary} equilibrium shapes under gravity.
A promising approach would be to blend the two analytic solutions into reasonably accurate approximation of the true equilibrium shape. The blending technique will then be augmented with the tools developed in this paper into flexible cable steering  scheme under gravity.

A more open ended extension will allow the flexible cable to deform
against obstacles during steering from start to target. When friction is sufficiently high at the cable's contacts with obstacles,
the flexible cable will exhibit rolling-like behavior with full segments of the cable contacting obstacles.
A promising approach would be to allow the flexible cable to
deform against a~single wall when steered by two robot hands, while avoiding all other obstacles.
Predicting the cable equilibrium shapes will require extension of this paper optimal control formulation to include
{\em pure state constraints} that capture contacts with obstacles, then use the elastica solutions to model
the contact free portions of the cable with endpoints now imposed by the robot hands and the environment. The flexible cable
steering scheme will likely require extra state variables that will measure the cable contacts against the environment.

The extensions will eventually allow steering {\em flexible strips} by two robot hands in {\small 3-D} under the influence of gravity.
Suitably designed robot grippers can impose {\em line contacts} at the flexible strip opposing ends.
By maintaining parallel line contacts during {\small 3-D} manipulation, flexible strip steering by two robot hands would
become a~natural generalization of the steering scheme considered in this paper.
\vspace{-.06in}



\section*{\bf Acknowledgment}
\noindent Prof. Segev from Ben-Gurion University provided highly appreciated input during the preparation of this paper. 

\section*{\bf Appendix A: Elliptic Function Details}

\noindent This appendix verifies the formulas for the flexible cable curvature $\Kg(s)$ and tangent $\phi(s)$ given in Section~II.
Let us first verify that the flexible cable curvature
 \begin{equation} \label{eq.app_curv}
\kp(s) \!=\! - 2 \mbox{\small $\sqrt{\Lgs}$} \mathrm{k} \!\cdot \! \mbox{\sf $cn$}( \mbox{\small $\sqrt{\Lgs}$} (s_0 \!+\! s), \mathrm{k} )
 \end{equation}
\noindent solves the elastic energy extremum conditions of Eqs.~\eqref{eq:AdjointSystem}-\eqref{eq.u_cond}.
The two extremum conditions are equivalent to the single condition $H(s) \!=\! H^*$ for $s \!\in\! [0,\mbox{\small $L$}]$, that can be written as
\begin{equation} \label{eq.app1}
 \hlf \mbox{\small $EI$} \cdot \kp^2(s) + \mbox{\small $H$}^* = \Lgs_r \cdot \cos( \phi\mbox{\small $(s)$} \!-\! \phi_0 ) \rtxt{$s\in [0,L]$.}
\end{equation}
\noindent  Eq.~\eqref{eq.app1} is next converted into an~equivalent condition that involves only the cable curvature, $\kp(s)$.
Taking the derivative of Eq.~\eqref{eq.app1} with respect to~$s$ gives
\begin{equation} \label{eq.app2}
   \mbox{\small $EI$} \cdot \frac{d}{ds} \kp(s)  = - \Lgs_r \cdot \sin( \phi\mbox{\small $(s)$} \!-\! \phi_0 )  \rtxt{$s\in [0,L]$.}
 \end{equation}
\noindent  where we used the chain rule and then canceled the common factor~$\kp(s) \!=\! \frac{d}{ds} \phi(s)$.
 Eqs \eqref{eq.app1}-\eqref{eq.app2} combine into a~single implicit differential equation in $\kp(s)$
 \begin{equation} \label{eq.app3}
 \big( \hlf \mbox{\small $EI$} \cdot \kp^2(s) + \mbox{\small $H$}^* \big)^2 + \big( \mbox{\small $EI$} \cdot \frac{d}{ds} \kp(s) \big)^2 = \Lgs^2_r.
 \end{equation}
\noindent We need an expression~for~$\mbox{\small $H$}^*$~in~this~\mbox{equation.}~Since~\mbox{\small $\mbox{\small $H$}(s) \!=\! \mbox{\small $H$}^*$} for $s \!\in\! [0,\mbox{\small $L$}]$,
the value of  $\mbox{\small $H$}^*$  can be determined at any point along  the flexible cable.  At the zero curvature points where $\kp(s^*) \!=\! 0$,  Eq.~\eqref{eq.Hstar} for $\mbox{\small $H$} (s) \!=\! \mbox{\small $H$}^*$ gives
{\small 
 \begin{equation} \label{eq.H*}
 H^*  \!=\! \Lg_r \cdot ( \cos\phi\mbox{\small $(s^*)$}\cos\phi_0 \!+\! \sin\phi\mbox{\small $(s^*)$}\sin\phi_0 )
 \!=\! \Lg_r  (1 \!-\! 2\mathrm{k}^2),
 \end{equation}
 }
\noindent \hspace{-.9em} where we used the relation $ \cos(\phi(s^*) \!-\! \phi_0) \!=\! 1 \!-\! 2\mathrm{k}^2$ based on Eq.~\eqref{eq.k} for $\mathrm{k}$. Substituting for $\mbox{\small $H$}^*$ in Eq.~\eqref{eq.app3} gives
 \begin{equation} \label{ext_cond}
 \big( \hlf \mbox{\small $EI$} \cdot \kp^2(s) + \Lgs_r \cdot (1 \!-\! 2\mathrm{k}^2)  \big)^2 + \big( \mbox{\small $EI$} \cdot \frac{d}{ds} \kp(s) \big)^2 = \Lgs^2_r.
 \end{equation}
\noindent  The derivative of $\mbox{\sf $cn$}(u, \mathrm{k})$
is given by the rule: $\tfrac{d}{du}  \mbox{\sf $cn$}(u, \mathrm{k}) = -  \mbox{\sf $sn$}(u, \mathrm{k}) \cdot  \mbox{\sf $dn$}(u, k)$,  where $\mbox{\sf $sn$}(u, \mathrm{k} )$ is the elliptic sine function
 and $\mbox{\sf $dn$}(u, \mathrm{k} )$ is the {\em elliptic delta amplitude function}~\cite{ellip_book}.
 Applying this derivative to $\kp(s)$ in Eq.~\eqref{eq.app_curv} gives
 \begin{equation} \label{eq.curv_deriv}
  \mbox{\small $\frac{d}{ds}$} \kp(s) =
  2 \Lgs \cdot \mbox{\sf $sn$}( \mbox{\small $\sqrt{\Lgs}$} (s_0 \!+\! s), \mathrm{k} ) \cdot \mbox{\sf $dn$}( \mbox{\small $\sqrt{\Lgs}$}(s_0 \!+\! s), \mathrm{k} ).
 \end{equation}
\noindent where we used the chain rule.
 Substituting for $\kp(s)$ and $\tfrac{d}{ds} \kp(s)$ according to~\eqref{eq.app_curv} and~\eqref{eq.curv_deriv} in Eq.~\eqref{ext_cond}
 gives the equation
 \begin{equation}  \label{eq.app4}
 \barr
 & \big(2  \mathrm{k}^2 \cdot \mbox{\sf $cn$}^2(\mbox{\small $\sqrt{\Lg}$} (s_0 \!+\! s), \mathrm{k} ) + (1-2\mathrm{k}^2)  \big)^2 +\\
 & 4  \mathrm{k}^2  \cdot \mbox{\sf $sn$}^2( \mbox{\small $\sqrt{\Lg}$} (s_0 \!+\! s), \mathrm{k} ) \cdot  \mbox{\sf $dn$}^2( \mbox{\small $\sqrt{\Lg}$}(s_0 \!+\!  s), \mathrm{k} ) = 1
 \earr
 \end{equation}
\noindent  where we substituted $\Lg \!=\! \Lg_r / \mbox{\footnotesize $EI$}$ then canceled the resulting common factor $\Lg^2_r$.
The first summand in Eq.~\eqref{eq.app4} can be simplified as $( 1 \!-\! 2 \mathrm{k}^2 \mbox{\sf $sn$}^2(\mbox{\small $\sqrt{\Lg}$} (s_0 \!+\! s), \mathrm{k} ) )^2 $.
The function $\mbox{\sf $dn$}(u, \mathrm{k} )$ satisfies the relation $\mbox{\sf $dn$}^2(u, \mathrm{k}) \!=\! 1 \!-\!  \mathrm{k}^2 \!\cdot\! \mbox{\sf $sn$}^2(u, k)$. Substituting for $\mbox{\sf $dn$}^2(u, \mathrm{k})$ in Eq.~\eqref{eq.app4} and using the simplified summand gives the relation
 \[
 \barr
 & 1 - 4 \mathrm{k}^2  \cdot \mbox{\sf $sn$}^2(\mbox{\small $\sqrt{\Lg}$} (s_0 \!+\! s), \mathrm{k} ) )\big( 1 -  \mathrm{k}^2 \mbox{\sf $sn$}^2(\mbox{\small $\sqrt{\Lg}$} (s_0 \!+\! s), \mathrm{k} ) \big) \\[4pt]
 & + 4  \mathrm{k}^2  \cdot \mbox{\sf $sn$}^2( \mbox{\small $\sqrt{\Lg}$} (s_0 \!+\! s), \mathrm{k} ) \big( 1 - \mathrm{k}^2 \mbox{\sf $sn$}^2( \mbox{\small $\sqrt{\Lg}$} (s_0 \!+\! s), \mathrm{k} ) \big) = 1
 \earr
 \]
\noindent which validates formula \eqref{eq.app_curv} for the flexible cable curvature.\\[2pt]
\indent Let us next verify the flexible cable tangent formula:
 \begin{equation} \label{eq.app5}
\sin\big( \tfrac{1}{2}(\phi(s) - \phi_0) \big) = - \mathrm{k} \cdot \mbox{\sf $sn$}\big( \mbox{\small $\sqrt{\Lg}$} \cdot (s_0 \!+\!  s), \mathrm{k} \big).
\end{equation}
The derivative of $\mbox{\sf $sn$}( u, \mathrm{k} )$ is given by the rule:
$\tfrac{d}{du}  \mbox{\sf $sn$}(u, \mathrm{k}) \!=\! \mbox{\sf $cn$}(u, \mathrm{k}) \!\cdot\!  \mbox{\sf $dn$}(u,  \mathrm{k})$~(\cite{ellip_book}).
Using the  relation $\mbox{\sf $dn$}^2(u, \mathrm{k}) \!=\! 1 \!-\!  \mathrm{k}^2 \!\cdot\! \mbox{\sf $sn$}^2(u,  \mathrm{k})$ and the fact that $\mbox{\sf $dn$}(u, \mathrm{k}) \!\geq \! 0$,
the derivative rule for $\mbox{\sf $sn$}(u, \mathrm{k})$ gives the elliptic cosine expression
{\small
\begin{equation} \label{eq.cn_deriv}
\mbox{\sf $cn$}( u, \mathrm{k} ) \!=\! \mbox{\small  $\frac{1}{\sqrt{1 - \mathrm{k}^2\mbox{\sf $sn$}^2(u, \mathrm{k})}} \frac{d}{du} \mbox{\sf $sn$}( u, \mathrm{k} )
\!=\! \frac{1}{ \mathrm{k}} \frac{d}{du} \big(  \sin^{-1}( \mathrm{k} \!\cdot\!  \mbox{\sf $sn$}( u, \mathrm{k} ) ) \big)$}
\end{equation}
}
\noindent \hspace{-.75em} where we used the chain rule.
Recall now that $\kappa(s) \!=\! \tfrac{d}{ds} \phi(s)$. Hen\-ce, taking the integral of both sides of Eq.~\eqref{eq.cn_deriv} using \eqref{eq.app_curv} gives the flexible cable tangent 
\[
\phi(s) = \phi(0) + \! \int_0^s \!\! \kappa(t) dt = \phi(0) - 2   \mbox{\small $\sqrt{\Lgs}$} \mathrm{k} \! \int_0^s \! \! \mbox{\sf $cn$}\big( \mbox{\small $\sqrt{\Lgs}$} (s_0 \!+\! t), \mathrm{k} \big) dt .
\]
\noindent Now set the integration variable as $\mu \!=\! \mbox{\small $\sqrt{\Lgs}$} (s_0 \!+\! t)$ so that $d\mu \!=\! \mbox{\small $\sqrt{\Lgs}$} \, dt$. The integration limits
$t \!=\! 0$ and $t \!=\! s$ become $\mu \!=\! \mbox{\small $\sqrt{\Lgs}$} s_0$ and $\mu \!=\! \mbox{\small $\sqrt{\Lgs}$} (s_0 \!+\! s)$.
The expression for $\phi(s)$ now takes the form
\[
\phi(s) =  \phi(0) - 2   \mathrm{k} \int_{\mbox{\scriptsize $\sqrt{\Lgs}$} s_0}^{\mbox{\scriptsize $\sqrt{\Lgs}$} (s_0 \!+\! s)} \!\! \mbox{\sf $cn$}( \mu, \mathrm{k} ) d\mu .
\]
\noindent Substituting for  $\mbox{\sf $cn$}( \mu, \mathrm{k} )$ according to Eq.~\eqref{eq.cn_deriv} gives
\begin{equation} \label{eq.app_phi}
\barr
\phi(s) =&  \phi(0) - 2   \big(
 \sin^{-1}( \mathrm{k} \!\cdot\!  \mbox{\sf $sn$}( \mbox{\small $\sqrt{\Lgs}$} (s_0 \!+\! s), \mathrm{k} ) ) \\[2pt]
 & - \sin^{-1}( \mathrm{k} \!\cdot\!  \mbox{\sf $sn$}( \mbox{\small $\sqrt{\Lgs}$} s_0, \mathrm{k} ) )
 \big)
 \earr
\end{equation}
\noindent where $\mathrm{k}$ has been canceled out. The first two summands in \eqref{eq.app_phi} describe the elastica axis angle $\phi_0$.
As seen in Fig.~\ref{el_extra.fig}, $\phi_0$ forms the elastica tangent angle at its highest point $s \! =\! -s_0$.
Substituting $s \! =\! -s_0$ in the expression for $\phi(s)$ and noting that $\mbox{\sf $sn$}( 0, \mathrm{k} ) \!=\! 0$
gives the elastica axis angle
\begin{equation} \label{eq.phi0}
\phi_0 = \phi(\mbox{\small $0$}) + 2\sin^{-1}\big(\mathrm{k}\cdot \mbox{\sf $sn$}( \mbox{\small $\sqrt{\Lg}$} \!\cdot\! s_0, \mathrm{k}) \big) .
\end{equation}
The expression for $\phi(s)$ thus takes the form
 \[
\mbox{\small $\frac{1}{2}$} ( \phi(s) -  \phi_0 ) = - \sin^{-1}( \mathrm{k} \!\cdot\!  \mbox{\sf $sn$}\big( \mbox{\small $\sqrt{\Lgs}$} (s_0 \!+\! s), \mathrm{k} \big)
 \]
\noindent which validates formula \eqref{eq.app5} for the flexible cable tangent.
\section*{\bf Appendix B:  Relation of Elastica Parameters to Flexible Cable Endpoint Positions and Tangents}

\noindent This appendix formulates three equations that relate
the flexible cable elastica parameters $\Lg$, $\mathrm{k}$ and $s_0$
to the flexible cable relative endpoint positions and tangents, used
by the robot grippers during steering.
Consider the flexible cable $(x,y)$ coordinates of Eq.~\eqref{eq.xy}. This expression uses the elastica amplitude parameter {\small $A$}.
To express the amplitude in terms of the elastica parameters,
consider that the elastica highest point with respect to its axis occurs at $s \!=\! -s_0$, where
$s \!+\! s_0 \!=\! 0$ and $\mbox{\sf $cn$}( 0,\mathrm{k}) \!=\! 1$ (Fig.~\ref{el_extra.fig}).
 From Appendix~A,
substituting $H^* \!=\! \Lg_r  (1 \!-\! 2\mathrm{k}^2)$ then
evaluating $H(s)$ at the elastica highest point
$s \!=\! -s_0$ using the relation $\Lg \!=\! \Lg_r / \mbox{\small $EI$}$
gives the elastica amplitude
\begin{equation} \label{eq.ms_0}
\mbox{\small $
\Lg_r  -  \hlf \mbox{\small $EI$} \cdot (\Lg  \mbox{\small $A$})^2
\!=\! \Lg_r  (1 \!-\! 2\mathrm{k}^2)  \,\, \Rightarrow \,\,
\mbox{\small $A$} \!=\! \frac{2\mathrm{k}}{\mbox{\footnotesize $\sqrt{\Lgs}$}} \, . $}
\end{equation}
\noindent
The flexible cable $(x,y)$ coordinates of Eq.~\eqref{eq.xy} evaluated at $s \!=\! \mbox{\small $L$}$  give two equations that relate the elastica parameters $\Lg$, $\mathrm{k}$ and $s_0$ to the flexible cable relative endpoint positions:
{\small
\begin{equation} \label{eq.xy_L}
\begin{aligned}
  &\begin{pmatrix} x(L) \\ y(L) \end{pmatrix} - \begin{pmatrix} x(0) \\ y(0) \end{pmatrix} \\
  &\quad = \frac{1}{\lambda_r} \begin{bmatrix} \cos\phi_0 & \sin\phi_0 \\ \sin\phi_0 & -\cos\phi_0 \end{bmatrix} \\
  &\qquad \cdot \begin{pmatrix} \int_0^L \frac{1}{2} EI \cdot \kappa^2(s) ds + \lambda_r (1 - 2k^2) \cdot L \\ EI \cdot ( \kappa(L) - \kappa(0) ) \end{pmatrix} \\
  &\quad = \frac{1}{\lambda} \begin{bmatrix} \cos\phi_0 & \sin\phi_0 \\ \sin\phi_0 & -\cos\phi_0 \end{bmatrix} \\
  &\qquad \cdot \begin{pmatrix} \int_0^L \frac{1}{2} \kappa^2(s) ds + \lambda (1 - 2k^2) \cdot L \\ \kappa(L) - \kappa(0) \end{pmatrix}
\end{aligned}
\end{equation}
}
\noindent \hspace{-1.4em} where we substituted $\Lg \!=\! \Lg_r / \mbox{\small $EI$}$ and $\! H^*   \!=\! \Lg_r  (1 \!-\! 2\mathrm{k}^2) \!$
fr\-om Appendix~A.
The endpoint curvatures in Eq.~\eqref{eq.xy_L} are
$\kp(0) \!=\! -  2\mathrm{k}\mbox{\small $\sqrt{\Lg}$}   \cdot  \mbox{\sf $cn$}( \mbox{\small $\sqrt{\Lg}$} \cdot s_0,\mathrm{k})$
and $\kp(\mbox{\small $L$}) \!=\! -  2\mathrm{k}\mbox{\small $\sqrt{\Lg}$}    \! \cdot\!  \mbox{\sf $cn$}( \mbox{\small $\sqrt{\Lg}$} \!\cdot \! (s_0 \!+\! \mbox{\small $L$}),\mathrm{k})$, where we substituted for {\small $A$} and used Eq.~\eqref{eq.kp} for $\kappa(s)$. The elastica axis angle $\phi_0$ in  Eq.~\eqref{eq.xy_L} is specified by Eq.~\eqref{eq.phi0}.
\noindent The third equation relates the elastica parameters to the relative endpoint tangent angles. This relation is obtained by evaluating Eq.~\eqref{eq.phi} for $\phi(s)$ at the flexible cable endpoints $s \!=\! 0$ and $s \!=\! \mbox{\small $L$}$:
{\small
\begin{equation} \label{eq.phi_0L}
\barr
& \tfrac{1}{2} \big( \phi(\mbox{\small $L$}) - \phi(0) \big) = \\[6pt]
&  \sin^{-1}\big(\mathrm{k}\!\cdot\! \mbox{\sf $sn$}( \mbox{\small $\sqrt{\Lgs}$} \!\cdot\! s_0,\mathrm{k})\big)
- \sin^{-1}\big(\mathrm{k}\!\cdot\! \mbox{\sf $sn$}( \mbox{\small $\sqrt{\Lgs}$} \!\cdot\! (s_0 \!+\! L), \mathrm{k}) \big).
\earr
\end{equation}
}
\noindent \hspace{-1em} The three equations \eqref{eq.xy_L}-\eqref{eq.phi_0L} can be used to map the  elastica parameters {\small $\Lg$}, $\mathrm{k}$ and~$s_0$ to the flexible cable relative endpoint positions and tangents,
used by the robot grippers during steering.

Finally consider the subsets of elastica parameters that describe a~flexible cable held with equal endpoint tangents.
When a~flexible cable is held with equal endpoint tangents, $\phi(\mbox{\small $L$}) \!-\! \phi(0) \! = \! 0$,
Eq.~\eqref{eq.phi_0L} gives the relation
\[
\mbox{\sf $sn$}( \mbox{\small $\sqrt{\Lgs}$} \!\cdot\! s_0,\mathrm{k})
= \mbox{\sf $sn$}( \mbox{\small $\sqrt{\Lgs}$} \!\cdot\! (s_0 \!+\! L), \mathrm{k} ).
\]
\noindent There are two types of flexible cable equilibrium shapes with equal endpoint tangents. Either $L \!=\! \tilde{L}$ (two inflection points), or $L \!<\! \tilde{L}$ (one inflection point).
The $L \!=\! \tilde{L}$ shapes have equal endpoint tangents since
$\mbox{\sf $sn$}( \mbox{\small $\sqrt{\Lgs}$} \!\cdot\! s_0,\mathrm{k})
\!=\! \mbox{\sf $sn$}( \mbox{\small $\sqrt{\Lgs}$} \!\cdot\! (s_0 \!+\! \Tilde{L}), \mathrm{k})$ by the $\Tilde{L}$-periodicity
of the elliptic sine function. These shapes correspond to the horizontal cells on the bottom of Fig.~\ref{cells.fig}(a).
The $L \!<\! \tilde{L}$  shapes have two possible phase parameter values when held with equal endpoint tangents.
The first phase parameter $s_0 \!=\! \tfrac{3\Tilde{L}}{4} \!+\! \tfrac{\Tilde{L}-L}{2}$ gives equal endpoint tan\-gents due to symmetry of
the elastica about its first inflection point at $ s \!=\! \tfrac{\Tilde{L}}{4}$: $\mbox{\sf $sn$}( \tfrac{\Tilde{L}}{4} \!-\! \tfrac{L}{2}, \mathrm{k}) \!=\!
\mbox{\sf $sn$}( \tfrac{\Tilde{L}}{4} \!+\! \tfrac{L}{2},\mathrm{k})$.
This phase parameter gives the left diagonal cell $\tfrac{3\Tilde{L}}{4} \!-\! s_0 \!=\!  \tfrac{L}{2}$ that runs
parallel to the $\mathrm{k}$-axis in Fig.~\ref{cells.fig}(a).
The second phase parameter $s_0 \!=\! \tfrac{\Tilde{L}}{4} \!+\! \tfrac{\Tilde{L}-L}{2}$ gives equal endpoint tangents due to symmetry of
the elastica about its second inflection point at $ s \!=\! \tfrac{3\Tilde{L}}{4}$:
$\mbox{\sf $sn$}( \tfrac{3\Tilde{L}}{4} \!-\! \tfrac{L}{2}, \mathrm{k}) \!=\!
\mbox{\sf $sn$}( \tfrac{3\Tilde{L}}{4} \!+\! \tfrac{L}{2},\mathrm{k})$.
This phase parameter gives the right diagonal cell $ \tfrac{3\Tilde{L}}{4} \!-\! s_0 \!=\!  \tfrac{L}{2}$ that runs
parallel to the $\mathrm{k}$-axis in Fig.~\ref{cells.fig}(a).



\section*{\bf Appendix C: Self-Intersection of the Inflectional Elastica}

\noindent In this appendix, we first consider the canonical elastica shape without adding a phase $s_0$, translation $(x(0),y(0))$, or rotation $\phi(0)$. That is, an infinite elastica that depends on two parameters, the elliptic modulus parameter $k$ or $\mu$ and the co-state parameter $\lambda$.

We now write the equations of the elastica curve $(x(s),y(s))$ according to the parametrization in~\cite{djondjorov}:
\begin{equation} \label{eq.cablexy}
    \begin{pmatrix}
        x(s) \\
        y(s)
    \end{pmatrix} \!=\!  \begin{pmatrix}
        \frac{2}{\sqrt{\lambda}}E\bigg{(}\mathrm{am}(\sqrt{\lambda}s,\mathrm{k})\bigg{)}-s \\
        A\cdot\mathrm{cn}(\sqrt{\lambda}s,\mathrm{k})
    \end{pmatrix}, \qquad s\in[0,L]
\end{equation}
where the following relations hold:
\begin{equation*}
\mathrm{k}=\sqrt{\frac{1-\mu}{2}},\qquad a=\sqrt{\frac{2(1-\mu)}{\lambda}},\qquad \lambda=\left(\frac{4\,\mathrm{K}(\mathrm{k})}{\tilde{L}}\right)^{\!2}.
\end{equation*}

In~\cite{djondjorov}, there is an expression for the height $y_{fold}$, which corresponds to what we referred to as {\em fold points}. From this expression, it is possible to derive an expression for $s_{fold}$, the length along the cable where the tangent angle with respect to the elastica axis satisfies $\phi(s_{fold}) \pm 90^{\circ}$: 
\begin{equation}
y_{fold}=\sqrt{-\frac{2\mu}{\lambda}} \ \ \Longrightarrow \ \ s_{\mathrm{fold}}=\frac{1}{\sqrt{\lambda}}\,F\!\Big(\cos^{-1}\!\sqrt{\frac{\mu}{\mu-1}},\,\mathrm{k}\Big).
\end{equation}
Additionally, we note that the points where the elastica curvature $\kappa(s)$ vanishes are
\begin{equation*}
s(\kappa=0)=\frac{\tilde{L}}{4}+n\,\frac{\tilde{L}}{2},\qquad n=1,2,\dots .
\end{equation*}

We also recall that $\mathrm{k}_{int}$ is the maximum elliptic modulus parameter for which there is no self-intersection (there exist aself-tangency point), $\mathrm{k}_{c}$ is the elliptic modulus parameter corresponding to the figure-eight shape. Thus, we divide the elastica shape into three regions:
\begin{align*}
\textbf{Region 1: } & \mathrm{k}\in[0,\mathrm{k}_{\max}),\\
\textbf{Region 2: } & \mathrm{k}\in[k_{\max},\mathrm{k}_c),\\
\textbf{Region 3: } & \mathrm{k}\in[\mathrm{k}_c,1).
\end{align*}

\section*{Region 1}
In this region, there is no self-intersection for any phase $s_0$ and any full period $\Tilde{L}$, as shown in section~\ref{sec.self_intersect}.

\section*{Region 2}
For convenience, consider a full period of the elastica in the domain $s \in \left[-\tfrac{\tilde{L}}{2}, \tfrac{\tilde{L}}{2}\right]$. We look for the first $s \geq 0$ self-intersection point, denoted as $s_{int}$.

\begin{figure}[H]\centering
\includegraphics[width=\linewidth]{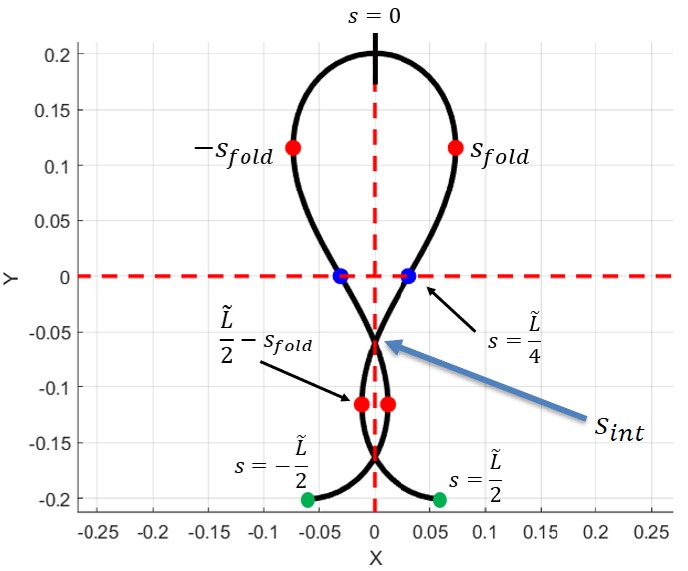}
\caption{A full period of the elastica in Region 2 of $\mathrm{k}$ values.}
\end{figure}

From Figure~1, several points can be noted that help us understand the self-intersection point of $k$ values:

\begin{itemize}
    \item The first self-intersection point in Region~2 always satisfies 
    $x(s_{int}) = 0$. Therefore, Eq.~(1) equals zero---this is the implicit equation we need to solve to find the intersection point $s_{int}$.
    
    \item In Region~2, the self-intersection point $s_{int}$ will always lie in the interval
    $\left[\tfrac{\tilde{L}}{4}, \tfrac{\tilde{L}}{2}  - s_{fold}\right]$.
    
    \item In Region~2, it always holds that 
    $x\!\left(\tfrac{\tilde{L}}{4}\right) > 0$
    and 
    $x\!\left(\tfrac{\tilde{L}}{2} - s_{fold}\right) < 0$, 
    so the function $x(s)$ of Eq.~(1) has exactly {\em one root} in the interval 
    $\left[\tfrac{\tilde{L}}{4}, \tfrac{\tilde{L}}{2} - s_{fold}\right]$. 
    Therefore, the {\em bisection method} can be used to find $s_{int}$.
    
    \item In Region~2, the first self-intersection point is always in the negative part of the $y$-axis (Fig.~1).
    Note that the second self-intersection point always lies in the interval 
    $\left[\tfrac{\tilde{L}}{2} - s_{fold}, \tfrac{\tilde{L}}{2}\right]$, 
    but this point is not needed to detect self-intersection.
\end{itemize}

Additionally, two special cases are worth noting:
\begin{itemize}
    \item When $\mathrm{k} = \mathrm{k}_{max}$, it holds that 
    $s_{int} = \tfrac{\tilde{L}}{2} - s_{fold}$.
    
    \item When $\mathrm{k} = \mathrm{k}_c$, it holds that 
    $s_{int} = \tfrac{\tilde{L}}{4}$, 
    and the self-intersection point of the canonical elastica occurs at the origin.
\end{itemize}

After presenting Region~3, we will describe an algorithm that examines all possible cases for self-intersection, based on an analysis I performed, which involved tracking the increase of $s$ and searching for the next self-intersection points.

\section*{Region 3}
In this region as well, we consider a full period of the canonical elastica in the domain $s \in \left[-\tfrac{\tilde{L}}{2}, \tfrac{\tilde{L}}{2}\right]$
We look for the first positive self-intersection point, denoted by $s_{int}$.

\begin{figure}[H]\centering
\includegraphics[width=\linewidth]{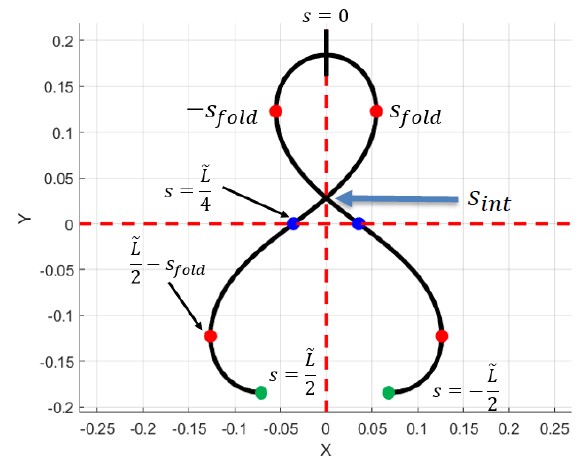}
\caption{A full period of the elastica in Region~3 of $\mathrm{k}$ values.}
\end{figure}
\textbf{Important note:} This paper do not consider Region~3 of $\mathrm{k}$-values for local stability reasons. 
However, locally stable shapes can exist in Region~3. 
To determine whether a shape is locally stable, one must check the determinant of the {\em endpoint mapping Jacobian}~\cite{sachkov_conjugate}.

From Figure~2, several points can be observed that help us understand the self-intersection point in Region~3 of $\mathrm{k}$-values:
\begin{itemize}
    \item The first self-intersection point in Region~3 always satisfies 
    $x(s_{int}) = 0$. Therefore, Eq.~(1) equals zero---this is the implicit equation we need to solve to find the intersection point $s_{int}$.
    
    \item In Region~3, the self-intersection point will always lie within the interval $s \in \left[s_{fold}, \tfrac{\tilde{L}}{4}\right]$.
    
    \item In Region~3, it always holds that 
    $x(s_{fold}) > 0$ and $x\!\left(\tfrac{\tilde{L}}{4}\right) < 0$, 
    so the function $x(s)$ of Eq.~(1) has exactly {\em one root} in the interval 
    $\left[ s_{fold}, \tfrac{\tilde{L}}{4} \right]$. 
    Therefore, the {\em bisection method} can be used to find $s_{int}$.
    
    \item In Region~3, the self-intersection point always lies in the positive part of the $y$-axis (Fig.~2).
\end{itemize}

\section*{Self-Collision Check Algorithm}
First, let us specify how we compute $s_{int}$. 
For this, we use the {\em bisection method}, denoted as:
\[
s_{int} = \text{Bisection}(s_{min}, s_{max}, \epsilon),
\]
where $s_{min}$ is the lower bound of the interval where Eq.~(1) crosses zero, 
$s_{max}$ is the upper bound, and $\epsilon$ represents the tolerance for the function to be considered zero.

The following algorithm illustrates the computation of $s_{int}$.

{\fontsize{10}{10}\selectfont
\begin{algorithm}{}
\caption{\bf GetSint}
\begin{spacing}{0.95}
\begin{algorithmic}[0]
\renewcommand{\thealgorithm}{}

\Statex {\bf Input:} elastica parameters $\mathrm{k}$, $\tilde{L}$, tolerance $\epsilon$.
\Statex {\bf Data structures:} fold location $s_{\text{fold}}$, intersection $s_{\text{int}}$.
\Statex {\bf Compute:} $s_{\text{fold}}$.

\Statex {\bf Cases:}
\begin{enumerate}{}
  \item If $\mathrm{k} \leq \mathrm{k}_{\max}$, set $s_{\text{int}} = \emptyset$.
  \item If $\mathrm{k} \in [\mathrm{k}_{max},\, \mathrm{k}_c]$, set 
        $s_{\text{int}} = \text{Bisection}\!\big(\tfrac{\tilde{L}}{4},\, \tfrac{\tilde{L}}{2}-s_{\text{fold}},\, \epsilon\big)$.
  \item If $\mathrm{k} \in [k_c,\, 1)$, set 
        $s_{\text{int}} = \text{Bisection}\!\big(s_{\text{fold}},\, \tfrac{\tilde{L}}{4},\, \epsilon\big)$.
\end{enumerate}

\Statex {\bf Output:} $s_{\text{int}}$.
\end{algorithmic}
\end{spacing}
\end{algorithm}
}

Without loss of generality, let the phase parameter $s_{0}$ be confined to the interval $s_{0} \in [0, \tilde{L})$
and let the full period length parameter $\tilde{L}$ be limited to at most one full period. Consider two periods of the elastica in Region~2 and Region~3. 
In the corresponding figures, we mark the relevant self-intersection points, and then express these points as functions of $s_{int}$ and $\tilde{L}$. 
Recall that $s_{int}$ is determined by the {\em bisection method}.

\begin{figure}[H]\centering
\includegraphics[width=\linewidth]{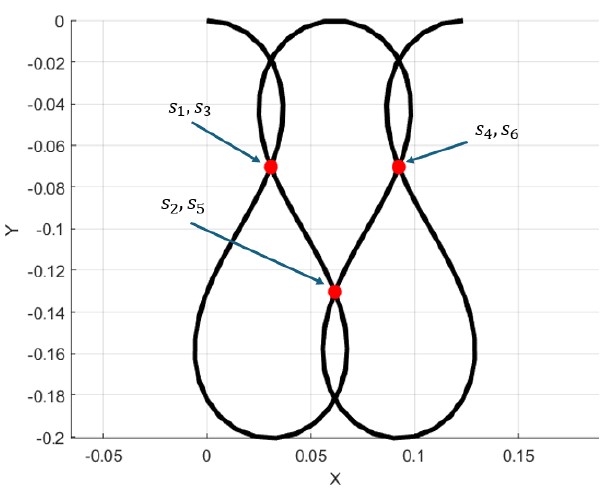}
\caption{\centering Two full periods of the elastica ($\tilde{L}=L/2$) and the six self-intersection points along the periods for Region~2 of $\mathrm{k}$ values.}

\end{figure}
For Region~2 of $\mathrm{k}$ values, Figure~3 shows the six possible self-intersection points, 
where some points satisfy $(x(s_i), y(s_i)) = (x(s_j), y(s_j))$. Based on this, we perform the self-intersection analysis.

For Region~2 of $\mathrm{k}$ values, the values $s_1 < s_2 < s_3 < s_4 < s_5 < s_6$ are computed as:
\begin{equation*}
s_1 = \tfrac{\tilde{L}}{2} - s_{int}, \qquad
s_2 = \tilde{L} - s_{int}, \qquad
s_3 = \tfrac{\tilde{L}}{2} + s_{int}, 
\end{equation*}
\begin{equation*}
s_4 = \tfrac{3\tilde{L}}{2} - s_{int}, \qquad
s_5 = \tilde{L} + s_{int}, \qquad
s_6 = \tfrac{3\tilde{L}}{2} + s_{int}.
\end{equation*}

From Figure~3, we can see that:
\begin{itemize}
    \item If $s_0$ starts between $0$ and $s_1$, the next self-intersection point is $s_3$. 
    Therefore, if $s_0 + L < s_3$ there is no self-intersection.
    
    \item If $s_0 \in (s_1, s_2]$, the next self-intersection point is $s_5$. 
    Therefore, if $s_0 + L < s_5$ there is no self-intersection.
    
    \item If $s_0 \in (s_2, \tilde{L})$, the next self-intersection point is $s_6$. 
    Therefore, if $s_0 + L < s_6$, there is no self-intersection.
\end{itemize}

The next figure also shows two elastica periods, but for Region~3 of $\mathrm{k}$ values. 
In this case, $s_1, \ldots, s_6$ are computed differently.

\begin{figure}[H]\centering
\includegraphics[width=\linewidth]{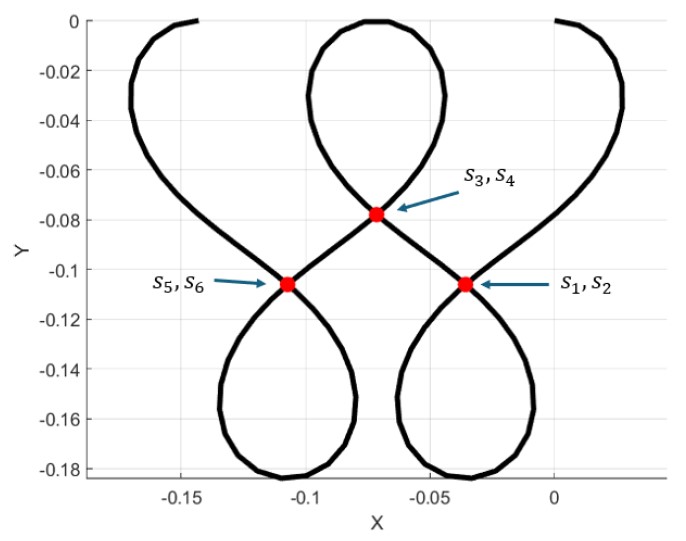}
\caption{\centering Two full periods of the elastica ($\tilde{L}=L/2$) and the six self-intersection points along the periods for Region~3 of $\mathrm{k}$ values.}
\end{figure}

For Region~3 of $\mathrm{k}$ values, the values $s_1 < s_2 < s_3 < s_4 < s_5 < s_6$ are computed as:

\begin{equation*}
s_1 = \tfrac{\tilde{L}}{2} - s_{int}, \qquad
s_2 = \tfrac{\tilde{L}}{2} + s_{int}, \qquad
s_3 = \tilde{L} - s_{int}, \qquad
\end{equation*}
\begin{equation*}
s_4 = \tilde{L} + s_{int}, \qquad
s_5 = \tfrac{3\tilde{L}}{2} - s_{int}, \qquad
s_6 = \tfrac{3\tilde{L}}{2} + s_{int}.
\end{equation*}

From Figure~4, we can see that:
\begin{itemize}
    \item If $s_0$ starts between $0$ and $s_1$, the next self-intersection point is $s_2$. 
    Therefore, if $s_0 + L < s_2$ there is no self-intersection.
    
    \item If $s_0 \in (s_1, s_3]$, the next self-intersection point is $s_4$. 
    Therefore, if $s_0 + L < s_4$, there is no self-intersection.
    
    \item If $s_0 \in (s_3, \tilde{L})$, the next self-intersection point is $s_6$. 
    Therefore, if $s_0 + L < s_6$, there is no self-intersection.
\end{itemize}

After expressing these inequality conditions as functions of $s_{int}$ and $\tilde{L}$, 
we obtain identical expressions and domains for both $\mathrm{k}$ regions. 
Therefore, Algorithm~6 describes the self-intersection check for all $\mathrm{k}$ values, 
$0 \leq \mathrm{k} < 1$, within the inflectional elastica domain (including the figure-eight shape).

{\fontsize{10}{10}\selectfont
\begin{algorithm}{}
\caption{\bf CheckSelfIntersection}
\begin{spacing}{0.95}
\begin{algorithmic}[0]
\renewcommand{\thealgorithm}{}

\Statex {\bf Input:} $\mathrm{k}$, $\tilde{L}$, $s_0$, cable length $L$, tolerance $\epsilon$.
\Statex {\bf Compute:} $s_{\text{int}} = \text{GetSint}(\mathrm{k}, \tilde{L}, \epsilon)$.
\Statex {\bf Output:} {\bf True} if a self–intersection occurs, otherwise {\bf False}.

\Statex {\bf Early exit:}
\begin{enumerate}{}
  \item If $\mathrm{k} \leq \mathrm{k}_{\max}$, {\bf return} {\bf False}.
\end{enumerate}

\Statex {\bf Region 2 \& 3:} ($\mathrm{k} \in (\mathrm{k}_{\max},\, 1)$)
\begin{enumerate}{}
  \item If $s_0 \in \big[0,\, \tfrac{\tilde{L}}{2}-s_{\text{int}}\big]$:
        \begin{itemize}
          \item If $s_0 + L < \tfrac{\tilde{L}}{2} + s_{\text{int}}$, {\bf return} {\bf False}; else {\bf return} {\bf True}.
        \end{itemize}
  \item If $s_0 \in \big(\tfrac{\tilde{L}}{2}-s_{\text{int}},\, \tilde{L}-s_{\text{int}}\big]$:
        \begin{itemize}
          \item If $s_0 + L < \tilde{L} + s_{\text{int}}$, {\bf return} {\bf False}; else {\bf return} {\bf True}.
        \end{itemize}
  \item If $s_0 \in \big(\tilde{L}-s_{\text{int}},\, \tilde{L}\big)$:
        \begin{itemize}
          \item If $s_0 + L < \tfrac{3\tilde{L}}{2} + s_{\text{int}}$, {\bf return} {\bf False}; else {\bf return} {\bf True}.
        \end{itemize}
\end{enumerate}

\end{algorithmic}
\end{spacing}
\end{algorithm}
}

\section*{Examples}
This section presents experiments conducted to evaluate the algorithm for self-intersection detection.  

Example~1 demonstrates Region~2 of the $\mathrm{k}$ values.  
In part (a) of Fig.~5, one can see an elastica shape with parameters 
$\mathrm{k} = 0.88$, $\tilde{L} = L$, $s_{0} = 0.23\tilde{L}$.  
In this case, Algorithm~3 returned \textbf{FALSE}, meaning no self-intersection was detected.  

In part (b) of Fig.~5, a small change was made in the elastica parameters so that only $s_{0}$ was modified to $s_{0} = 0.2\tilde{L}$.  
In this case, the algorithm detected a self-intersection.

\begin{figure}[H]\centering
\includegraphics[width=\linewidth]{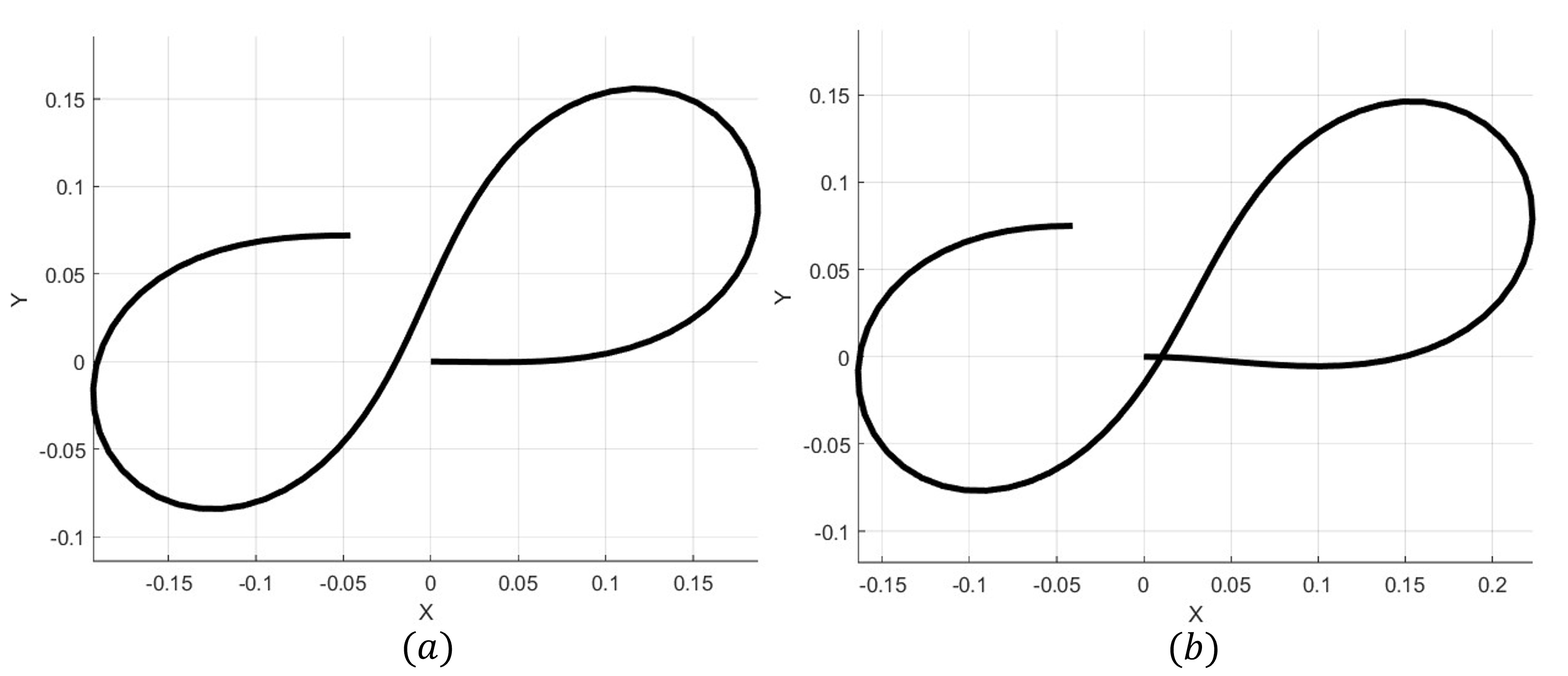}\hfill
\caption{Example of self-intersection detection by Algorithm~3 in Region~2 of the $\mathrm{k}$ values. 
Part (a) shows an elastica with parameters $\mathrm{k} = 0.88$, $\tilde{L} = L$, $s_{0} = 0.23\tilde{L}$, where no self-intersection was detected. 
Part (b) shows an elastica with parameters $\mathrm{k} = 0.88$, $\tilde{L} = L$, $s_{0} = 0.2\tilde{L}$, where a self-intersection was detected.}
\end{figure}

Example~2 demonstrates Region~3 of the $\mathrm{k}$ values.  
In part (a) of Fig.~6, one can see an elastica shape with parameters 
$\mathrm{k} = 0.99$, $\tilde{L} = 2.7L$, $s_{0} = 0.75\tilde{L}$.  
In this case, Algorithm~3 returned \textbf{FALSE}, meaning no self-intersection was detected.  

In part (b) of Fig.~6, a small change was made in the elastica parameters so that only $\tilde{L}$ was modified to $\tilde{L} = 2.5L$.  
In this case, the algorithm detected a self-intersection.

\begin{figure}[H]\centering
\includegraphics[width=\linewidth]{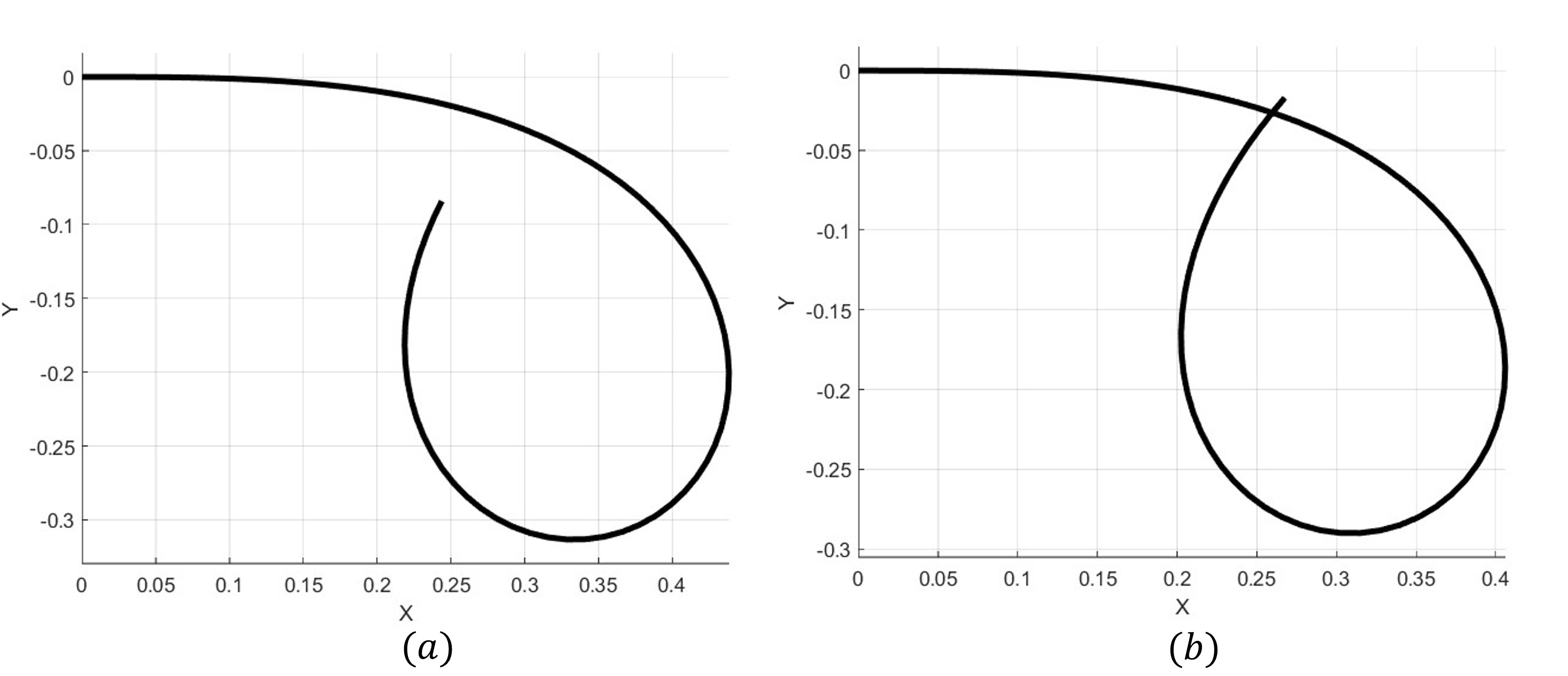}\hfill
\caption{Example of self-intersection detection by Algorithm~3 in Region~3 of the $\mathrm{k}$ values. 
Part (a) shows an elastica with parameters $\mathrm{k} = 0.99$, $\tilde{L} = 2.7L$, $s_{0} = 0.75\tilde{L}$, where no self-intersection was detected. 
Part (b) shows an elastica with parameters $\mathrm{k} = 0.99$, $\tilde{L} = 2.5L$, $s_{0} = 0.75\tilde{L}$, where a self-intersection was detected.}
\end{figure}

\vspace{-.06in} 
\section*{\bf Appendix D: Static Elastica Experiments}
\vspace{-.03in}
\noindent This appendix presents static-shape experiments that compare
the planar Euler elastica solutions used in the paper with the
shape of a real flexible cable. The purpose is to experimentally
validate that the elastica model provides an accurate description
of the cable shape during quasi-static manipulation.

\begin{figure*}[!t]
  \centering
  \includegraphics[width=\textwidth]{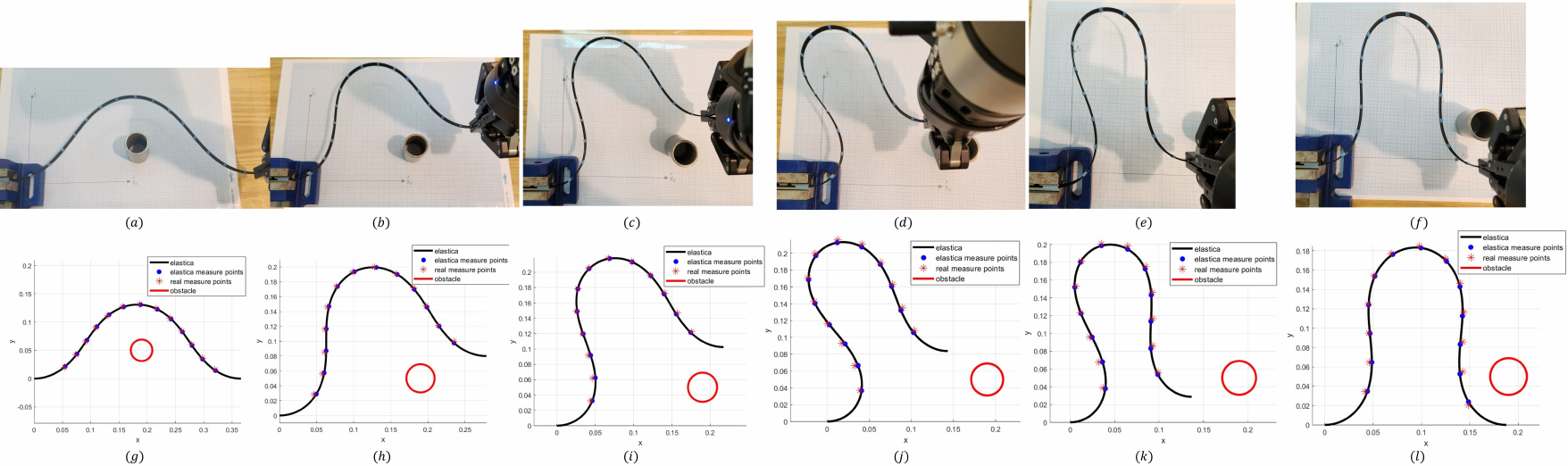}
  \caption{Static shapes of the cable tie and corresponding
  elastica solutions. Parts (a)–(f) show the measured cable-tie
  shapes in the horizontal plane. Parts (g)–(l) show the
  corresponding planar elastica solutions. In the elastica plots,
  the black curve represents the elastica, blue markers denote
  sampling points on the elastica, red markers denote the
  measured cable-tie points, and the red circle marks the
  cylindrical obstacle used in some of the shapes.}
  \label{fig:static-elastica-exp}
\end{figure*}

\subsection{Experimental Setup}

The experiments were conducted with a plastic cable tie (zip tie)
of length $L_{\mathrm{zip}} = 470\,\mathrm{mm}$, width
$w_{\mathrm{zip}} = 9\,\mathrm{mm}$, and thickness
$t_{\mathrm{zip}} = 1\,\mathrm{mm}$. One end was rigidly clamped, while
the other end was grasped by a Robotiq 2F-85 parallel gripper
mounted on a UR5e manipulator.

The cable tie was manipulated in a horizontal plane for
static-shape acquisition. A millimeter-grid paper sheet was placed beneath the tie for measurement, and the coordinates
of marked points along the tie were manually recorded.

Fifteen approximately uniform arc-length measurement
points were defined along the tie. In practice, points
$i = 2,\dots,14$ were measured due to limited access near the
clamped and grasped endpoints.

\subsection{Measurement Errors}

Three primary error sources affect the measurements:

\begin{itemize}
  \item \textbf{Grid and thickness error:}
  Due to the finite width of the millimeter-grid paper and the
  cable-tie thickness, each measured point has an inherent
  uncertainty of approximately $t_{\mathrm{zip}}/2$ in the direction
  normal to the elastica.

  \item \textbf{Arc-length projection error:}
  Manual projection of points onto the grid introduces small
  errors $\Delta s_i$ along the curve tangent, with an unknown sign.

  \item \textbf{Systematic errors:}
  Residual curvature of the tie, surface unevenness, and
  accumulated plastic deformation may introduce additional
  biases, but all remained small relative to the overall shape.
\end{itemize}
A preliminary error measurement
was performed on a fully stretched zip-tie. The average markers
measurement error was determined to be a disc of radius 1.2 mm.

\subsection{Tested Shapes and Notation}

Six planar shapes were recorded, denoted
$\mathrm{shape}\ 1,\dots,6$. For each shape $j$, the elastica and
measured coordinates of point $i$ are denoted respectively as:

\[
  \mathbf{p}^{\mathrm{e}}_{i,j}
  = (x^{\mathrm{e}}_{i,j}, y^{\mathrm{e}}_{i,j}), \qquad
  \mathbf{p}^{\mathrm{z}}_{i,j}
  = (x^{\mathrm{z}}_{i,j}, y^{\mathrm{z}}_{i,j}).
\]

The clamped endpoint is $\mathbf{p}^{\mathrm{e}}_{0,j} = (0,0)$ for all shapes.

\begin{table}[!t]
  \centering
  \caption{Elastica parameters and end points for the six static shapes.}
  \label{tab:appC-params}
  \small
  \begin{tabular}{c c c c c c}
    \hline
    Shape & $\mathrm{k}$ & $\tilde{L}$ & $s_0$ & $x(L) [mm]$ & $y(L) [mm]$  \\
    \hline
    1 & 0.4637 & $L_{zip}$ & 0.2350 & 365.87 & 0 \\
    2 & 0.6021 & $L_{zip}$ & 0.2193 & 278.81 & 79.99 \\
    3 & 0.6743 & $L_{zip}$ & 0.2135 & 217.25 & 102.37 \\
    4 & 0.7640 & $L_{zip}$ & 0.2135 & 143.21 & 83.76 \\
    5 & 0.7917 & $L_{zip}$ & 0.2272 & 136.59 & 28.75\\
    6 & 0.7395 & $L_{zip}$ &  0.2350 & 187.75 & 0 \\
    \hline
  \end{tabular}
\end{table}

\begin{table*}[!t]
  \centering
  \caption{Elastica coordinates $\mathbf{p}^{\mathrm{e}}_{i,j}$ (mm).}
  \label{tab:appC-elastica-pts}
  \small
  \setlength{\tabcolsep}{4pt}
  \begin{tabular}{ccccccc}
    \hline
    $i$ & Shape 1 & Shape 2 & Shape 3 & Shape 4 & Shape 5 & Shape 6 \\
    \hline
     1 & (28.79, 5.58) & (28.16, 7.57) & (27.74, 8.61) & (26.98,10.20) & (26.60,10.95) & (27.22, 9.80) \\
     2 & (54.73,21.36) & (49.60,28.82) & (46.16,32.36) & (40.68,36.74) & (39.51,37.96) & (44.15,34.70) \\
     3 & (75.64,43.49) & (60.22,57.23) & (50.24,62.33) & (36.38,66.59) & (36.33,68.06) & (48.62,64.72) \\
     4 & (93.18,67.83) & (62.87,87.08) & (44.07,91.62) & (21.08,92.29) & (24.68,95.66) & (46.66,94.65) \\
     5 & (110.75,91.51)& (63.09,116.58)& (34.25,119.43)& ( 2.40,115.11)& (12.11,122.34)& (45.38,124.09)\\
     6 & (132.50,112.80)& (66.63,146.81)& (26.46,148.87)& (-14.66,140.32)& ( 5.13,151.88)& (51.75,153.73)\\
     7 & (158.37,126.69)& (78.23,173.75)& (27.38,178.18)& (-22.29,168.57)& (11.67,180.24)& (70.29,176.17)\\
     8 & (188.43,130.53)& (100.88,193.69)& (42.03,204.47)& (-13.55,197.20)& (35.09,198.57)& (99.36,182.96)\\
     9 & (217.67,122.52)& (130.47,199.37)& (68.96,217.67)& (11.87,212.70)& (64.55,194.67)& (125.87,169.11)\\
    10 & (242.15,105.38)& (158.67,189.98)& (98.24,213.27)& (40.85,207.50)& (84.20,172.63)& (139.75,142.88)\\
    11 & (262.23, 82.47)& (181.47,169.91)& (122.56,195.21)& (62.91,186.85)& (91.14,143.12)& (142.39,112.60)\\
    12 & (279.26, 58.39)& (198.52,145.87)& (139.94,171.43)& (76.34,160.65)& (90.80,113.64)& (140.05, 83.19)\\
    13 & (297.92, 34.29)& (215.06,120.25)& (155.64,145.29)& (87.42,132.24)& (90.70, 83.16)& (139.72, 52.74)\\
    14 & (320.73, 14.16)& (235.19, 97.45)& (174.48,121.39)& (102.40,105.78)& (98.53, 53.89)& (148.60, 23.81)\\
    15 & (348.04,  2.11)& (261.09, 82.72)& (199.60,105.43)& (125.70, 87.38)& (119.24, 32.88)& (170.29,  3.78)\\
    \hline
  \end{tabular}
\end{table*}

\begin{table*}[!t]
  \centering
  \caption{Measured cable-tie coordinates $\mathbf{p}^{\mathrm{z}}_{i,j}$ (mm).}
  \label{tab:appC-zip-pts}
  \small
  \setlength{\tabcolsep}{4pt}
  \begin{tabular}{ccccccc}
    \hline
    $i$ & Shape 1 & Shape 2 & Shape 3 & Shape 4 & Shape 5 & Shape 6 \\
    \hline
     2 & (53.82,19.62) & (46.93,28.24) & (44.97,32.32) & (39.50,36.93) & (37.50,38.44) & (42.47,34.34) \\
     3 & (73.89,42.09) & (57.99,55.90) & (47.50,62.04) & (32.66,66.69) & (32.98,68.15) & (46.50,64.50) \\
     4 & (91.91,65.21) & (61.50,85.49) & (41.48,91.65) & (17.90,93.30) & (22.45,96.22) & (46.00,95.05) \\
     5 & (108.88,89.68) & (62.00,115.99)& (33.97,119.66)& ( 0.39,116.81)& (11.96,123.19)& (45.50,124.48) \\
     6 & (130.80,111.10)& (64.99,145.88)& (26.49,149.07)& (-15.05,142.72)& ( 6.50,153.02)& (51.46,153.80) \\
     7 & (156.66,126.53)& (76.91,173.22)& (27.99,178.38)& (-22.50,171.01)& (12.93,181.25)& (68.77,176.58) \\
     8 & (186.46,130.50)& (99.22,193.05)& (42.35,205.14)& (-14.10,199.20)& (35.13,200.27)& (97.93,184.50) \\
     9 & (215.78,123.55)& (127.96,199.50)& (67.57,218.00)& (12.08,215.51)& (64.25,197.57)& (124.64,171.66) \\
    10 & (239.66,106.13)& (156.74,190.07)& (97.30,213.04)& (40.76,210.06)& (84.55,174.79)& (139.52,146.38) \\
    11 & (260.60,83.70) & (180.12,171.18)& (121.33,195.66)& (62.59,188.72)& (92.00,145.97)& (143.50,116.52) \\
    12 & (278.59,60.21) & (197.08,147.23)& (140.07,172.74)& (77.04,163.31)& (92.50,116.52)& (142.00, 86.04) \\
    13 & (297.13,36.17) & (213.59,121.21)& (156.58,146.73)& (88.54,134.80)& (92.50, 85.96)& (142.00, 54.95) \\
    14 & (319.72,15.58) & (234.67, 99.12)& (175.15,122.14)& (102.61,108.19)& (98.56, 55.77)& (148.07, 20.25) \\
    \hline
  \end{tabular}
\end{table*}

\subsection{Error Metrics}

The absolute error of point $i$ in shape $j$ is

\[
e_{i,j}
= \bigl\| \mathbf{p}^{\mathrm{e}}_{i,j}
       - \mathbf{p}^{\mathrm{z}}_{i,j} \bigr\|.
\]

Following the measurement protocol in the experimental
document, the relative error is defined as

\[
\varepsilon_{i,j}
= 100\,
  \frac{\big|\|\mathbf{p}^{\mathrm{e}}_{i,j}\| - \| \mathbf{p}^{\mathrm{z}}_{i,j}\|\big|}
       {\|\mathbf{p}^{\mathrm{e}}_{i,j} - \mathbf{p}^{\mathrm{e}}_{0,j}\|},
\]
that is, the absolute deviation divided by the Euclidean
distance from the clamped endpoint.


\begin{table}[!t]
  \centering
  \caption{Absolute errors $e_{i,j}$ (mm).}
  \label{tab:appC-errors-mm}
  \small
  \setlength{\tabcolsep}{3pt}
  \begin{tabular}{ccccccc}
    \hline
    $i$ & Shape 1 & Shape 2 & Shape 3 & Shape 4 & Shape 5 & Shape 6 \\
    \hline
     2 & 1.96 & 2.73 & 1.19 & 1.20 & 2.06 & 1.71 \\
     3 & 2.24 & 2.59 & 2.75 & 3.72 & 3.35 & 2.13 \\
     4 & 2.90 & 2.09 & 2.59 & 3.34 & 2.29 & 0.77 \\
     5 & 2.61 & 1.24 & 0.36 & 2.63 & 0.85 & 0.40 \\
     6 & 2.39 & 1.88 & 0.20 & 2.43 & 1.77 & 0.30 \\
     7 & 1.72 & 1.42 & 0.64 & 2.45 & 1.61 & 1.57 \\
     8 & 1.96 & 1.77 & 0.74 & 2.07 & 1.69 & 2.10 \\
     9 & 2.14 & 2.50 & 1.43 & 2.81 & 2.90 & 2.83 \\
    10 & 2.60 & 1.93 & 0.96 & 2.56 & 2.18 & 3.50 \\
    11 & 2.03 & 1.84 & 1.30 & 1.89 & 2.97 & 4.07 \\
    12 & 1.94 & 1.98 & 1.31 & 2.75 & 3.35 & 3.45 \\
    13 & 2.04 & 1.75 & 1.72 & 2.79 & 3.32 & 3.17 \\
    14 & 1.74 & 1.75 & 1.00 & 2.42 & 1.87 & 3.59 \\
    \hline
    Mean & 2.17 & 1.96 & 1.24 & 2.54 & 2.39 & 2.27 \\
    Std.~dev. & 0.35 & 0.43 & 0.75 & 0.62 & 0.79 & 1.27 \\
    \hline
  \end{tabular}
\end{table}

\begin{table}[!t]
  \centering
  \caption{Relative errors $\varepsilon_{i,j}$ (\%).}
  \label{tab:appC-errors-rel}
  \small
  \setlength{\tabcolsep}{3pt}
  \begin{tabular}{ccccccc}
    \hline
    $i$ & Shape 1 & Shape 2 & Shape 3 & Shape 4 & Shape 5 & Shape 6 \\
    \hline
     2 & 2.49 & 4.52 & 1.77 & 1.35 & 1.99 & 2.73 \\
     3 & 2.53 & 3.04 & 2.40 & 2.14 & 1.86 & 1.77 \\
     4 & 2.21 & 1.94 & 1.04 & 0.36 & 0.01 & 0.06 \\
     5 & 1.81 & 0.78 & 0.11 & 1.45 & 0.66 & 0.30 \\
     6 & 1.37 & 0.94 & 0.13 & 1.72 & 0.78 & 0.02 \\
     7 & 0.70 & 0.54 & 0.16 & 1.44 & 0.60 & 0.08 \\
     8 & 0.70 & 0.60 & 0.34 & 1.03 & 0.83 & 0.32 \\
     9 & 0.45 & 0.52 & 0.04 & 1.31 & 1.29 & 0.62 \\
    10 & 0.74 & 0.46 & 0.25 & 1.17 & 1.08 & 1.17 \\
    11 & 0.42 & 0.04 & 0.11 & 0.84 & 1.68 & 1.83 \\
    12 & 0.09 & 0.14 & 0.49 & 1.52 & 2.28 & 1.92 \\
    13 & 0.19 & 0.32 & 0.78 & 1.73 & 2.61 & 1.95 \\
    14 & 0.29 & 0.06 & 0.46 & 1.28 & 0.83 & 0.69 \\
    \hline
    Mean      & 1.08 & 1.07 & 0.63 & 1.34 & 1.27 & 1.04 \\
    Std.~dev. & 0.89 & 1.33 & 0.72 & 0.44 & 0.76 & 0.91 \\
    \hline
  \end{tabular}
\end{table}

\subsection{Discussion}

Across all six shapes, the absolute and relative errors remain
small. Absolute deviations are typically a few millimeters,
while relative errors mostly remain below $2\%$. These results
validate the suitability of Euler’s elastica as an accurate model
for the planar equilibrium shapes of the flexible cable used in
this work.

\newpage

\printbibliography

@book{ben-asher,
         AUTHOR     ="J. Z. Ben-Asher",
         TITLE      ="Optimal Control Theory with Aerospace Applications",
         PUBLISHER  ="AIAA Inc.",
         YEAR       ="2010",
         ADDRESS     ="Reston, Virginia"
         }

@book{gelfand,
         AUTHOR     = "I. M. Gelfand and S. V. Fomin",
         TITLE      = "Calculus of Variations",
         PUBLISHER  ="Moscow State University, published in english by Prentice-Hall then by Dover",
         YEAR       ="1963",
         ADDRESS     ="NY"
         }

@book{love,
         AUTHOR     ="A. Love",
         TITLE      = "The Mathematical Theory of Elasticity",
         PUBLISHER  ="Dover, New York",
         YEAR       ="1944",
         ADDRESS     =""
         }

@book{pontryagin,
         AUTHOR     ="L. S. Pontryagin and V. G. Boltyanskii  and R. V. Gamkrelidze and E. F. Mishchenko",
         TITLE      ="The Mathematical Theory of Optimal Processes",
         PUBLISHER  ="Interscience Publishers John Wiley \& Sons, New York-London",
         YEAR       ="1962",
         ADDRESS     =""
         }

@book{ellip_book,
         AUTHOR     = "V. Prasolov and Y. Solovyev",
         TITLE      = "Elliptic Functions and Elliptic Integrals",
         PUBLISHER  = "",
         YEAR       = "1997",
         ADDRESS     = "AMS, Providence, RI"
         }

@book{timoshenko&goodier:1969,
    AUTHOR = {S. P. Timoshenko and J. N. Goodier},
    TITLE  = {Theory of Elasticity},
    EDITION = {3rd},
    PUBLISHER = {McGraw-Hill},
    Address = {New York},
    YEAR = {1969}
}

@String{SIAM="SIAM Journal on Control and Optimization"}

@ARTICLE{adaptive_LDO22,
     AUTHOR  = "O. Aghajanzadeh and  M. Aranda and  J. A. C. Ramon and C. Cariou and R. Lenain and  Y. Mezouar",
     TITLE   = "Adaptive Deformation Control for Elastic Linear Objects",
     JOURNAL = "Frontiers in Robotics and AI",
     VOLUME  = "9",
     NUMBER  = "",
     PAGES   = "1-13",
     YEAR    = "Article 868459, 2022",
    }

@ARTICLE{balkcom15,
      AUTHOR=   "W. Wang and M. Bell and D. Balkcom",
      TITLE =   "Towards Arranging and Tightening Knots and Unknots With Fixtures",
      JOURNAL=  "IEEE Trans. on Automation Science and Engineering",
      YEAR   =  "2015",
      VOLUME  = "12",
     NUMBER  = "",
      PAGES=   "1318-1331"
     }

@ARTICLE{batista_matlab,
      AUTHOR=   "M. Batista",
      TITLE =   "\mbox{Elfun18} – A collection of \mbox{MATLAB} functions for the computation of
                          elliptic integrals and Jacobian elliptic functions of real arguments",
      JOURNAL=  "SoftwareX, http://www.sciencedirect.com/journal/softwarex",
      YEAR   =  "2019",
      VOLUME  = "10",
      NUMBER  = "",
      PAGES=   "1-10"
     }

@ARTICLE{bezier,
      AUTHOR=   "P. E. B\'ezier and S. Sioussio",
      TITLE =   "Semi-Automatic System for Defining Free-Form Curves and Surfaces",
      JOURNAL=  "Computer Aided Design",
      YEAR   =  "1983",
      VOLUME  = "5",
     NUMBER  = "2",
      PAGES=   "65-72"
     }

@ARTICLE{brander,
      AUTHOR=   "D. Brander and A. Baerentzen and A.S. Fisker and J. Gravesen",
      TITLE =   "B\'ezier curves that are close to elastica",
      JOURNAL=  "Science Direct",
      YEAR   =  "2017",
      VOLUME  = "104",
     NUMBER  = "",
      PAGES=   "36-44"
     }

@INPROCEEDINGS{djondjorov,
         AUTHOR   ="P. A. Djondjorov and M. T.  Haddzhilazova and I. M. Mladenov and V. M. Vassilev",
         TITLE    ="Explicit Parametrization of Euler's Elastica",
         BOOKTITLE="Int Conf. on Geometry, Integrability and Quantization",
         PAGES = "175-186",
         YEAR     ="2008"
         }

@ARTICLE{bretl_ijrr14,
     AUTHOR  = "T. Bretl and Z. McCarthy",
     TITLE   = "Quasi-Static Manipulation of a Kirchhoff Elastic Rod Based on Geometric Analysis of Equilibrium Configurations",
     JOURNAL = "The Int, J. of Robotics Research",
     VOLUME  = "33",
     NUMBER  = "1",
     PAGES   = "48-68",
      YEAR    = "2014"
   }

@INPROCEEDINGS{bretl_multiple,
     AUTHOR   ="M. Mukadam and A. Borum and T. Bretl",
     TITLE    ="\mbox{IEEE/RSJ} Int. Conf. on Intelligent Robots and Systems",
     BOOKTITLE="Quasi-Static Manipulation of a Planar Elastic Rod Using Multiple Robotic Grippers",
     PAGES    ="55-60",
     YEAR     ="2014"
     }

@ARTICLE{opt_survey,
     AUTHOR  = "R. F. Hartl and S. P. Sethi and R. G. Vickson",
     TITLE   = "A Survey of the Maximum Principles for Optimal Control Problems with State Constraints",
     JOURNAL = "\mbox{SIAM} Review",
     YEAR    = "1995",
     VOLUME  = "37",
     NUMBER  = "2",
     PAGES   = "181-218"
        }

@article{hirai_knot,
author = {H. Wakamatsu and E. Arai and S. Hirai},
title ={Knotting/Unknotting Manipulation of Deformable Linear Objects},
journal = {The Int. J. of Robotics Research},
volume = {25},
number = {4},
pages = {371-395},
year = {2006}
}

@INPROCEEDINGS{jackson&cavusogl,
           AUTHOR  =  "R. C. Jackson and M. C. Cavusoglu",
           TITLE   =  "Needle Path Planning for Autonomous Robotic Surgical Suturing",
           BOOKTITLE= "IEEE Int. Conf. on Robotics and Automation",
           YEAR    = "2013",
           PAGES   = "1669-1675"
               }

@INPROCEEDINGS{javdani,
  author={S. Javdani and S. Tandon and J. Tang and J. F. O'Brien and P. Abbeel},
  booktitle={IEEE Int. Conf. on Robotics and Automation},
  title={Modeling and Perception of Deformable One-Dimensional Objects},
  year={2011},
  pages={1607-1614}
  }

@ARTICLE{cable_routing11,
     AUTHOR  = "X.   Jiang  and  K.-M.   Koo  and  K.   Kikuchi  and  A.  Konno  and  M.  Uchiyama ",
     TITLE   = "Robotized Assembly of a Wire Harness in a Car Production Line",
     JOURNAL = "Advanced Robotics",
     VOLUME  = "25",
     NUMBER  = "3-4",
     PAGES   = "473-489",
     YEAR    = "2011",
    }

@ARTICLE{kavraki06,
      AUTHOR  = "M. Moll and L. E. Kavraki",
      TITLE =   "Path Planning for Deformable Linear Objects",
      JOURNAL=  "IEEE Transactions on Robotics",
      YEAR   =  "2006",
      VOLUME  = "22",
      NUMBER  = "4",
      PAGES=   "625–636"
     }

@ARTICLE{lagneau20,
     AUTHOR  = "R. Lagneau and A. Krupa and M. Marchal",
     TITLE   = "Automatic Shape Control of Deformable Wires based on Model-Free Visual Servoing",
     JOURNAL = " IEEE Robotics and Automation Letters,",
     VOLUME  = "5",
     NUMBER  = "4",
     PAGES   = "5252 - 5259",
     YEAR    = "2020"
    }

@article{sachkov07,
author = {Y. L. Sachkov},
title = {Maxwell Strata in the Euler Elastic Problem},
journal = {J. of Dynamical and Control Systems},
volume = {14},
pages = {169-234},
year = {2008}
}

@article{sachkov_conjugate,
  title={Conjugate Points in Euler's Elastic Problem},
  author={Y. L. Sachkov},
  journal={J. of Dynamical and Control Systems},
  volume = {14},
  pages = {409-439},
  year={2008}
}

@article{sachkov_vertices,
  title={Stability of Inflectional Elastica Centered at Vertices or Inflection Points},
  author={Y. L. Sachkov and S. V. Levyakov},
  journal={Proceedings of the Steklov Institute of Mathematics },
  volume = {271},
  pages = {177–192},
  year={2011}
}

@PHDTHESIS{goss_thesis,
         AUTHOR   ="V. G. A. Goss",
         TITLE    ="Snap Buckling, Writhing and Loop Formation in Twisted Rods",
         SCHOOL="Center for Nonlinear Dynamics, University College of London",
         YEAR     ="2003",
         }

@ARTICLE{sintov20,
  author={A. Sintov and S. Macenski and A. Borum and T. Bretl},
  journal={IEEE Robotics and Automation Letters},
  title={Motion Planning for Dual-Arm Manipulation of Elastic Rods},
  year={2020},
  volume={5},
  number={4},
  pages={6065-6072}
}

@article{till17,
author = {J. Till and D. Rucker},
year = {2017},
month = {},
pages = {1-16},
title = {Elastic Stability of Cosserat Rods and Parallel Continuum Robots},
volume = {33},
number = {3},
journal = {IEEE Transactions on Robotics}
}

@INPROCEEDINGS{cable_rss22,
         AUTHOR   ="V. Viswanath and K. Shivakumar and J. Kerr and al.",
         TITLE    ="Autonomously Untangling Long Cables",
         BOOKTITLE="Robotics:  Science and Systems, DOI:10.15607/RSS.2022.XVIII.034",
         YEAR     ="2022",
         PAGES    ="",
         ADDRESS     =""
       }

@ARTICLE{food_survey22,
     AUTHOR  = "Z. Wang and S. Hirai and S. Kawamura",
     TITLE   = "Challenges and Opportunities in Robotic Food Handling: A Review",
     JOURNAL = "Frontiers in Robotics and AI",
     VOLUME  = "8",
     NUMBER  = "",
     PAGES   = "1-12",
     YEAR    = "Article 789107, 2022",
    }

@ARTICLE{levyakov09,
     AUTHOR  = "S.V. Levyakov",
     TITLE   = "Stability Analysis of Curvilinear Configurations of an Inextensible Elastic Rod with Clamped Ends",
     JOURNAL = "Mechanics Research Communications",
     VOLUME  = "36",
     NUMBER  = "5",
     PAGES   = "612-617",
     YEAR    = "2009"
    }

@ARTICLE{levyakov10,
     AUTHOR  = "S.V. Levyakov and V. V. Kuznetsov",
     TITLE   = "Stability Analysis of Planar Equilibrium Configurations of Elastic Rods Subjected to End Loads",
     JOURNAL = "Acta Mech",
     VOLUME  = "211",
     NUMBER  = "",
     PAGES   = "73–87",
     YEAR    = "2010"
    }

@ARTICLE{batista15,
     AUTHOR  = "M. Batista",
     TITLE   = "On Stability of Elastic Rod Planar Equilibrium Configurations",
     JOURNAL = "Int. J. of Solids and Structures",
     VOLUME  = "72",
     NUMBER  = "",
     PAGES   = "144-152",
     YEAR    = "2015"
    }

@ARTICLE{zhu_survey22,
     AUTHOR  = "J. Zhu and A. Cherubini and C. Dune and D. Navarro-Alarcon and F. Alambeigi and D. Berenson and F. Ficuciello
     and K. Harada and J. Kober and X. LI and J. Pan and W. Yuan and M. Gienger",
     TITLE   = "Challenges and Outlook in Robotic Manipulation of Deformable Objects",
     JOURNAL = "IEEE Robotics  Automation Magazine",
     VOLUME  = "",
     NUMBER  = "",
     PAGES   = "2-12",
     YEAR    = "Early Access, 2022"
   }

@INPROCEEDINGS{zhu_iros18,
  author={J. Zhu  and B. Navarro and P. Fraisse and A. Crosnier and A. Cherubini},
  booktitle={2018 IEEE/RSJ Int. Conf. on Intelligent Robots and Systems},
  title={Dual-Arm Robotic Manipulation of Flexible Cables},
  year={2018},
  volume={},
  number={},
  pages={479-484}
  }
\end{document}